\UseRawInputEncoding
\documentclass{book}
\usepackage{roboto}

\usepackage{footmisc}

\usepackage[english]{babel}
\usepackage[letterpaper,top=2cm,bottom=2cm,left=3cm,right=3cm,marginparwidth=1.75cm]{geometry}

\usepackage{pdflscape}
\usepackage{lscape}

\usepackage{rotating}
\usepackage{longtable}
\usepackage{array}
\usepackage{float}
\usepackage{amsmath}
\usepackage{graphicx}
\usepackage{inconsolata}
\usepackage[colorlinks=true, allcolors=blue]{hyperref}
\usepackage{enumitem}

\usepackage{tikz} 

\usepackage{listings}
\usepackage{xcolor}

\usepackage[T1]{fontenc}
\usepackage{inconsolata}  

\definecolor{bgcolor}{rgb}{0.97,0.97,0.97}
\definecolor{codeblue}{rgb}{0.1,0.1,0.8}
\definecolor{codegreen}{rgb}{0,0.4,0}
\definecolor{codegray}{rgb}{0.4,0.4,0.4}
\definecolor{codepurple}{rgb}{0.5,0,0.5}
\definecolor{codered}{rgb}{0.6,0.2,0.2}
\definecolor{lightgray}{rgb}{0.9,0.9,0.9}
\definecolor{darkgray}{rgb}{0.6,0.6,0.6} 

\makeatletter
\renewcommand{\paragraph}{%
  \@startsection{paragraph}{4}{\z@}{1ex}{-1em}{\normalfont\normalsize\bfseries\color{darkgray}}}
\makeatother

\lstdefinestyle{python}{
    language=Python,
    basicstyle=\ttfamily\small\color{black}\usefont{T1}{zi4}{m}{n},  
    keywordstyle=\bfseries\color{codeblue},  
    stringstyle=\color{codegreen},  
    commentstyle=\slshape\color{codegray},  
    showstringspaces=false,
    numbers=left,
    numberstyle=\tiny\color{codegray},  
    stepnumber=1,
    numbersep=8pt,
    frame=single,
    rulecolor=\color{darkgray},  
    breaklines=true,
    backgroundcolor=\color{bgcolor},
    tabsize=4,
    captionpos=b,
    morekeywords={self}, 
}

\lstdefinestyle{text}{
    language=,
    basicstyle=\ttfamily\small\color{black}\usefont{T1}{zi4}{m}{n},  
    stringstyle=\color{codered},
    commentstyle=\color{codegray},
    showstringspaces=false,
    numbers=none,
    frame=single,
    rulecolor=\color{lightgray},  
    frameround=tttt,
    breaklines=true,
    backgroundcolor=\color{bgcolor},
    tabsize=4,
    captionpos=b,
}

\lstdefinestyle{cmd}{
    language=bash,
    basicstyle=\ttfamily\small\color{black}\usefont{T1}{zi4}{m}{n},  
    keywordstyle=\bfseries\color{blue},
    stringstyle=\color{codegreen},
    commentstyle=\itshape\color{gray},
    showstringspaces=false,
    numbers=none,
    frame=single,
    rulecolor=\color{darkgray},  
    breaklines=true,
    backgroundcolor=\color{bgcolor},
    tabsize=4,
    captionpos=b,
}

\title{Deep Learning and Machine Learning: Advancing Big Data Analytics and Management}

\date{}

\begin{document}

\begin{titlepage}
    \centering
    \includegraphics[width=1.0\textwidth]{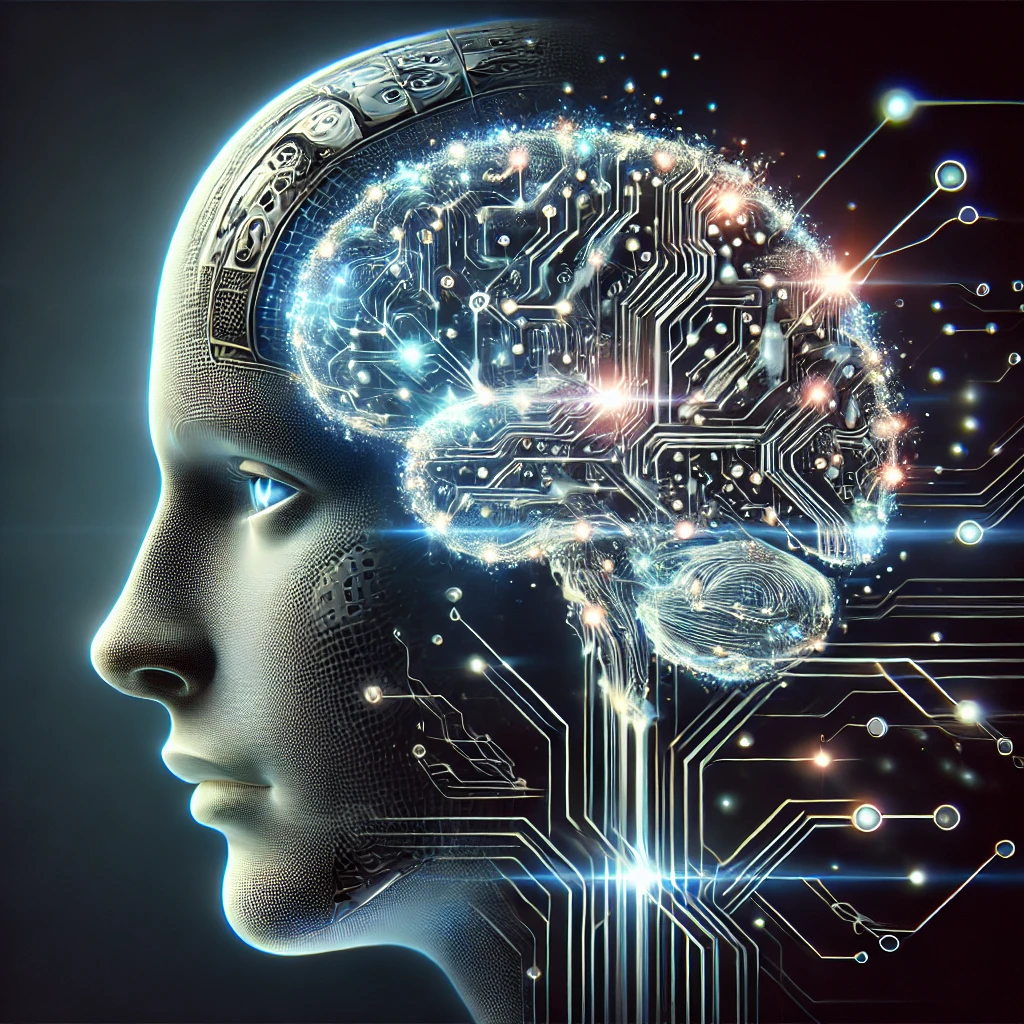}
    \vfill
    {\LARGE \textbf{Deep Learning and Machine Learning: Advancing Big Data Analytics and Management}} \\
    \vspace{1cm}
    
    {\Large
    Benji Peng\textsuperscript{1*}, Xuanhe Pan\textsuperscript{2*}, Yizhu Wen\textsuperscript{3*}, Ziqian Bi\textsuperscript{4*}, Keyu Chen\textsuperscript{5}, Ming Li\textsuperscript{5}, Ming Liu\textsuperscript{6}, Qian Niu\textsuperscript{7}, Junyu Liu\textsuperscript{7}, Jinlang Wang\textsuperscript{2}, Sen Zhang\textsuperscript{8}, Jiawei Xu\textsuperscript{6}, Xinyuan Song\textsuperscript{10}, Zekun Jiang\textsuperscript{11}, Tianyang Wang\textsuperscript{12}, Pohsun Feng\textsuperscript{9$\dagger$} \\
    }
    
    \vspace{0.5cm}
    {\small
    \textsuperscript{1}AppCubic, \textsuperscript{2}University of Wisconsin-Madison, \textsuperscript{3}Department of Electrical and Computer Engineering, University of Hawaii at Manoa, \textsuperscript{4}Indiana University, \textsuperscript{5}Georgia Institute of Technology, \textsuperscript{6}Purdue University, \textsuperscript{7}Kyoto University, \textsuperscript{8}Rutgers University, \textsuperscript{9}National Taiwan Normal University, \textsuperscript{10}Emory University, \textsuperscript{11}Sichuan University, \textsuperscript{12}Xi’an Jiaotong-Liverpool University \\
    *Equal contribution \\
    $\dagger$Corresponding author, 41075018h@ntnu.edu.tw
    }
    
    \vspace{1.5cm}
    \vfill
    \date{}
\end{titlepage}


\tableofcontents  
\cleardoublepage

\section{Introduction}

In the past decade, the rapid development of Artificial Intelligence (AI), Machine Learning (ML), and Deep Learning (DL) technologies has fundamentally transformed numerous industries, driving global innovation and productivity to new heights~\cite{The_transformative_AI}. Data has become the core asset of modern enterprises and organizations~\cite{Valuing_Data_as_an_Asset}, and the ability to extract valuable insights and knowledge from vast amounts of data has become a key competitive advantage~\cite{Competitive_advantage}. With the widespread use of big data, machine learning, and deep learning have become essential technologies for automating processes, optimizing decisions, and improving efficiency~\cite{ML_DL_Reviwe, Sarker2021, ML_DL_app_in_health}.

Although deep learning and machine learning have tremendous potential, many people find these technologies complex and difficult to understand. This book aims to simplify these complex mathematical formulations, provide intuitive visualizations, and provide comprehensive explanations of core concepts of deep learning and machine learning, allowing readers to easily grasp these technologies and understand their wide-ranging applications~\cite{computers12050091}.

\subsection{Simplifying Deep Learning: From Complex to Simple}

Deep learning (DL) technology based on artificial neural networks typically involves complex mathematical computations such as linear algebra, calculus, and probability theory. These technical foundations have contributed to the success of deep learning, but also present a barrier for many beginners.

One of the core goals of this book is to simplify these complex concepts. We will use simplified mathematical derivations, practical example codes, and graphical demonstrations to make the key concepts of deep learning easier to understand. Using visualization techniques, readers will be able to intuitively grasp the workings of forward propagation and backpropagation in neural networks, how convolutional neural networks (CNNs)~\cite{oshea2015introductionconvolutionalneuralnetworks} operate, and how model parameters are optimized using gradient descent.

In addition, this book will provide practical projects and case studies to help readers understand how these technologies can be applied in real-world scenarios. This approach will not only help technical professionals master deep learning but will also provide business managers with insights into how to leverage these technologies in the era of big data.

\subsection{Applications of Machine Learning and Deep Learning}

Machine learning and deep learning have become the foundation of modern technology systems, driving innovation and efficiency across industries. In the following, we will introduce several classic models and technologies that have played an irreplaceable role in various fields. The models and technologies that we introduce are not limited to these, but these represent some of the most widely applied and impactful models in the industry.

\subsubsection{Transformer}
\begin{figure}
    \centering
    \includegraphics[width=0.4\linewidth]{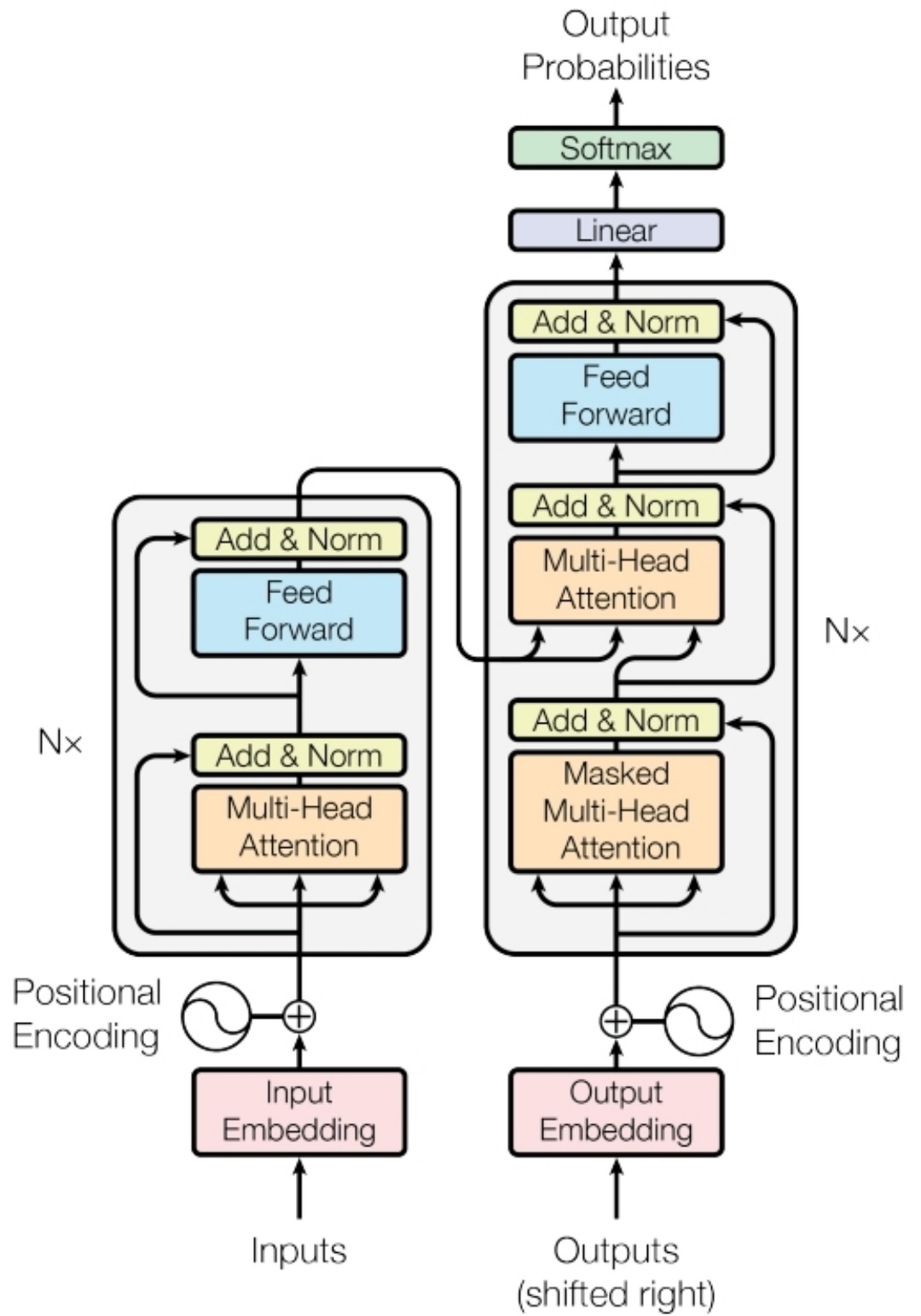}
    \caption{Transformer Model Architecture}
    \label{fig:transformer}
\end{figure}
The Transformer is a deep learning architecture~\cite{vaswani2023attentionneed} based on self-attention mechanisms (Figure~\ref{fig:transformer} from~\cite{vaswani2023attentionneed}), originally designed for Natural Language Processing (NLP) but now extended to image processing and speech recognition. Unlike traditional sequence models such as RNNs~\cite{schmidt2019recurrentneuralnetworksrnns} and LSTMs~\cite{LSTM}, the Transformer can process input sequences in parallel, making it more efficient for handling long texts or complex sequences. The Transformer architecture~\cite{vaswani2023attentionneed} is built on models like GPT~\cite{radford2018improving} and BERT~\cite{devlin2019bertpretrainingdeepbidirectional}, used widely in NLP applications such as language translation, text generation, and question answering~\cite{wolf-etal-2020-transformers}.

\subsubsection{ChatGPT and Generative Pre-trained Models (GPT)}
\begin{center}
\includegraphics[width=1\textwidth]{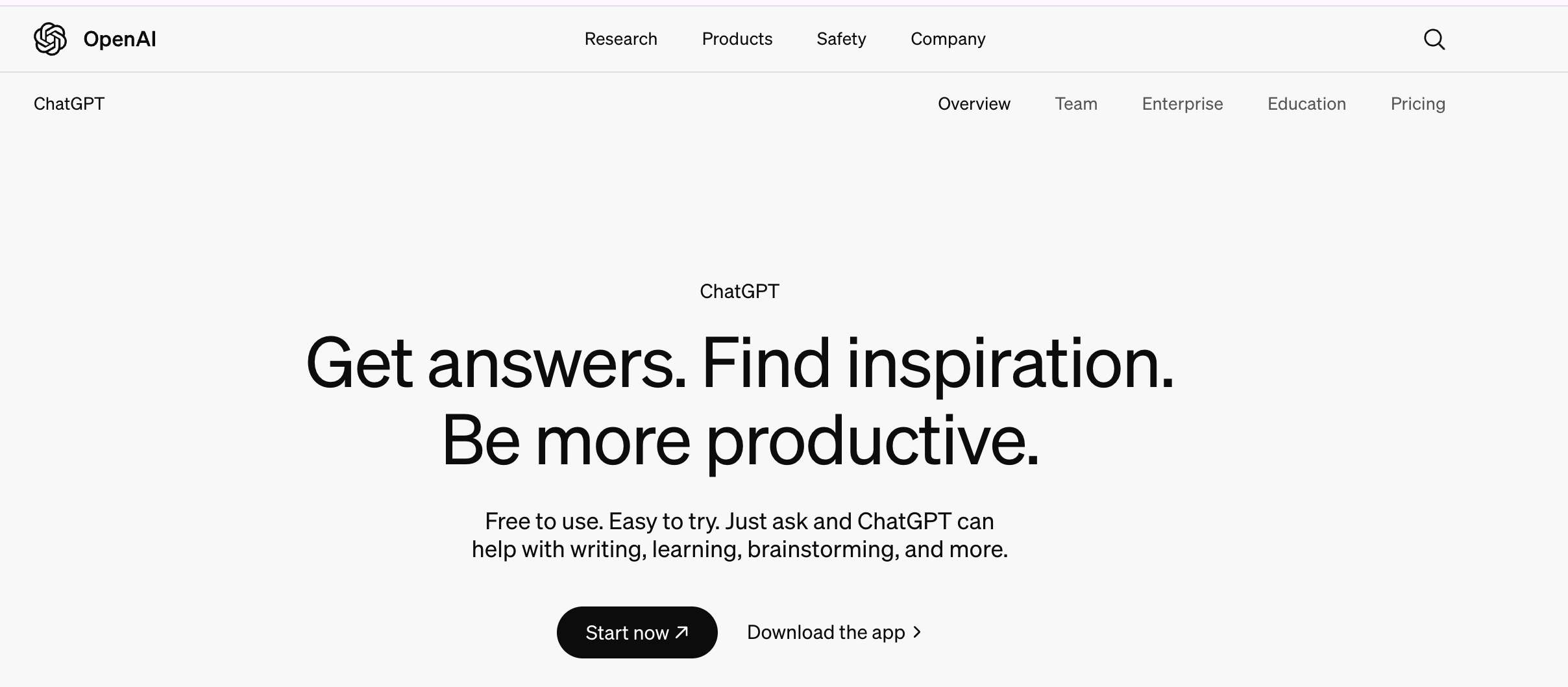}
\end{center}
ChatGPT is based on OpenAI’s Generative Pre-trained Transformer (GPT)~\cite{openai2024gpt4technicalreport}, trained on vast amounts of text data, capable of generating coherent, natural conversations. It has been widely applied in chatbots, text generation, and virtual assistants~\cite{Shahriar_2023}. The GPT series of models has become an essential technology in natural language processing, pushing the boundaries of language models in content generation, automated customer service, and writing assistance.

\subsubsection{XGBoost (eXtreme Gradient Boosting)}
\begin{center}
\includegraphics[width=1\textwidth]{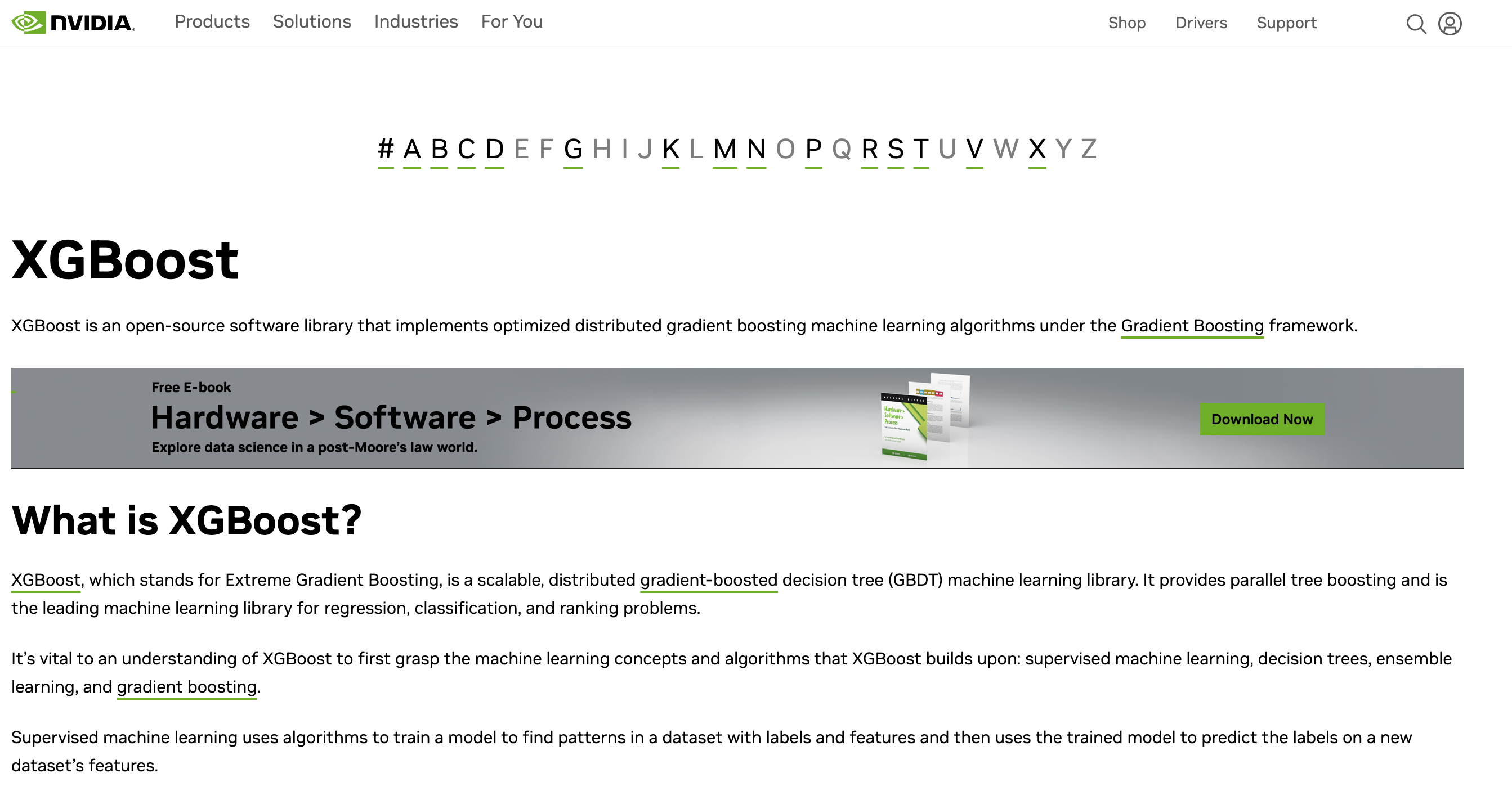}
\end{center}
XGBoost~\cite{Chen_2016} is a decision-tree-based ensemble machine learning model that is particularly suited for handling structured data. Its efficiency and accuracy have made it a popular tool, especially in Kaggle competitions. A wide range of industries, including finance, healthcare, and marketing, use XGBoost to analyze structured data, making it indispensable~\cite{Bain_etal, NIAZKAR2024105971}.

\subsubsection{AlphaGo Zero}
\begin{figure}
    \centering
    \includegraphics[width=0.5\linewidth]{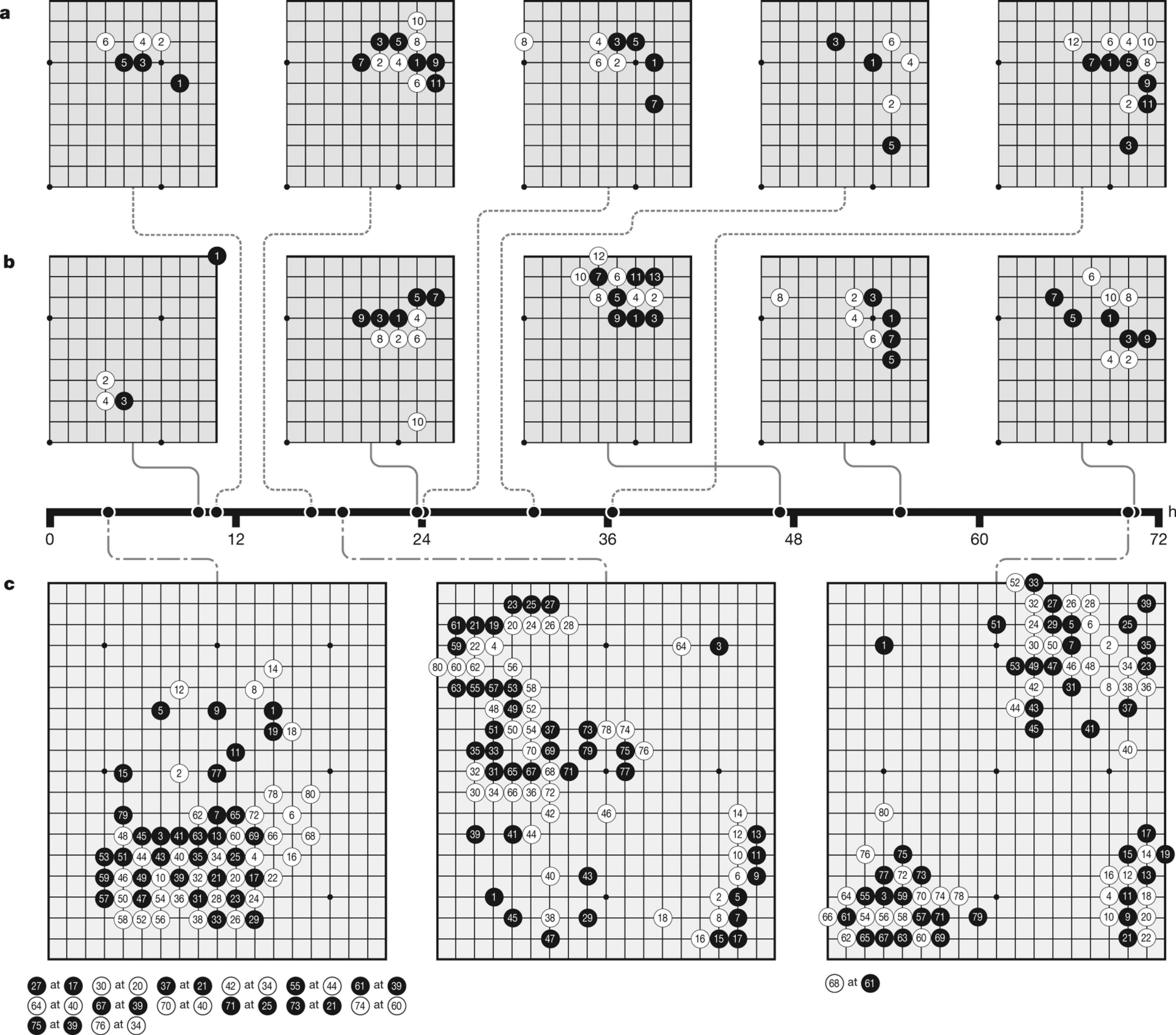}
    \caption{AlphaGo Zero}
    \label{fig:alpha_go}
\end{figure}
AlphaGo Zero~\cite{Silver2017} is an AI system based on deep reinforcement learning, which learns complex Go strategies through self-play. Unlike earlier AlphaGo versions, AlphaGo Zero (Figure~\ref{fig:alpha_go} from~\cite{Silver2017}) continuously improves through self-competition, eventually surpassing humans. It has applications in autonomous driving, robotic control, and more, and it illustrates the immense potential of deep reinforcement learning for complex decision-making and automation.

\subsubsection{ResNet (Residual Networks)}
\begin{figure}
    \centering
    \includegraphics[width=0.4\linewidth]{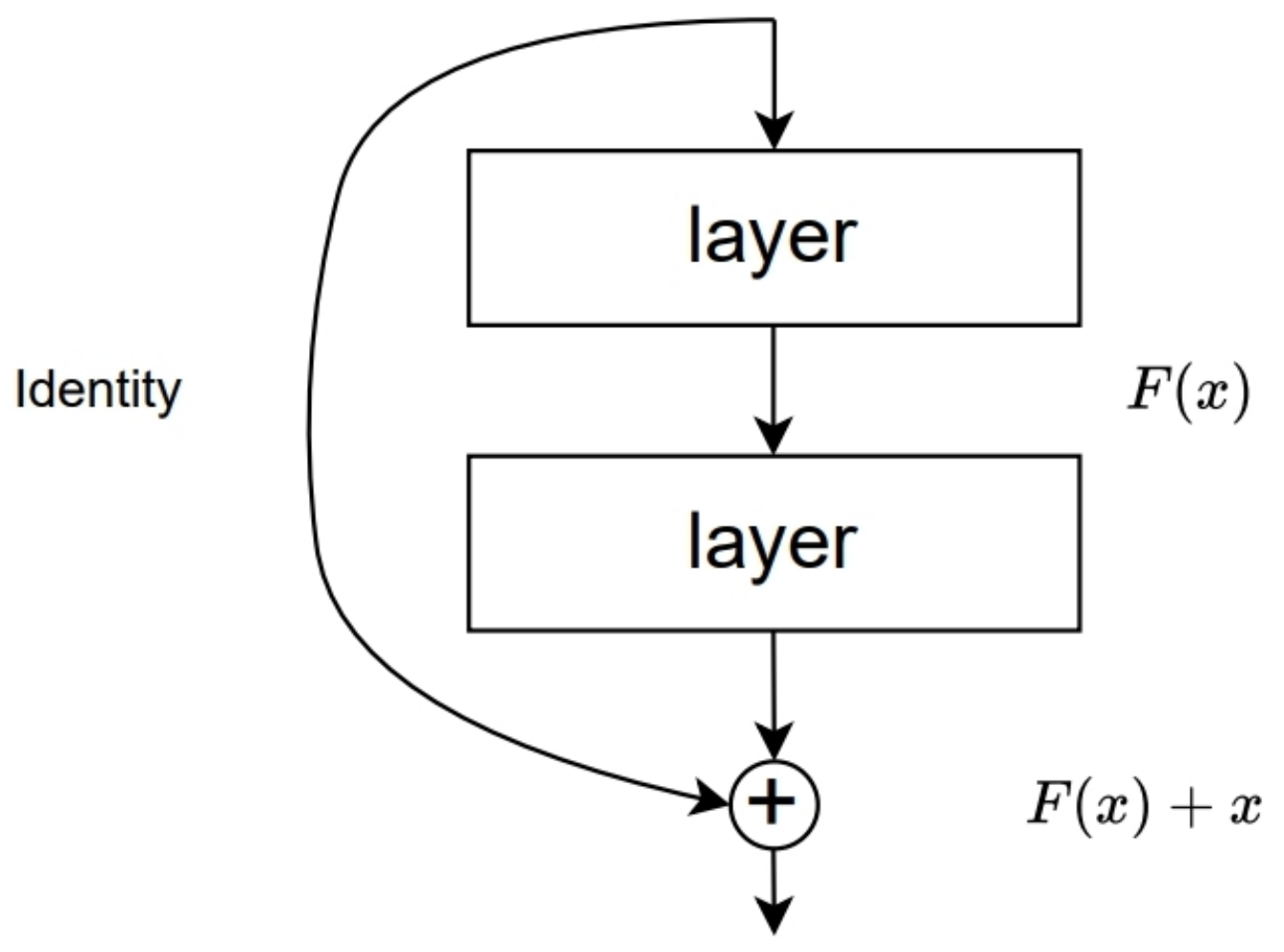}
    \caption{Residual Networks Architecture}
    \label{fig:resnet}
\end{figure}
ResNet~\cite{he2015deepresiduallearningimage} is a specific convolutional neural network model, that addresses the vanishing gradient problem in deep networks through residual connections (Figure~\ref{fig:resnet}), enabling much deeper networks to be trained. It is widely applied in medical image analysis, autonomous driving, and visual processing in autonomous vehicles. ResNet is capable of analyzing images, detecting objects, and classifying them.

\subsubsection{BERT (Bidirectional Encoder Representations from Transformers)}
\begin{figure}
    \centering
    \includegraphics[width=0.4\linewidth]{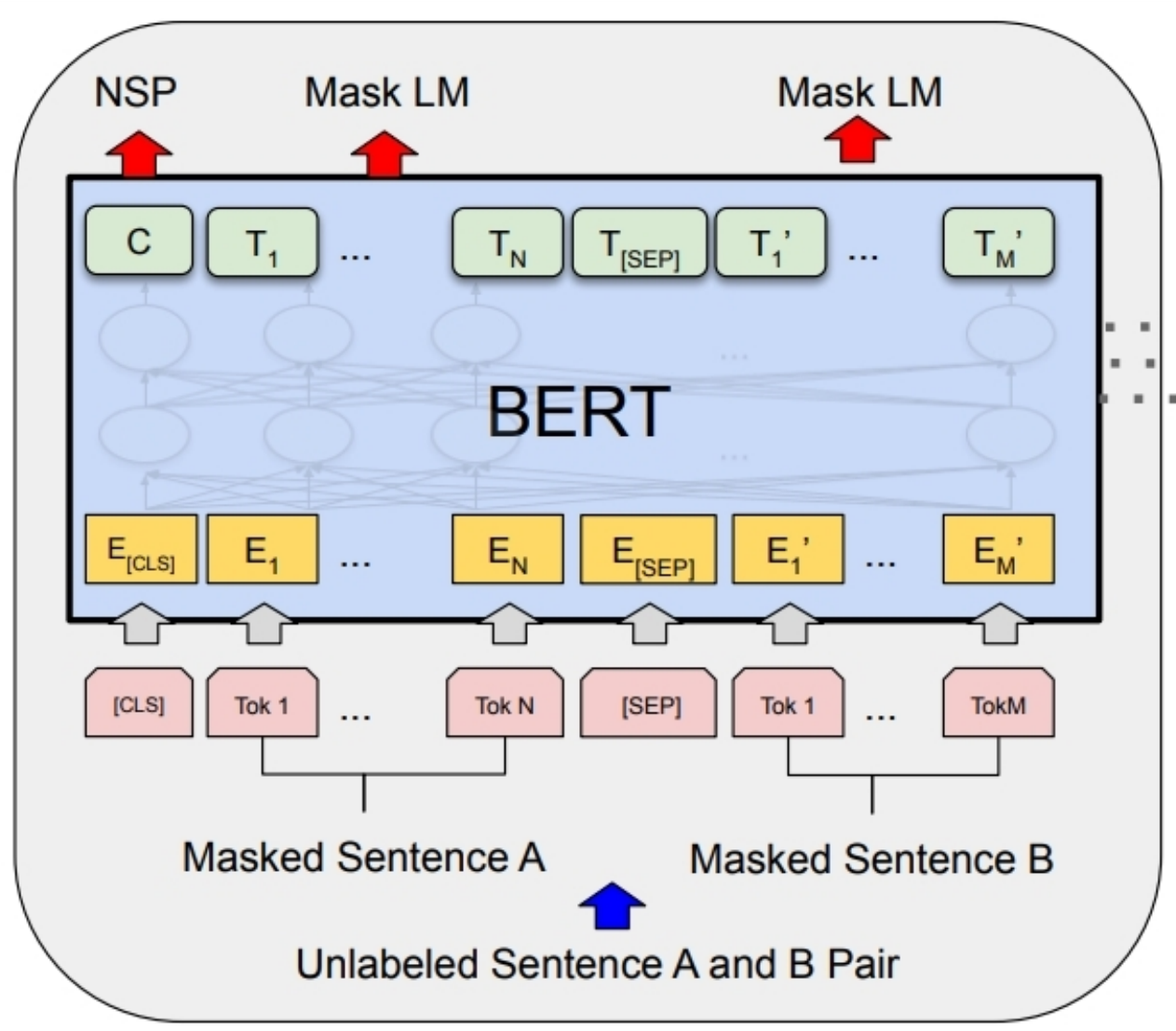}
    \caption{BERT Pre-training Architecture}
    \label{fig:bert}
\end{figure}
BERT, developed by Google~\cite{devlin-etal-2019-bert}, is a bidirectional encoder model that excels at understanding the contextual semantics of words in text (Figure~\ref{fig:bert} from~\cite{devlin-etal-2019-bert}). In tasks like sentiment analysis, question answering, and text classification, BERT has become a benchmark technology. Unlike GPT’s unidirectional generative model, BERT’s bidirectional encoding captures richer contextual information, providing powerful support for various NLP tasks.

\subsubsection{YOLO (You Only Look Once)}
\begin{figure}
    \centering
    \includegraphics[width=0.8\linewidth]{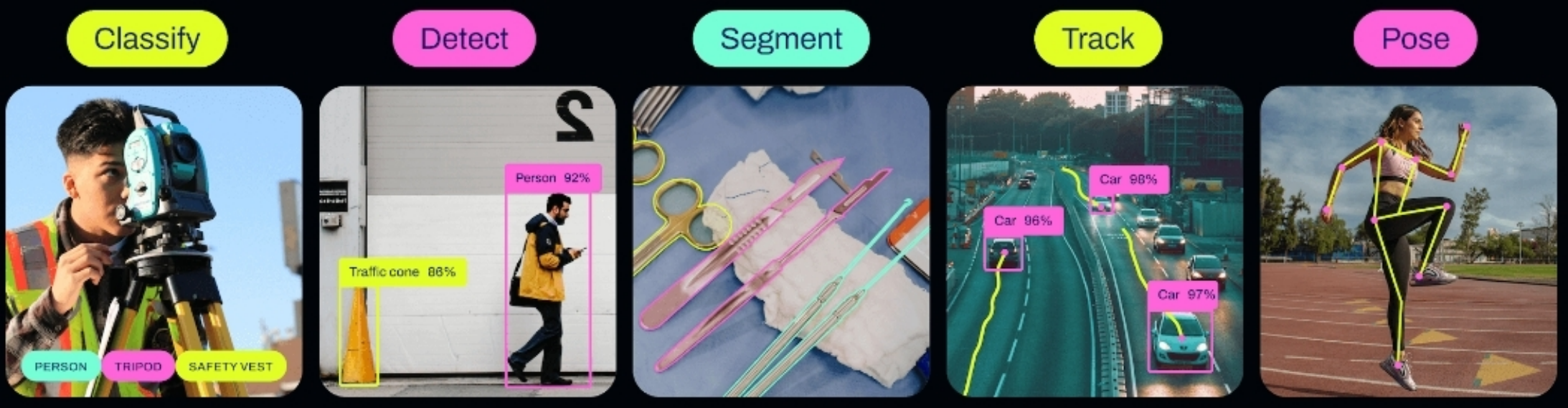}
    \caption{YOLO Models}
    \label{fig:yolo}
\end{figure}
YOLO~\cite{redmon2016lookonceunifiedrealtime, Terven_2023} is a real-time object detection model that can locate and classify multiple objects in a single image (Figure~\ref{fig:yolo} from~\cite{redmon2016lookonceunifiedrealtime}). Due to its outstanding speed and accuracy, YOLO is widely used in real-time detection applications such as autonomous driving, security surveillance, smart cities, and industrial manufacturing.

\subsubsection{AlphaFold}
\begin{figure}
    \centering
    \includegraphics[width=0.4\linewidth]{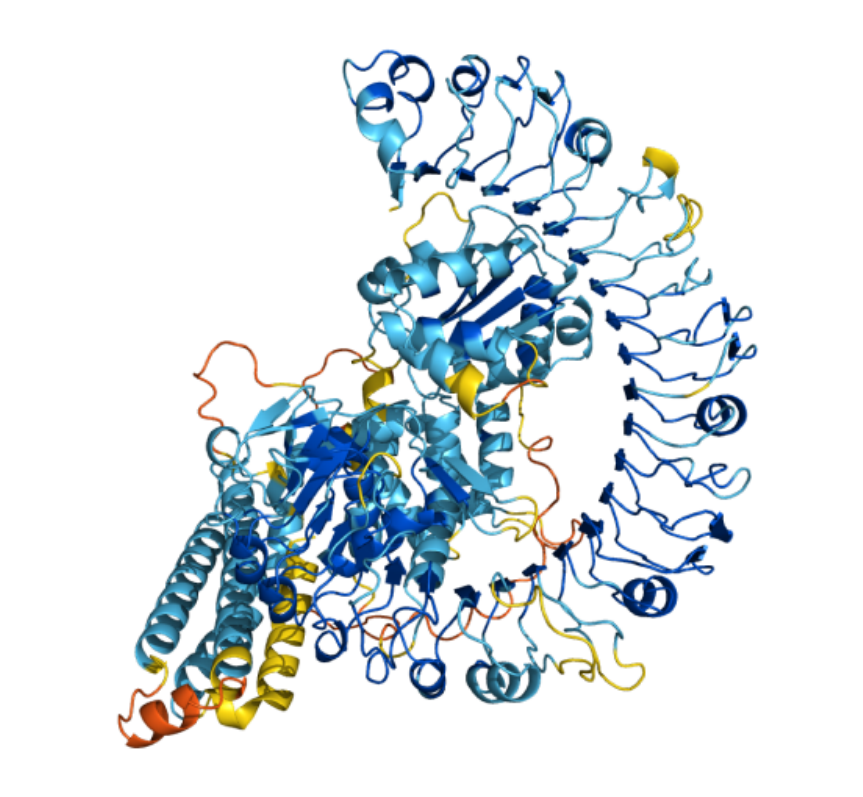}
    \caption{Q8W3K0: A potential plant disease resistance protein. Mean pLDDT 82.24.}
    \label{fig:alphafold}
\end{figure}
AlphaFold~\cite{Jumper2021} is a deep learning model developed by Google DeepMind that predicts the three-dimensional structure of proteins (Figure~\ref{fig:alphafold} from~\cite{alphafold_image}). This model solves a major challenge in biology, significantly accelerating drug discovery and disease research, and bringing revolutionary changes to life sciences and biotechnology.

\subsubsection{GAN (Generative Adversarial Networks)}
\begin{figure}
    \centering
    \includegraphics[width=0.7\linewidth]{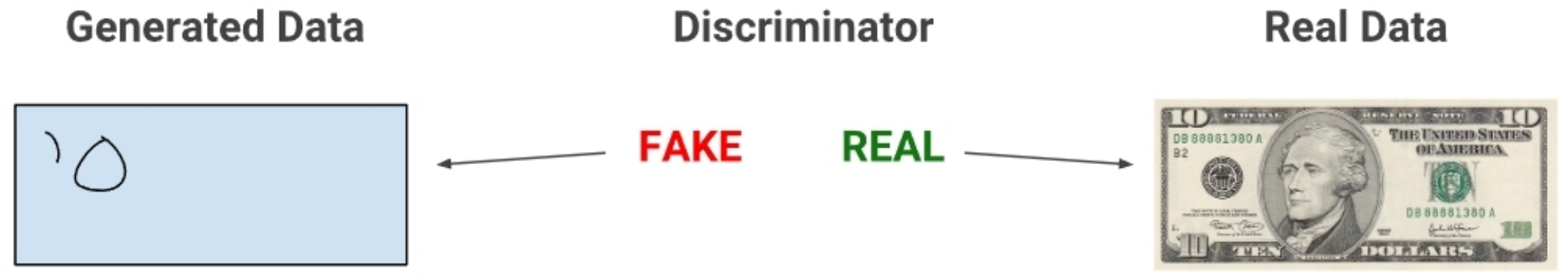}
    \caption{GAN Mechanism}
    \label{fig:GAN}
\end{figure}
Generative Adversarial Networks (GANs)~\cite{goodfellow2014generativeadversarialnetworks, Creswell_2018} consists of a generator and a discriminator (Figure\ref{fig:GAN} from~\cite{google_gan_structure}), where the generator creates data and the discriminator distinguishes between real and fake data. GANs are widely used in image generation, style transfer, image restoration, and artistic creation, and can generate realistic images, videos, and audio, greatly expanding the applications of deep learning in content creation.

\subsubsection{Reinforcement Learning}
\begin{figure}
    \centering
    \includegraphics[width=0.6\linewidth]{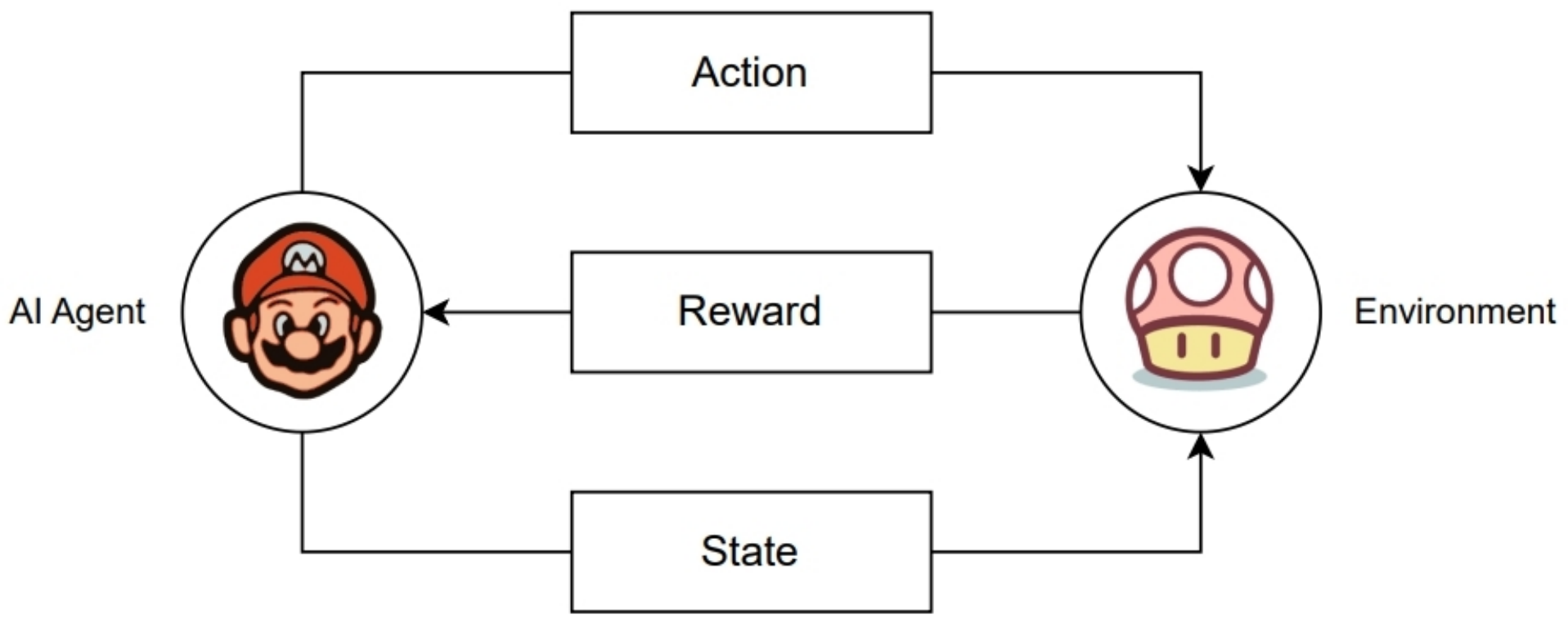}
    \caption{Reinforcement Learning}
    \label{fig:RL}
\end{figure}
Reinforcement~\cite{mnih2013playingatarideepreinforcement, hausknecht2017deeprecurrentqlearningpartially, wang2016duelingnetworkarchitecturesdeep, vanhasselt2015deepreinforcementlearningdouble} learning is a machine learning method focused on learning optimal behaviors through interactions with the environment (Figure~\ref{fig:RL}). Through trial and error, reinforcement learning optimizes decision-making. It has shown great potential in fields such as autonomous driving, robotic control, financial market optimization, and resource management.

\subsection{Pre-trained Models: Enhancing Model Performance and Accuracy}

\subsubsection{The Concept of Pre-trained Models}
Pre-trained models \cite{chen2024deeplearningmachinelearning} are trained on large datasets before being applied to specific tasks. In the development of deep learning, pre-trained models have significantly reduced training time while improving performance and accuracy on tasks with limited data.

\subsubsection{Common Pre-trained Models}
\begin{itemize}
    \item \textbf{BERT (Bidirectional Encoder Representations from Transformers):} Used in natural language processing, BERT is a bidirectional model that considers the context of a sentence from both directions to understand semantics. Pre-trained on massive text datasets, BERT is well-suited for various language tasks, such as question-answering systems and language translation.
    
    \item \textbf{GPT (Generative Pre-trained Transformer):} GPT models are generative models used to produce text and understand context, widely applied in conversation generation and content creation. ChatGPT is one such application of this technology.
    
    \item \textbf{ResNet (Residual Networks):} ResNet is a pre-trained model used for image recognition. By introducing residual structures, it solves the degradation problem in deep neural networks. After being pre-trained on large-scale image datasets, ResNet can be fine-tuned for specific image recognition tasks.
\end{itemize}

\subsubsection{Advantages of Pre-trained Models}
\begin{itemize}
    \item \textbf{Improved Learning Efficiency:} Pre-trained models have already learned general features from large datasets, allowing them to achieve high performance on new tasks with minimal fine-tuning.
    
    \item \textbf{Reduced Computational Cost:} Compared to training models from scratch, using pre-trained models saves significant computational resources and time.
    
    \item \textbf{Wide Range of Applications:} Pre-trained models are widely applied across various fields, from natural language processing and image recognition to medical image analysis and autonomous driving.
\end{itemize}

\subsubsection{Future Developments of Pre-trained Models}
With the continuous expansion of dataset sizes and improvements in computational power, pre-trained models will play a larger role in more fields, and more efficient architectures are likely to emerge, such as models capable of handling multi-modal data.

\subsubsection{How to Use Pre-trained Models in Applications}
In practice, the main methods for using pre-trained models include:
\begin{enumerate}
    \item \textbf{Transfer Learning:} Fine-tuning pre-trained weights for specific tasks.
    \item \textbf{Feature Extraction:} Using intermediate layers of pre-trained models to extract features from data and training custom models with these features.
\end{enumerate}

\subsection{Big Data Management and Processing Technologies}

The success of deep learning and machine learning is inseparable from the support of vast amounts of data. However, effectively storing, managing, and processing this data presents a significant challenge. This book will introduce the key technologies of modern data management, helping readers understand how these tools provide the necessary support for machine learning and deep learning models.

\subsubsection{SQL and NoSQL Databases}
\begin{figure}
    \centering
    \includegraphics[width=0.6\linewidth]{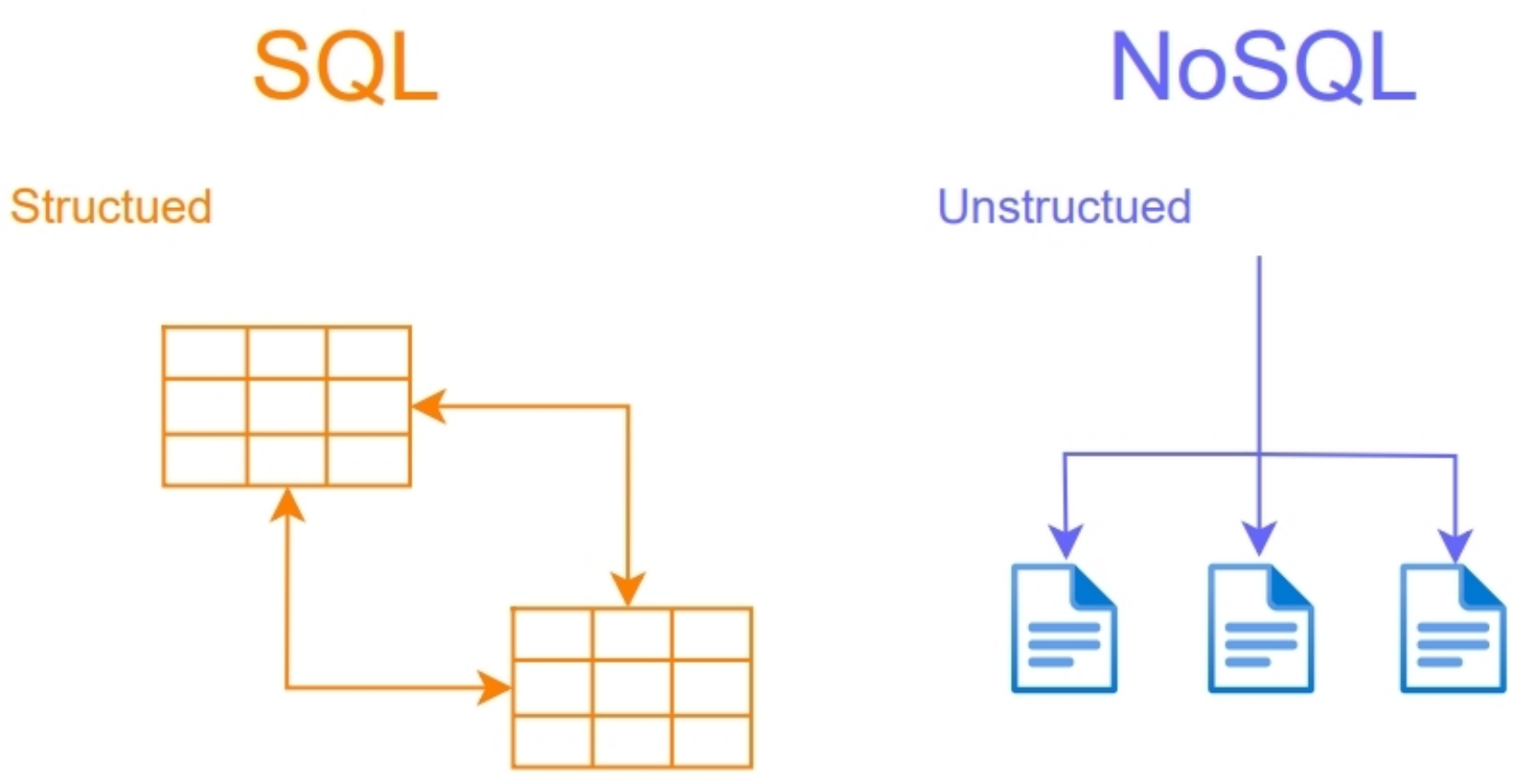}
    \caption{SQL and NoSQL Database}
    \label{fig:database}
\end{figure}
\paragraph{SQL Databases} Traditional relational databases (such as MySQL, PostgreSQL) are well-suited for handling structured data~\cite{Ferreira_2017, Silva2016} (see the left side of Figure~\ref{fig:database}), especially in scenarios that require transactional consistency and strict data management. We will demonstrate how to use SQL databases for data storage and queries and their applications in industries like finance and retail.

\paragraph{NoSQL Databases} To cope with the explosive growth of unstructured data, NoSQL databases~\cite{Gupta2017} (such as MongoDB, Cassandra) provide flexible solutions for storing and processing large-scale data (see the right side of Figure~\ref{fig:database}). NoSQL databases are widely applied in scenarios like IoT and social media data analysis.

\subsubsection{Graph Databases and Relational Data Analysis}

\paragraph{Graph Databases} Graph databases~\cite{tian2022worldgraphdatabasesindustry} (such as Neo4j) efficiently store and query complex relationship networks among nodes and edges, making them especially suitable for use in social networks, recommendation systems, and fraud detection. We will demonstrate how graph databases help process and analyze complex relational data~\cite{Gao2023}.

\subsubsection{Cluster Computing and Distributed Systems}

\paragraph{Apache Hadoop} Hadoop~\cite{hadoop} is a distributed storage and computing framework that efficiently processes large-scale data through the MapReduce model, suitable for batch processing tasks. We will explain how to use Hadoop to handle unstructured data and demonstrate its applications in e-commerce and finance.

\paragraph{Apache Spark} Spark~\cite{spark} offers a faster in-memory computing framework than Hadoop, supporting real-time data stream processing and complex data analysis tasks. This book will show how to leverage Spark’s advantages in real-time data analysis, machine learning, and large-scale data processing.

\subsection{Future Competitiveness: Essential Skills in Deep Learning and Data Management}

With the widespread adoption of AI technologies, deep learning, and big data management are no longer tools exclusively for data scientists but have become core skills required for professionals across industries. Whether you are a software developer, enterprise manager, or researcher, mastering these technologies will not only help you face future challenges but will also add irreplaceable value to your work~\cite{ML_4.0}.

Numerous industries have already been dramatically transformed by the rapid development of AI technologies. Deep learning and big data management skills are becoming increasingly important for technical roles. Deep learning is necessary for developing AI-driven applications, automating business processes, and optimizing system performance. Through big data and artificial intelligence, enterprise managers can optimize decision-making processes to improve efficiency and drive digital transformation~\cite{Sarker2021, Choudhary2022}.

However, the risk of not mastering these skills is evident. As automation systems and intelligent decision-making systems become more prevalent, many traditional roles will be replaced. Only those who can fully utilize deep learning and big data technologies will remain competitive in the evolving job market.

Through learning this book, readers will gain the key technical skills in these two critical fields and develop the ability to apply them practically, positioning themselves favorably for future career success.

\subsection{Goals of This Book}

The goal of this book is to simplify complex technologies, provide intuitive visualizations, and offer real-world case studies to help readers fully grasp the core concepts of deep learning and machine learning while gaining a deep understanding of the key technologies in big data management and analysis. Whether readers are beginners, technical experts, or business managers, this book will offer systematic content to benefit everyone at various levels.

\subsubsection{Simplifying Complex Technologies to Lower the Learning Curve}

Deep learning and machine learning typically involve complex mathematical derivations and algorithmic implementations, such as linear algebra, calculus, and probability theory. This book aims to simplify these complex topics, helping readers understand their core principles. By using simplified mathematical formulas, intuitive explanations, and accessible analysis, readers will be able to easily grasp key concepts without being overwhelmed by technical details.

\subsubsection{Comprehensive Coverage of Applications and Classic Models}

To help readers fully understand the practical applications of deep learning and machine learning, this book will cover a wide range of scenarios, from natural language processing (NLP) to computer vision, reinforcement learning, and generative adversarial networks (GANs)~\cite{goodfellow2014generativeadversarialnetworks}. We will provide detailed explanations of classic models such as Transformer, ChatGPT~\cite{OpenAI2023}, AlphaGo Zero~\cite{Silver2017}, BERT~\cite{devlin-etal-2019-bert}, ResNet~\cite{he2015deepresiduallearningimage}, and YOLO~\cite{redmon2016lookonceunifiedrealtime}, and show how these models work in practice, covering real-world applications in industries such as finance, healthcare, autonomous driving, and security.

These models will be presented in order of their impact on the industry, helping readers prioritize the most important technologies. Through rich case studies and real-world application examples, readers will gain a clear understanding of how these technologies are transforming workflows and decision-making processes in various industries.

\subsubsection{Core Data Management Technologies and Tools}

The efficient operation of these models is dependent on modern data management technologies, which go beyond deep learning and machine learning. This book will explain in detail how to use SQL and NoSQL databases to store and manage large-scale data, how to handle complex relational data with graph databases (such as Neo4j), and how to process vast amounts of data using cluster computing frameworks (such as Hadoop~\cite{hadoop} and Spark~\cite{spark}).

Through real-world case studies, we will demonstrate how to integrate these data management technologies with deep learning to build scalable solutions, helping readers master the entire process of big data analysis, from data collection, storage, and cleaning to modeling and prediction.

\subsubsection{Visualization and Practice-Driven Learning}

By using visualization tools throughout the book, readers will gain a better understanding of how neural network layers function, how CNNs process images, and how model parameters are optimized using gradient descent. This complex algorithm can be intuitively understood by using diagrams and data visualizations.

In addition, this book will provide practical projects and code examples to demonstrate how theory can be applied in practice. Whether it’s training machine learning models or building big data platforms, readers will be guided step by step through the entire process, easily mastering the transition from theory to application. Each key concept will be accompanied by concrete examples, helping readers apply what they have learned in real-world scenarios.

\subsubsection{A Comprehensive and Systematic Learning Path}

The content of this book spans from foundational theory to cutting-edge technologies, striving for comprehensive coverage. Readers will follow a systematic learning path, starting from the basic theories of machine learning and deep learning, and gradually progressing to big data management, model optimization, and cluster computing. Whether you are a beginner or an experienced professional, you will find content tailored to your needs and gradually master the entire technological ecosystem.

This book provides readers with a comprehensive understanding of the big data management and analysis ecosystem, such as deep learning and machine learning. Whether you are looking to enhance your current skills or prepare for future career growth, this book will provide you with a complete set of knowledge and skills to help you stay competitive in an AI-driven era.

\part{Handy Appetizer}

\chapter{Getting Started with Python}

\section{Why Python?}

Python is one of the easiest programming languages to learn. It’s known for its clean, readable syntax—writing Python is almost like writing English. Whether you're interested in data science, building websites, or creating machine learning models, Python~\cite{python} is the perfect starting point.

\subsection{What is Python Used For?}

\begin{itemize}
    \item \textbf{Data Science:} Analyze large datasets with libraries like \texttt{pandas} and \texttt{numpy}.
    \item \textbf{Web Development:} Build web applications using frameworks like \texttt{Flask} and \texttt{Django}.
    \item \textbf{Machine Learning:} Create predictive models with libraries such as \texttt{scikit-learn} and \texttt{TensorFlow}.
    \item \textbf{Automation:} Automate repetitive tasks like renaming files or scraping data from websites.
\end{itemize}

\section{What is an IDE?}

An Integrated Development Environment (IDE) is a tool that helps you write, run, and test code. Python comes with a built-in IDE called \textbf{IDLE}. It’s simple, easy to use, and perfect for beginners. While there are other IDEs available, such as VS Code and PyCharm, we will focus on IDLE for now.

In future chapters, we’ll explore other IDEs and tools that might be more suitable for larger projects.

\section{Setting Up Python}

Let’s get Python up and running on your computer. These instructions are for complete beginners, so follow along step by step.

\subsection{Step 1: Download Python}

\begin{enumerate}
    \item Open your web browser and go to the official Python website: \href{https://www.python.org/downloads/}{python.org/downloads}.
    \item Depending on your operating system, choose the appropriate version to download:
        \begin{itemize}
            \item For \textbf{Windows}, click the large button that says \textbf{Download Python 3.x.x} (e.g., Python 3.12.6) as shown in the image below.
            
            \begin{center}
            \includegraphics[width=0.8\textwidth]{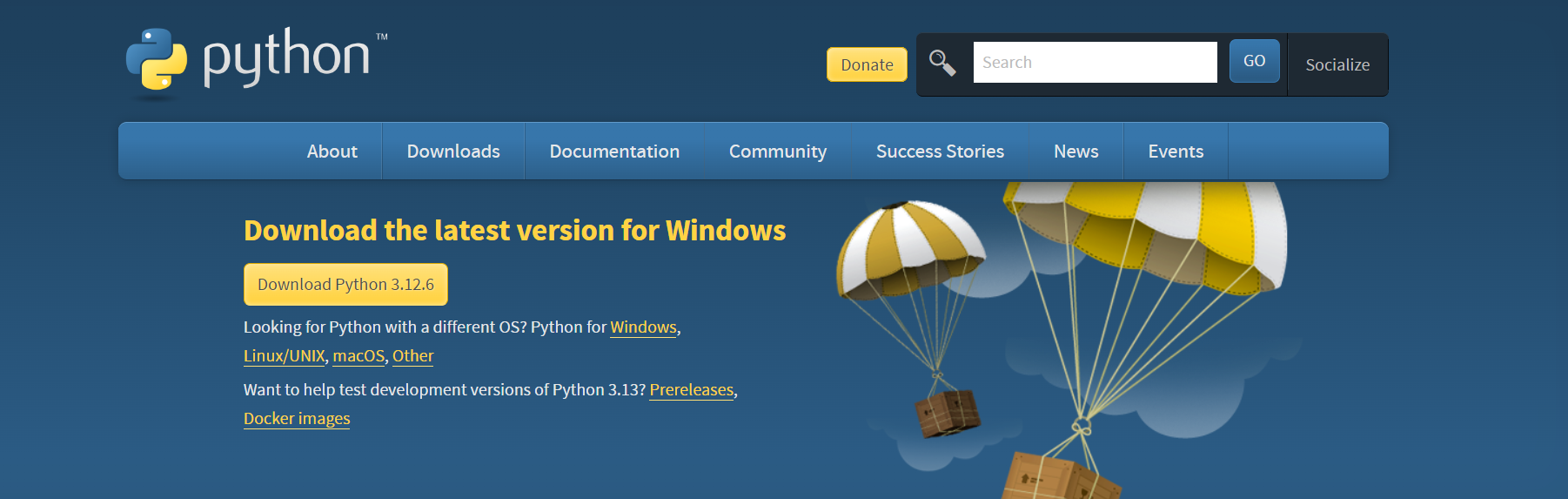}
            \end{center}
            
            \item For \textbf{MacOS}, follow this link: \href{https://www.python.org/downloads/macos/}{macOS version}. Choose the latest stable release and download the universal installer, as shown in the two images below.

            \begin{center}
            \includegraphics[width=0.8\textwidth]{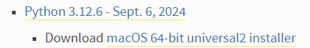}
            \end{center}
        \end{itemize}
    \item Make sure you download the version marked as Python 3.x (e.g., Python 3.11).
\end{enumerate}

\subsection{Step 2: Install Python}

\textbf{For Windows:}

\begin{enumerate}
    \item Once the download is complete, run the installer by double-clicking on it.
    \item Before proceeding, check the box that says \textbf{Add Python to PATH}. This is important!
    \item Click \textbf{Install Now} and wait for the installation to complete. You’ll see a message that says \textbf{Setup was successful}.
\end{enumerate}

\textbf{For MacOS:}

\begin{enumerate}
    \item Open the downloaded \texttt{.pkg} file.
    \item Follow the prompts by clicking \textbf{Continue} and \textbf{Agree}.
    \item Once the installation is finished, open \textbf{Terminal} and type the following to verify Python is installed:
    \begin{lstlisting}[style=cmd]
    python3 --version
    \end{lstlisting}
\end{enumerate}

\textbf{For Linux:}

\begin{enumerate}
    \item Many Linux distributions come with Python pre-installed. You can check by opening \textbf{Terminal} and typing:
    \begin{lstlisting}[style=cmd]
    python3 --version
    \end{lstlisting}
    \item If Python is not installed, use the following command:
    \begin{lstlisting}[style=cmd]
    sudo apt-get install python3
    \end{lstlisting}
\end{enumerate}

\section{Step 3: Using IDLE - Your First Python IDE}

Python comes with its own built-in IDE called \textbf{IDLE} (Integrated Development and Learning Environment). You don’t need to install anything extra! Once Python is installed, you can start using IDLE right away to write and run Python code.

\begin{itemize}
    \item \textbf{Windows and Mac:} After installing Python, search for \texttt{IDLE} in your start menu (or Finder on Mac) and open it. The image below shows how it looks on a Windows machine:
    
    \begin{center}
    \includegraphics[width=0.4\textwidth]{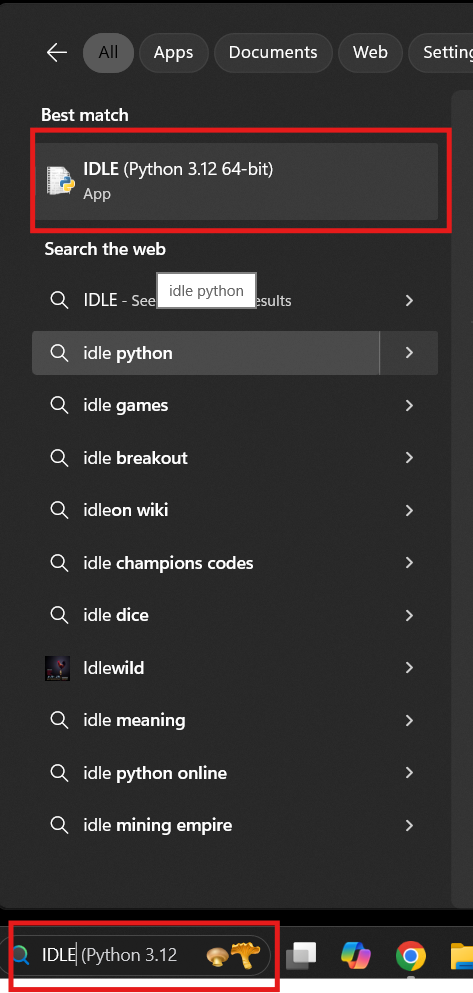}
    \end{center}
    
    \item \textbf{Linux:} Open Terminal and type \texttt{idle3}.
\end{itemize}

In IDLE, you can type Python code and run it instantly. It’s a perfect way to get familiar with Python, especially if you’ve never programmed before.

\section{Your First Python Program}

Now that you have Python installed and IDLE open, let’s write your first Python program! This will be the simplest program you can imagine, but it’s a great way to see Python in action.

\subsection{Step-by-Step Instructions}

\begin{enumerate}
    \item Open \textbf{IDLE}.
    \item In the IDLE window, type the following code:
    \begin{lstlisting}[style=python]
    print("Hello, World!")
    \end{lstlisting}
    \item Press \texttt{Enter}.
\end{enumerate}

You should see the following output:

\begin{center}
\includegraphics[width=1\textwidth]{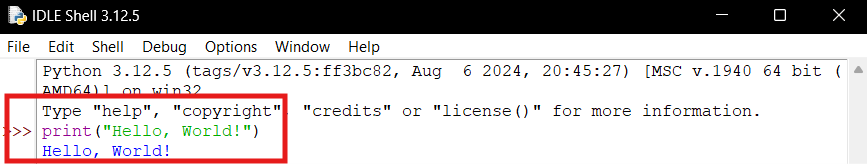}
\end{center}

\begin{lstlisting}[style=cmd]
Hello, World!
\end{lstlisting}

That’s it! You’ve just written and run your first Python program. This simple code tells Python to print the text \texttt{"Hello, World!"} on the screen.

\section{Quick Recap}

\begin{itemize}
    \item Python is a beginner-friendly language that’s easy to learn and powerful to use.
    \item You installed Python and used its built-in IDE, \textbf{IDLE}, to write and run Python code.
    \item You successfully wrote your first Python program!
\end{itemize}

In the next chapter, we’ll dive into real-world Python applications. You'll see how Python can be used for data analysis, automation, and more.

\chapter{Understanding Python Packages and \texttt{pip}}

\section{What are Python Packages?}

Python packages are collections of modules (i.e., pre-written Python code) that provide functions and tools for specific tasks. For example, if you need to work with data visualization, you can use the \texttt{matplotlib} package. If you’re doing data analysis, \texttt{pandas} can be useful. Packages make coding faster and more efficient by providing reusable code.

\section{Why Do We Need Packages?}

Packages save time by providing pre-built tools, so you don’t have to write everything from scratch. If you want to plot graphs, instead of writing your own plotting code, you can install a package like \texttt{matplotlib} or \texttt{seaborn} and use its functions. This way, you focus on solving the problem rather than building the tools from scratch.

\section{How to Use Packages in Python}

To use a package in Python, you first need to install it. Python uses a tool called \texttt{pip} to download and install packages from the Python Package Index (PyPI). Once installed, you can import the package into your code.

\section{What is \texttt{pip}?}

\texttt{pip} is a package manager that allows you to install and manage Python packages. It’s included with most Python installations, so you don’t need to install \texttt{pip} separately~\cite{pip}.

\section{Why Do We Need \texttt{pip}?}

\texttt{pip} helps manage all the external libraries you want to use in your Python projects. Whether you need libraries for data analysis, machine learning, or web development, \texttt{pip} makes the installation process easy.

\section{Example: Installing \texttt{seaborn} Using \texttt{pip}}

Let’s say you want to use the \texttt{seaborn} package, which is a powerful library for creating statistical graphics. You’ll first need to install it using \texttt{pip}.

\begin{lstlisting}[style=cmd]
pip install seaborn
\end{lstlisting}

The following screenshot shows what the installation process looks like in a command line:

\begin{center}
\includegraphics[width=1.0\textwidth]{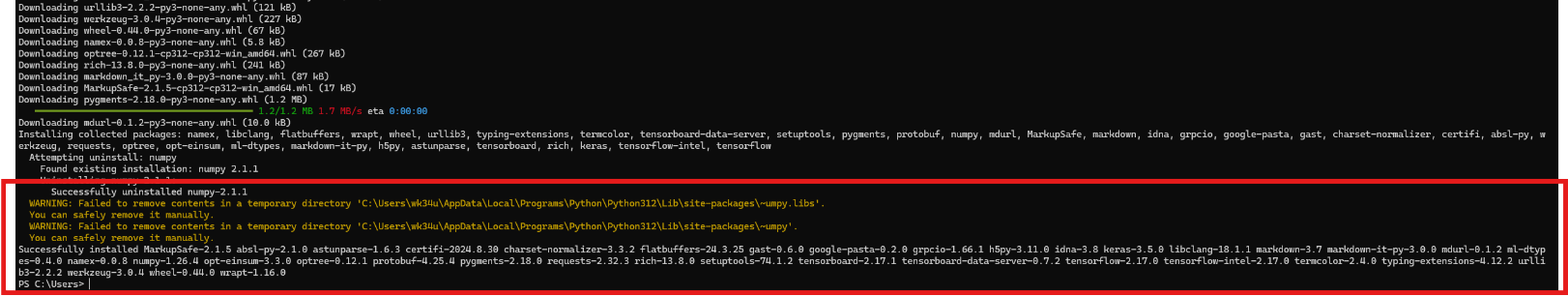}
\end{center}

As you can see, \texttt{pip} automatically downloads all the required dependencies, installs them, and finishes with a success message.

\section{Use \texttt{seaborn} to plot}

Seaborn~\cite{Waskom2021} is a powerful and user-friendly Python library built on top of Matplotlib~\cite{Hunter2007}, designed specifically for creating informative and attractive statistical graphics. It simplifies the process of creating complex visualizations by providing high-level interfaces for drawing various types of plots, such as categorical plots, distribution plots, and regression plots. Seaborn also integrates well with Pandas, allowing for easy plotting of data directly from DataFrames, making it a convenient tool for data exploration and analysis. Its default styles and color palettes also enhance the aesthetics of the plots without requiring extensive customization.

\subsection{Example: Using \texttt{seaborn} to Create a Simple Plot}

Now that \texttt{seaborn} is installed, let’s use it to create a simple plot. Here’s how you can use \texttt{seaborn} inside IDLE to generate a plot:

\begin{lstlisting}[style=python]
import numpy as np
import seaborn as sns
import matplotlib.pyplot as plt

# Create data for the heart shape
t = np.linspace(0, 2 * np.pi, 1000)
x = 16 * np.sin(t) ** 3
y = 13 * np.cos(t) - 5 * np.cos(2 * t) - 2 * np.cos(3 * t) - np.cos(4 * t)

# Set the theme for the plot
sns.set(style="whitegrid")

# Create the plot
plt.figure(figsize=(6, 6))
plt.plot(x, y, color="red")
plt.title("Heart Shape")
plt.axis("equal")
plt.show()
\end{lstlisting}

The image below shows how this looks when running the code in IDLE:
\begin{center}
\includegraphics[width=0.5\textwidth]{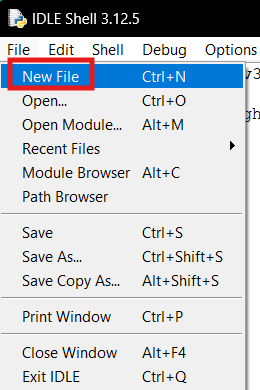}
\includegraphics[width=0.4\textwidth]{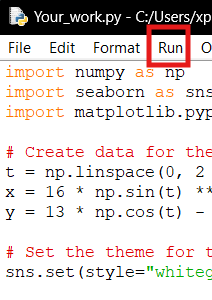}
\end{center}

This code uses \texttt{seaborn} to load a sample dataset of restaurant tips and create a scatter plot showing the relationship between the total bill and the tip amount.

After running this code, you will see the following plot:

\begin{center}
\includegraphics[width=0.6\textwidth]{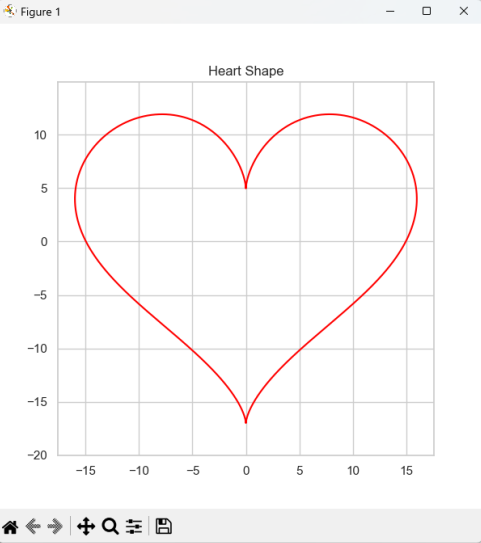}
\end{center}

As you can see, the plot visually represents the relationship between the total bill and the tip amount. \texttt{seaborn} makes it very easy to create beautiful visualizations with just a few lines of code.

\subsection{Another Example: Visualizing the \texttt{tips} Dataset}

The second example demonstrates how to use the \texttt{seaborn} package to create a simple scatter plot from the popular "tips" dataset~\cite{waskom2023pythontips}, which is included in \texttt{seaborn}. This dataset contains information about restaurant tips, such as total bill, tip amount, and various other factors.

Here’s the code to visualize the relationship between the total bill and the tip amount:

\begin{lstlisting}[style=python]
import seaborn as sns
import matplotlib.pyplot as plt

# Load the 'tips' dataset
tips = sns.load_dataset("tips")

# Set the theme for the plot
sns.set(style="whitegrid")

# Create a scatter plot
sns.scatterplot(x="total_bill", y="tip", data=tips)

# Show the plot
plt.title("Scatter Plot: Total Bill vs Tip")
plt.xlabel("Total Bill")
plt.ylabel("Tip Amount")
plt.show()
\end{lstlisting}

\begin{center}
\includegraphics[width=0.8\textwidth]{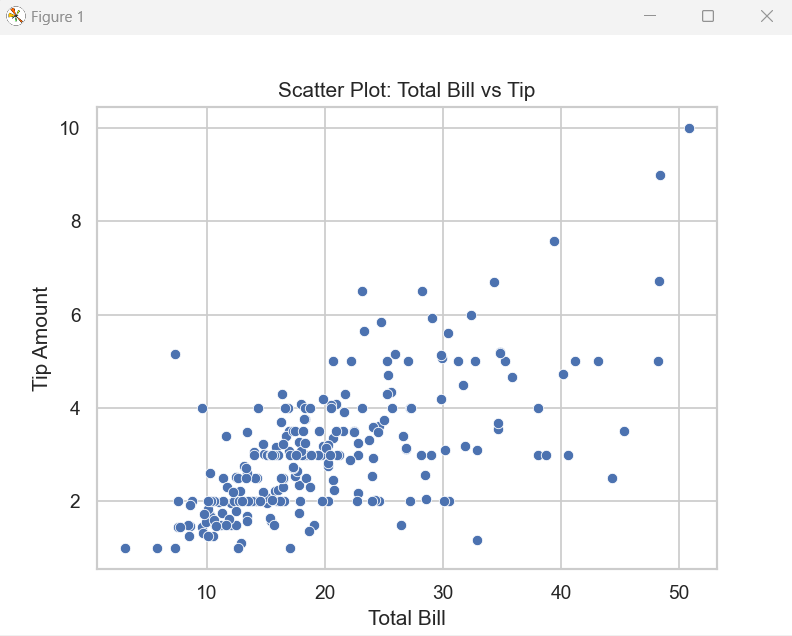}
\end{center}

user may see output shown as above. Now we have seen the example of using function and dataset in packages, in the next chapter, we will dive into more advanced topics, including machine learning and using Python to build models.

\section{Summary}

\begin{itemize}
    \item Python packages are collections of code that allow you to perform various tasks efficiently.
    \item \texttt{pip} is the tool used to install and manage Python packages.
    \item We installed the \texttt{seaborn} package using \texttt{pip}, and used it to generate a simple scatter plot.
\end{itemize}

\chapter{Machine Learning 101}

\section{What is Machine Learning?}

\textbf{Machine Learning (ML)} is a field of computer science that enables computers to learn from data without being explicitly programmed. Think of it as teaching a computer to recognize patterns, make decisions, or predict outcomes based on examples (data) that we give it. 

Instead of writing rules for the computer to follow, we provide it with data, and it learns the rules by itself.

\section{What Can Machine Learning Do?}

Machine Learning is everywhere around us. Here are a few examples of what ML can do:
\begin{itemize}
    \item \textbf{Predicting Outcomes:} ML can predict what might happen next. For example, predicting whether an email is spam or not, or predicting whether a customer will buy a product.
    \item \textbf{Image Recognition:} ML models can recognize faces or objects in photos and videos.
    \item \textbf{Personalization:} Online services like Netflix or Spotify use ML to recommend movies or songs based on your preferences.
    \item \textbf{Speech Recognition:} Voice assistants like Siri or Alexa use ML to understand spoken commands.
\end{itemize}

\section{Understanding Machine Learning}

It is easiest to understand ML by thinking of it as teaching a child. When you teach a child to recognize animals, you don't explain every animal's rules. Instead, you show them many pictures of animals and tell them what each one is. Eventually, they learn to recognize new animals by comparing them to previous examples.

It works similarly to ML. By providing a computer with a lot of data (examples), we allow it to find patterns on its own. As soon as it learns those patterns, it can make predictions based on new, unseen data.

\section{Introducing \texttt{scikit-learn}}

\texttt{scikit-learn} is one of the most popular and beginner-friendly packages for machine learning in Python~\cite{Pedregosa2011}. It provides easy-to-use tools for building machine learning models, like decision trees, linear regression, and more.

\subsection{Installing \texttt{scikit-learn}}

To install \texttt{scikit-learn}, use \texttt{pip} as we did in the previous chapter. Open your terminal and run:

\begin{lstlisting}[style=cmd]
pip install scikit-learn
\end{lstlisting}

Once installed, you can import it and use it in your Python code to create machine learning models.

\section{Training a Linear Regression Model on the Diabetes Dataset}

Now, let’s use \texttt{scikit-learn} to train a \textbf{Linear Regression} model on the Diabetes dataset\cite{diabetes}. 

\textbf{What is Linear Regression?} Linear Regression is one of the simplest machine learning models. It’s used to predict a numerical value (a continuous variable). For example, predicting a person’s weight based on their height. The idea is to find the best-fitting line that represents the relationship between input features (e.g., height) and the output (e.g., weight).

In this case, we will predict a diabetes progression value based on certain medical features.

\subsection{Step 1: Load the Diabetes Dataset and Train a Linear Regression Model}

Here’s the Python code for trainning a Linear Regression model on the Diabetes dataset using \texttt{scikit-learn}. We can load the dataset directly from /texttt{scikit-learn}.

\begin{lstlisting}[style=python]
# Import necessary libraries
import numpy as np
import matplotlib.pyplot as plt
from sklearn.datasets import load_diabetes
from sklearn.model_selection import train_test_split
from sklearn.linear_model import LinearRegression
from sklearn.metrics import mean_squared_error

# Step 1: Load the Diabetes dataset
diabetes = load_diabetes()
X = diabetes.data[:, np.newaxis, 2]  # Use only one feature (BMI) for simplicity
y = diabetes.target  # The target variable (progression of diabetes)

# Step 2: Split the data into training and testing sets
X_train, X_test, y_train, y_test = train_test_split(X, y, test_size=0.3, random_state=42)

# Step 3: Create and train the Linear Regression model
lr = LinearRegression()
lr.fit(X_train, y_train)

# Step 4: Make predictions on the test set
y_pred = lr.predict(X_test)

# Step 5: Calculate the Mean Squared Error (MSE)
mse = mean_squared_error(y_test, y_pred)
print(f"Mean Squared Error: {mse:.2f}")

# Step 6: Visualize the scatter plot and regression line
plt.figure(figsize=(8, 6))

# Scatter plot of the training data
plt.scatter(X_train, y_train, color="blue", label="Training data", alpha=0.6)

# Scatter plot of the test data
plt.scatter(X_test, y_test, color="green", label="Test data", alpha=0.6)

# Regression line (use entire dataset for better visualization)
plt.plot(X, lr.predict(X), color="red", linewidth=2, label="Regression line")

# Add labels and title
plt.title("Linear Regression on Diabetes Dataset (BMI vs Progression)")
plt.xlabel("BMI (Body Mass Index)")
plt.ylabel("Diabetes Progression")
plt.legend()

# Show the plot
plt.show()
\end{lstlisting}

\textbf{Explanation of the Code:}
\begin{itemize}
    \item \texttt{X = diabetes.data[:, np.newaxis, 2]}: We’re selecting only one feature, BMI (Body Mass Index), to make the regression simpler and easier to visualize. This limits the model to predicting diabetes progression based on BMI alone.
    \item \texttt{plt.scatter()}: Creates scatter plots of the training data (in blue) and test data (in green) to show the relationship between BMI and diabetes progression.
    \item \texttt{plt.plot()}: Draws the regression line (in red) that represents the linear relationship between BMI and diabetes progression.
    \item The plot includes clear labels, colors, and a legend to distinguish between training data, test data, and the regression line.
\end{itemize}

\section{Visualizing the Result}

This plot visually shows how well the linear regression model fits the data. The blue and green points represent the actual data (BMI vs diabetes progression), and the red line is the linear regression line predicted by the model. The closer the points are to the red line, the better the model's predictions are.

\begin{itemize}
    \item \textbf{Blue points:} Training data used to fit the model.
    \item \textbf{Green points:} Test data, which the model hasn’t seen before.
    \item \textbf{Red line:} The linear regression line representing the predicted relationship between BMI and diabetes progression.
\end{itemize}

\begin{center}
\includegraphics[width=0.8\textwidth]{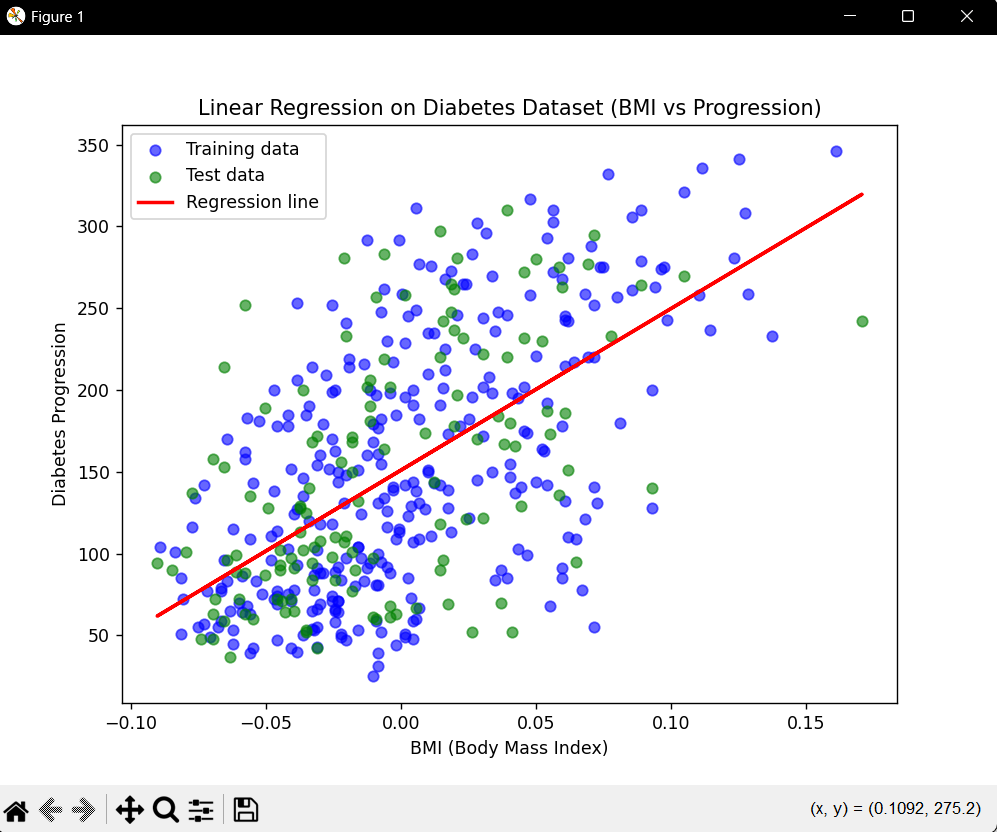}
\end{center}

\section{Training a Logistic Regression Model on the Breast Cancer Dataset}

Let’s use scikit-learn to train a Logistic Regression model on the Breast Cancer dataset.

\textbf{What is Logistic Regression?} Logistic Regression is a widely used machine learning algorithm for binary classification tasks. It models the probability that a given input belongs to one of two classes by fitting the data to a logistic function (also known as the sigmoid function). Unlike Linear Regression, which predicts continuous values, Logistic Regression is used to classify inputs into categories, such as "yes" or "no" (binary classification).

In this case, we will use the Breast Cancer dataset from scikit-learn to predict whether a tumor is benign or malignant based on specific medical features.

\subsection{Load the Breast Cancer Dataset and Train a Logistic Regression Model}

\begin{figure}
    \centering
    \includegraphics[width=0.7\linewidth]{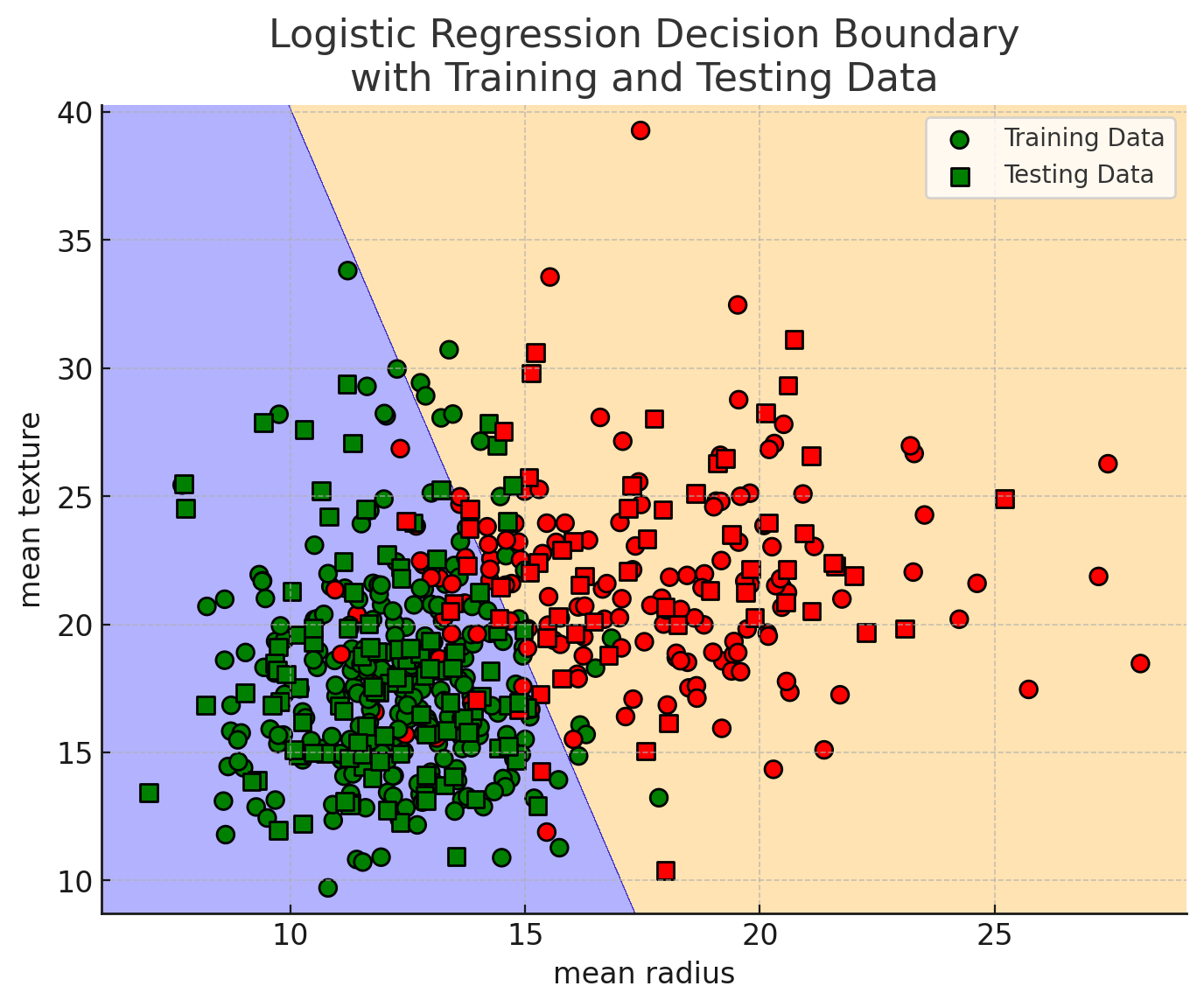}
    \caption{The graph displays the decision boundary of a Logistic Regression model trained on two features from the Breast Cancer dataset. The circles represent the training data, while the squares represent the testing data. Colors indicate the class labels: red for malignant tumors and green for benign tumors. The decision boundary separates the two classes, with the orange and blue regions indicating the predicted classification zones for malignant and benign cases, respectively. The plot effectively visualizes how well the model differentiates between the two tumor types.}
    \label{fig:logsitic-regression}
\end{figure}

Here’s the Python code to train a Logistic Regression model on the Breast Cancer dataset, see \textbf{Figure \ref{fig:logsitic-regression}} for our final result:

\begin{lstlisting}[style=python]
# Import necessary libraries
import numpy as np
import matplotlib.pyplot as plt
from sklearn.datasets import load_breast_cancer
from sklearn.model_selection import train_test_split
from sklearn.linear_model import LogisticRegression
from sklearn.metrics import accuracy_score, confusion_matrix, roc_curve, auc

# Step 1: Load the Breast Cancer dataset
cancer = load_breast_cancer()
X = cancer.data  # Using all features for this model
y = cancer.target  # Binary target (1: malignant, 0: benign)

# Step 2: Split the data into training and testing sets
X_train, X_test, y_train, y_test = train_test_split(X, y, test_size=0.3, random_state=42)

# Step 3: Create and train the Logistic Regression model
lr = LogisticRegression(max_iter=10000)
lr.fit(X_train, y_train)

# Step 4: Make predictions on the test set
y_pred = lr.predict(X_test)

# Step 5: Calculate the accuracy of the model
accuracy = accuracy_score(y_test, y_pred)

# Step 6: Visualize the training and testing data with the decision boundary
x_min, x_max = X[:, 0].min() - 1, X[:, 0].max() + 1
y_min, y_max = X[:, 1].min() - 1, X[:, 1].max() + 1
xx, yy = np.meshgrid(np.arange(x_min, x_max, 0.01),
                     np.arange(y_min, y_max, 0.01))

plt.figure(figsize=(8, 6))
plt.contourf(xx, yy, Z, alpha=0.3, cmap=ListedColormap(('orange', 'blue')))

# Scatter plot for training data
plt.scatter(X_train[:, 0], X_train[:, 1], c=y_train, s=50, edgecolor='k', marker='o', 
            cmap=ListedColormap(('red', 'green')), label='Training Data')

# Scatter plot for testing data
plt.scatter(X_test[:, 0], X_test[:, 1], c=y_test, s=50, marker='s', edgecolor='k', 
            cmap=ListedColormap(('red', 'green')), label='Testing Data')

# Labels and title
plt.title('Logistic Regression Decision Boundary\nwith Training and Testing Data')
plt.xlabel(cancer.feature_names[0])
plt.ylabel(cancer.feature_names[1])
plt.legend()
plt.show()

\end{lstlisting}

\textbf{Explanation of the Code:}

\begin{itemize}[noitemsep]
    \item \texttt{X = cancer.data}: We are using all features available in the dataset to predict the outcome (malignant or benign).
    \item \texttt{y = cancer.target}: This represents the binary classification labels (1 for malignant tumors, 0 for benign).
    \item \texttt{LogisticRegression(max\_iter=10000)}: Creates a Logistic Regression model. The parameter \texttt{max\_iter=10000} ensures that the optimization converges, as the default value may be insufficient for some datasets.
    \item \texttt{predict\_proba()}: Returns the probability estimates for the positive class (malignant tumors). This is useful for generating the ROC curve.
\end{itemize}

\section{Using ChatGPT to Generate Machine Learning Code}

\textbf{ChatGPT} is an AI language model developed by OpenAI~\cite{OpenAI2023}. You can use it to assist with generating code, understanding complex topics, and even debugging issues. If you're unsure how to write certain Python code, like training a Linear Regression model, ChatGPT can help by generating the code for you.

To use ChatGPT, go to \href{https://chat.openai.com}{chat.openai.com}, log in (or sign up if you're a new user), and then start a conversation by typing your request in plain language.

\subsection{How to Prompt ChatGPT for Code}

When you interact with ChatGPT, the more specific you are in your prompt, the more accurate and helpful the response will be. To generate a Linear Regression training code, you can provide the following instructions:

\textbf{Example Prompt:}
\begin{verbatim}
Can you generate Python code using scikit-learn to train a Linear Regression model on 
the Diabetes dataset? Please include code to split the data into training and testing sets, 
train the model, make predictions, and calculate the Mean Squared Error. Add in-line comments 
to explain each step.
\end{verbatim}

\subsection{Example Response from ChatGPT}

Here’s an example of the code ChatGPT might generate for this prompt:

\begin{lstlisting}[style=python]
# Import necessary libraries
from sklearn.datasets import load_diabetes
from sklearn.model_selection import train_test_split
from sklearn.linear_model import LinearRegression
from sklearn.metrics import mean_squared_error

# Load the Diabetes dataset
diabetes = load_diabetes()
X = diabetes.data  # Features
y = diabetes.target  # Labels

# Split the data into training and testing sets
X_train, X_test, y_train, y_test = train_test_split(X, y, test_size=0.3, random_state=42)

# Create the Linear Regression model
lr = LinearRegression()

# Train the model
lr.fit(X_train, y_train)

# Make predictions on the test set
y_pred = lr.predict(X_test)

# Calculate and print the Mean Squared Error
mse = mean_squared_error(y_test, y_pred)
print(f"Mean Squared Error: {mse:.2f}")
\end{lstlisting}

\section{Summary}

\begin{itemize}
    \item Machine Learning enables computers to learn from data and make predictions.
    \item \texttt{scikit-learn} is a popular Python package for building ML models.
    \item We used \texttt{scikit-learn} to train a Linear Regression~\cite{Schneider2010} model on the Diabetes dataset and evaluated it using Mean Squared Error.
    \item Linear Regression is a simple model for predicting continuous values based on input features.
    \item ChatGPT can be a valuable tool for generating code and understanding machine learning concepts.
\end{itemize}

\chapter{Deep Learning 101}

\section{What is Deep Learning?}

\textbf{Deep Learning (DL)} is a type of Machine Learning that focuses on models called \textbf{neural networks}. These models are inspired by how the human brain works and are very powerful at recognizing patterns in large amounts of data.

In Deep Learning, neural networks have several layers (which is why it's called "deep"). Each layer learns to represent data in a more complex way. For example, if you give it images, early layers may recognize simple shapes like edges, while deeper layers recognize whole objects like numbers or faces.

\section{What Can Deep Learning Do?}

Deep Learning can solve very complex tasks that traditional algorithms struggle with. Some examples include:

\begin{itemize}
    \item \textbf{Image Recognition:} Recognizing objects in pictures, like identifying handwritten digits or faces.
    \item \textbf{Language Understanding:} Translating languages, understanding commands, or generating text (e.g., ChatGPT and Gemini).
    \item \textbf{Speech Recognition:} Turning spoken words into text or understanding voice commands, like in Siri or Alexa.
    \item \textbf{Game Playing:} Deep Learning models can learn to play and master complex games like Chess or Go.
\end{itemize}

\section{How Does Deep Learning Work?}

The key idea in Deep Learning is the \textbf{neural network}. A neural network is made up of layers of connected nodes, called \textbf{neurons}. Each connection between neurons has a weight, which changes as the network learns from data. 
\begin{center}
\includegraphics[width=0.7\textwidth]{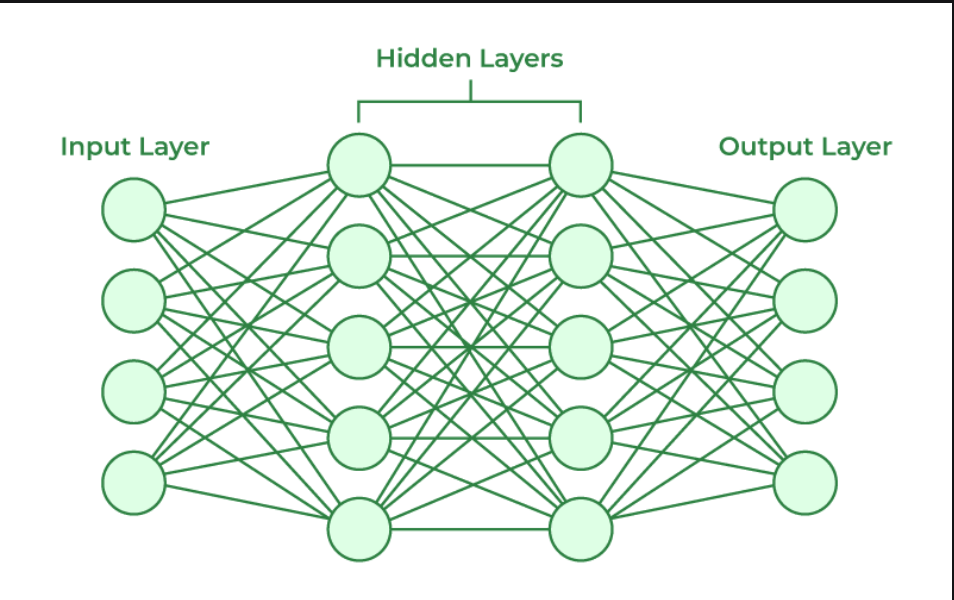}
\end{center}
In simple terms:
\begin{itemize}
    \item \textbf{Input Layer:} Receives the raw data (e.g., images or text).
    \item \textbf{Hidden Layers:} These layers process the data, learning patterns from it.
    \item \textbf{Output Layer:} Produces the final prediction, like classifying an image as a digit.
\end{itemize}

The more layers a network has, the more complex patterns it can learn. However, deep networks also need lots of data and computing power to work well.

\section{Mainstream Deep Learning Frameworks}

Deep learning involves many complex mathematical algorithms, which makes it difficult to build models from scratch. Fortunately, there are now several mainstream frameworks that allow us to bypass the mathematical details and write simple code to implement deep learning models. Two of the most popular frameworks are \textbf{TensorFlow 2}~\cite{tensorflow2015-whitepaper} and \textbf{PyTorch}~\cite{paszke2019pytorch}.

Below is a table comparing the pros and cons of each framework:

\begin{table}[H]
\centering
\begin{tabular}{|>{\centering\arraybackslash}p{7cm}|>{\centering\arraybackslash}p{7cm}|}
\hline
\textbf{TensorFlow 2} & \textbf{PyTorch} \\
\hline
\multicolumn{2}{|c|}{\textbf{Pros}} \\ \hline
\begin{itemize}[leftmargin=*, noitemsep]
    \item Simplified API, making it more beginner-friendly compared to previous versions.
    \item Strong support for deployment and production environments.
    \item Well-documented with a large and active community.
    \item Developed by Google, it updates quickly.
    \item It was once very popular and widely used.
\end{itemize} 
& 
\begin{itemize}[leftmargin=*, noitemsep]
    \item Easier to understand and use, especially for researchers and those who want flexibility.
    \item Dynamic computation graph allows for real-time tracking of the model's workflow and results.
    \item Increasingly popular in both academia and industry.
    \item It is the most used architecture currently and is still growing. Very promising.
\end{itemize} \\ \hline
\multicolumn{2}{|c|}{\textbf{Cons}} \\ \hline
\begin{itemize}[leftmargin=*, noitemsep]
    \item Despite improvements, still has a steeper learning curve compared to PyTorch.
    \item Debugging can sometimes be less intuitive due to its static computation graph.
    \item Updates are too fast and always use different APIs, which can cause serious incompatibility issues even if they are all TensorFlow 2.x.
    \item There is a lot of TensorFlow 1.x code on the market, but the incompatibility between TensorFlow 1.x and TensorFlow 2.x makes it difficult to migrate the code.
    \item The number of people using this architecture is decreasing.
\end{itemize}
& 
\begin{itemize}[leftmargin=*, noitemsep]
    \item Historically less mature for large-scale production and deployment compared to TensorFlow.
    \item Slightly smaller ecosystem for deployment tools.
    \item The code encapsulation level is low, Not very beginner friendly.
\end{itemize} \\ \hline
\multicolumn{2}{|c|}{\textbf{{\color{red}* It is strongly recommended that beginners focus their learning on PyTorch}}} \\ \hline
\end{tabular}
\caption{Comparison of TensorFlow 2 and PyTorch}
\end{table}

TensorFlow 2 and PyTorch APIs are quite similar, but switching between frameworks is still not a trivial task. However, with the advent of AI tools like ChatGPT, switching frameworks has become much easier. To provide readers with a clear comparison between TensorFlow 2 and PyTorch, we will present an example implemented in both frameworks below, allowing you to choose the framework that best suits your needs.

\section{Introducing \texttt{tensorflow} (TensorFlow 2)}

\texttt{tensorflow 2}, commonly known as TensorFlow, is a widely used tool for building and training deep learning models. It's scalable, beginner-friendly, and widely adopted in industry.

\subsection{Installing \texttt{tensorflow}}

To get started with \texttt{tensorflow}, you can install it using \texttt{pip}:

\begin{lstlisting}[style=cmd]
# Install TensorFlow
pip install tensorflow
\end{lstlisting}

\section{Training a Neural Network on the MNIST Dataset}

The \textbf{MNIST dataset}~\cite{MNIST} consists of 70,000 images of handwritten digits (0 to 9). We will use it to train a simple neural network that can recognize these digits.

\subsection{Step-by-Step: Training a Neural Network with TensorFlow 2}

Below is the equivalent Python code for training a neural network on the MNIST dataset using TensorFlow 2:

\begin{lstlisting}[style=python]
# Import necessary libraries
import tensorflow as tf
from tensorflow.keras import layers, models
from tensorflow.keras.datasets import mnist

# Step 1: Load and preprocess the MNIST dataset
(x_train, y_train), (x_test, y_test) = mnist.load_data()

# Normalize the images to values between 0 and 1
x_train, x_test = x_train / 255.0, x_test / 255.0

# Step 2: Define a simple neural network model
model = models.Sequential([
    layers.Flatten(input_shape=(28, 28)),    # Flatten the input image (28x28) to a 1D array
    layers.Dense(128, activation='relu'),    # Hidden layer with 128 neurons and ReLU activation
    layers.Dense(64, activation='relu'),     # Another hidden layer with 64 neurons
    layers.Dense(10)                         # Output layer with 10 neurons (one for each digit)
])

# Step 3: Compile the model
model.compile(optimizer='adam',              # Adam optimizer
              loss=tf.keras.losses.SparseCategoricalCrossentropy(from_logits=True),
              metrics=['accuracy'])

# Step 4: Train the neural network
model.fit(x_train, y_train, epochs=5, batch_size=64)

# Step 5: Test the neural network
test_loss, test_acc = model.evaluate(x_test, y_test, verbose=2)
print(f"Accuracy: {test_acc * 100:.2f}%")
\end{lstlisting}

\textbf{Explanation of the Code:}
\begin{itemize}
    \item \texttt{mnist.load\_data()}: Loads the MNIST dataset.
    \item \texttt{x\_train / 255.0}: Normalizes the pixel values between 0 and 1.
    \item \texttt{Sequential()}: Defines a sequential model where layers are added one by one.
    \item \texttt{Flatten()}: Converts the 28x28 images into a flat vector.
    \item \texttt{Dense()}: A fully connected layer. The last layer has 10 neurons for the 10 digit classes.
    \item \texttt{compile()}: Specifies the optimizer, loss function, and evaluation metrics.
    \item \texttt{fit()}: Trains the model on the training data for a specified number of epochs.
    \item \texttt{evaluate()}: Tests the model and returns the loss and accuracy.
\end{itemize}

\section{Training Output and Accuracy}

During the training process, you will see continuous updates on the model's performance after each batch of data. These updates include metrics such as the loss and accuracy. Below is an example of how this output looks during training, where the accuracy at the end of training is highlighted:

\begin{center}
\includegraphics[width=0.8\textwidth]{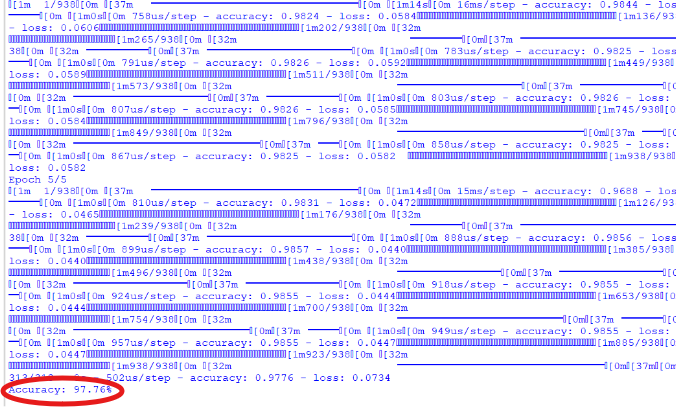}
\end{center}

As shown in the screenshot, the final accuracy of the model is 97.76\%. This accuracy value represents how well the model is performing on the test dataset after training. The lower the loss and the higher the accuracy, the better the model is at making correct predictions.

\section{Introducing \texttt{torch} (PyTorch)}

\texttt{torch}, also known as PyTorch, is another popular tool for building and training deep learning models. It's flexible, easy to use, and widely adopted by both researchers and developers.

\subsection{Installing \texttt{torch}}

To get started with \texttt{torch}, we need to install it. Here’s how you can install PyTorch using \texttt{pip}:

\begin{lstlisting}[style=cmd]
# Install PyTorch with CPU support (no GPU needed)
pip install torch torchvision
\end{lstlisting}

\texttt{torchvision} is an extra package that helps with handling image data, like the MNIST dataset we’ll use in the next section.

\section{Training a Neural Network on the MNIST Dataset}

The \textbf{MNIST dataset} is a collection of 70,000 images of handwritten digits (0 to 9). We’ll use it to train a simple neural network that can recognize these digits. 

\subsection{Step-by-Step: Training a Neural Network with PyTorch}

Here’s a Python example of training a neural network using the MNIST dataset with PyTorch. We will load the dataset, define the network, train it, and evaluate its accuracy.

\begin{lstlisting}[style=python]
# Import necessary libraries
import torch
import torch.nn as nn
import torch.optim as optim
from torch.utils.data import DataLoader
from torchvision import datasets, transforms

# Step 1: Prepare the MNIST dataset
transform = transforms.Compose([transforms.ToTensor(), transforms.Normalize((0.5,), (0.5,))])
train_data = datasets.MNIST(root='mnist_data', train=True, transform=transform, download=True)
test_data = datasets.MNIST(root='mnist_data', train=False, transform=transform, download=True)

# Step 2: Create data loaders for training and testing
train_loader = DataLoader(train_data, batch_size=64, shuffle=True)
test_loader = DataLoader(test_data, batch_size=64, shuffle=False)

# Step 3: Define a simple neural network
class NeuralNet(nn.Module):
    def __init__(self):
        super(NeuralNet, self).__init__()
        self.fc1 = nn.Linear(28 * 28, 128)  # Input layer (28x28 image = 784)
        self.fc2 = nn.Linear(128, 64)       # Hidden layer
        self.fc3 = nn.Linear(64, 10)        # Output layer (10 digits)

    def forward(self, x):
        x = x.view(-1, 28 * 28)            # Flatten the image
        x = torch.relu(self.fc1(x))        # Apply ReLU activation to input
        x = torch.relu(self.fc2(x))        # Apply ReLU to hidden layer
        x = self.fc3(x)                    # Output the result (logits)
        return x

# Step 4: Initialize the model, loss function, and optimizer
model = NeuralNet()
criterion = nn.CrossEntropyLoss()           # Loss function for classification
optimizer = optim.Adam(model.parameters(), lr=0.001)  # Optimizer to update weights

# Step 5: Train the neural network
epochs = 5
for epoch in range(epochs):
    for images, labels in train_loader:
        optimizer.zero_grad()               # Clear gradients from previous step
        output = model(images)              # Forward pass (make prediction)
        loss = criterion(output, labels)    # Compute loss
        loss.backward()                     # Backpropagation
        optimizer.step()                    # Update weights

    print(f"Epoch {epoch+1}/{epochs}, Loss: {loss.item()}")

# Step 6: Test the neural network
correct = 0
total = 0
with torch.no_grad():  # No need to compute gradients during testing
    for images, labels in test_loader:
        output = model(images)
        _, predicted = torch.max(output.data, 1)  # Get predicted digit
        total += labels.size(0)
        correct += (predicted == labels).sum().item()

accuracy = 100 * correct / total
print(f"Accuracy: {accuracy:.2f}%")
\end{lstlisting}

\textbf{Explanation of the Code:}
\begin{itemize}
    \item \texttt{transforms.Compose()}: Normalizes and converts images to tensors.
    \item \texttt{DataLoader()}: Loads the dataset in batches for faster training.
    \item \texttt{NeuralNet()}: Defines a basic neural network with one hidden layer.
    \item \texttt{optimizer.zero\_grad()}: Resets gradients before updating weights.
    \item \texttt{loss.backward()}: Computes the gradient for backpropagation.
    \item \texttt{torch.max()}: Finds the highest probability class (digit) for each prediction.
\end{itemize}

\section{Understanding Neural Networks (Simplified)}

Neural networks are made of layers of neurons, and each connection between neurons has a weight that the network adjusts during training. Here’s a simple breakdown of how it works:

\begin{itemize}
    \item \textbf{Input Layer:} Receives raw data (in this case, an image of a digit).
    \item \textbf{Hidden Layer:} Processes the input and looks for patterns (e.g., lines, shapes).
    \item \textbf{Output Layer:} Produces the final result (the digit the network thinks it is).
\end{itemize}

The network learns by adjusting its weights based on the errors it makes during training. Over time, it gets better at making accurate predictions.

\section{Using ChatGPT to Generate Deep Learning Code}

If you're unsure how to write deep learning code or need help, ChatGPT can be a helpful tool. You can ask it to generate code for training a neural network or to explain a concept in detail.

\subsection{How to Prompt ChatGPT for Code}

When asking ChatGPT for deep learning code, it's important to be clear in your request. Here’s an example prompt you could use:

\subsubsection{TensorFlow 2 Version}

\textbf{Example Prompt:}
\begin{verbatim}
Can you generate Python code using TensorFlow 2 to train a neural network on the MNIST dataset? 
Please include code to load the dataset, define a simple neural network, train it, and 
evaluate its accuracy. Add in-line comments to explain each step.
\end{verbatim}

\textbf{Example Response from ChatGPT}

Here’s an example of what ChatGPT might generate in TensorFlow 2:

\begin{lstlisting}[style=python]
# Import necessary libraries
import tensorflow as tf
from tensorflow.keras import layers, models
from tensorflow.keras.datasets import mnist
import numpy as np

# Load MNIST dataset
(x_train, y_train), (x_test, y_test) = mnist.load_data()
x_train, x_test = x_train / 255.0, x_test / 255.0

# Define the neural network
model = models.Sequential([
    layers.Flatten(input_shape=(28, 28)),
    layers.Dense(128, activation='relu'),
    layers.Dense(64, activation='relu'),
    layers.Dense(10)
])

# Compile the model
model.compile(optimizer='adam',
              loss=tf.keras.losses.SparseCategoricalCrossentropy(from_logits=True),
              metrics=['accuracy'])

# Train the model
model.fit(x_train, y_train, epochs=5, batch_size=64)

# Evaluate the model
test_loss, test_acc = model.evaluate(x_test, y_test, verbose=2)
print(f'Accuracy: {test_acc:.2f}%')
\end{lstlisting}

\subsubsection{PyTorch Version}

\textbf{Example Prompt:}
\begin{verbatim}
Can you generate Python code using PyTorch to train a neural network on the MNIST dataset? 
Please include code to load the dataset, define a simple neural network, train it, and 
evaluate its accuracy. Add in-line comments to explain each step.
\end{verbatim}

\textbf{Example Response from ChatGPT}

Here’s an example of what ChatGPT might generate:

\begin{lstlisting}[style=python]
# Import necessary libraries
import torch
import torch.nn as nn
from torchvision import datasets, transforms
from torch.utils.data import DataLoader

# Load MNIST dataset
transform = transforms.Compose([transforms.ToTensor(), transforms.Normalize((0.5,), (0.5,))])
train_data = datasets.MNIST(root='mnist_data', train=True, transform=transform, download=True)
test_data = datasets.MNIST(root='mnist_data', train=False, transform=transform, download=True)

# Create data loaders
train_loader = DataLoader(train_data, batch_size=64, shuffle=True)
test_loader = DataLoader(test_data, batch_size=64, shuffle=False)

# Define the neural network
class NeuralNet(nn.Module):
    def __init__(self):
        super(NeuralNet, self).__init__()
        self.fc1 = nn.Linear(28*28, 128)
        self.fc2 = nn.Linear(128, 64)
        self.fc3 = nn.Linear(64, 10)

    def forward(self, x):
        x = x.view(-1, 28*28)
        x = torch.relu(self.fc1(x))
        x = torch.relu(self.fc2(x))
        x = self.fc3(x)
        return x

# Initialize model, loss function, and optimizer
model = NeuralNet()
criterion = nn.CrossEntropyLoss()
optimizer = torch.optim.Adam(model.parameters(), lr=0.001)

# Train the model
for epoch in range(5):
    for images, labels in train_loader:
        optimizer.zero_grad()
        output = model(images)
        loss = criterion(output, labels)
        loss.backward()
        optimizer.step()

# Evaluate the model
correct = 0
total = 0
with torch.no_grad():
    for images, labels in test_loader:
        output = model(images)
        _, predicted = torch.max(output.data, 1)
        total += labels.size(0)
        correct += (predicted == labels).sum().item()

accuracy = 100 * correct / total
print(f'Accuracy: {accuracy:.2f}%')
\end{lstlisting}

\section{Summary}

\begin{itemize}
    \item Deep Learning is a type of Machine Learning that uses neural networks to learn from large amounts of data.
    \item We used PyTorch to train a neural network on the MNIST dataset.
    \item Neural networks consist of layers that learn patterns in data, adjusting their weights to improve accuracy.
    \item ChatGPT can assist with generating Deep Learning code and explaining concepts step-by-step.
\end{itemize}

\chapter{Cool AI Applications That Are Changing the Game}

\begin{figure}[h]
    \centering
    \includegraphics[width=1.0\linewidth]{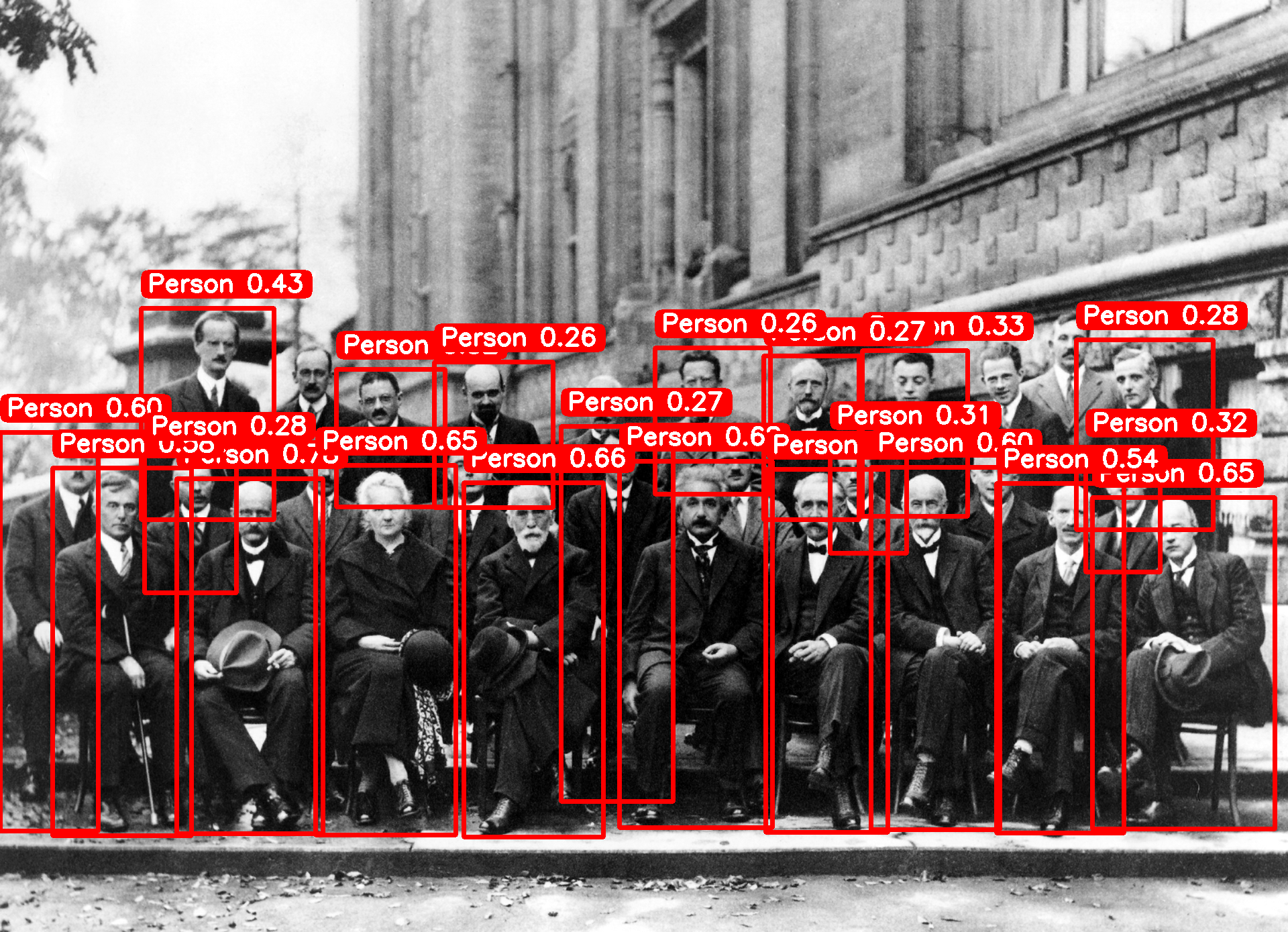}
    \caption{Real-world applications of AI object detection (by YOLO v8~\cite{ultralytics2023yolov8}). Notice how multiple objects are detected, segmented, and classified in real time, showcasing the efficiency of modern CV algorithms.}
    \label{fig:scientist}
\end{figure}

AI isn’t just hype anymore—it’s everywhere, making things that seemed impossible just a few years ago look effortless today. From generating code to identifying objects in images and understanding human language, AI has become the backbone of modern technology. This chapter dives into some of the coolest AI applications that are not just impressive but are transforming industries and daily life in big ways.

\section{Cutting-Edge Computer Vision (CV) Tools}

\subsection{Segment Anything Model (SAM)}

Ever wish you could instantly cut out an object from a picture or video, no matter how complex the background? \textbf{Segment Anything Model (SAM)}~\cite{kirillov2023segment} is making that wish come true. It lets you segment objects from images and videos in one click—perfect for designers, media editors, or even autonomous cars. The SAM web-based GUI is intuitive, allowing you to select and isolate objects with minimal effort. With a simple drag and click, objects are segmented, which makes the editing process more efficient and reduces the need for manual pixel-level adjustments.

\begin{figure}[h]
    \centering
    \includegraphics[width=0.48\linewidth]{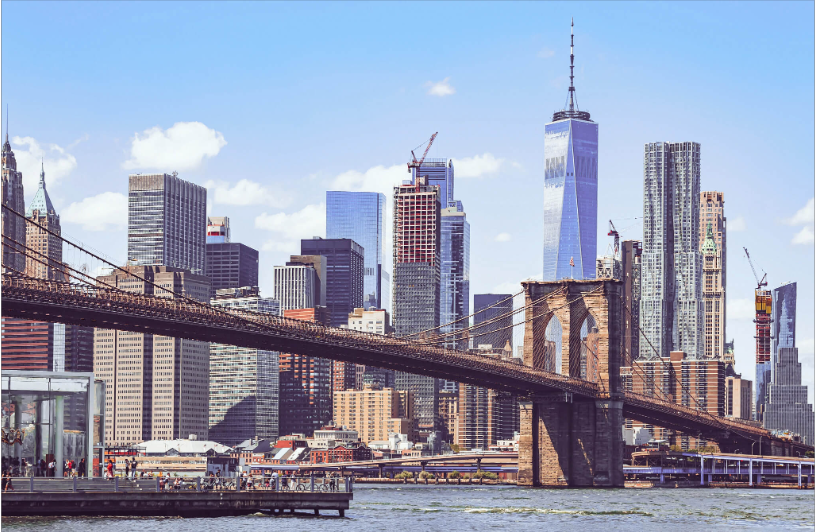}
    \includegraphics[width=0.48\linewidth]{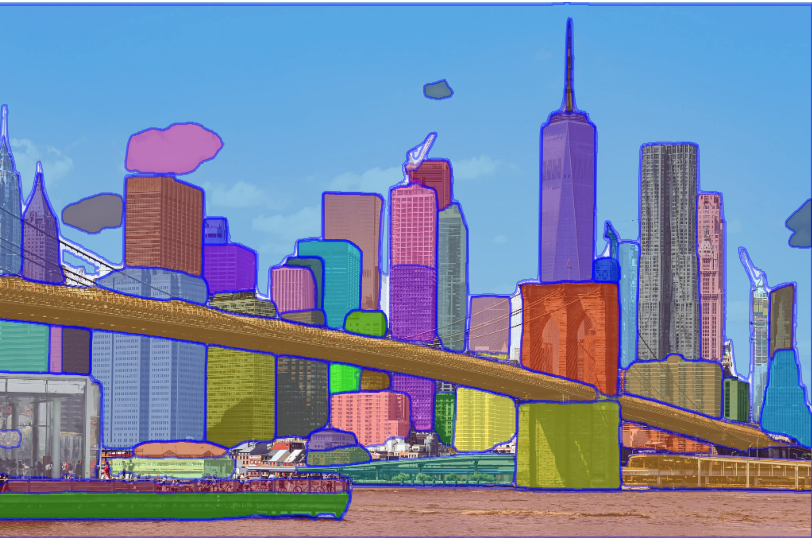}
    \caption{Left: Original image before SAM processing; Right: SAM's segmented objects. SAM can handle complex scenes, segmenting objects from different categories with precision, making it a revolutionary tool in image processing.}
    \label{fig:sam-combined}
\end{figure}

\textbf{What makes SAM so fascinating:}
SAM is powered by convolutional neural networks (CNNs)~\cite{oshea2015introductionconvolutionalneuralnetworks}, which have been trained on massive datasets to recognize all kinds of objects, no matter the shape, size, or environment. It can handle everything from cutting out a person in a crowded room to isolating a product from a shelf. The web interface provides real-time previews, so users can see the results instantly, making workflows faster.

\textbf{Inspiration behind SAM:} SAM reflects the future of AI popularization — AI that requires minimal human input to achieve complex tasks. Developers behind SAM have made it clear that their goal is to make image editing as simple as possible, taking away hours of tedious manual work.

\subsection{Digital Avatars}

From video games to social media, AI-generated \textbf{Digital Avatars}~\cite{Digital_Avatars} are here, and they look more lifelike than ever. These avatars can express emotions, make gestures, and even speak, making them an integral part of virtual reality worlds, online influencers, and more. The user interface of these avatar-creating tools is often straightforward, letting users craft highly realistic avatars with just a few clicks. Whether you’re a gamer or a content creator, Digital Avatars allow for personal customization that mirrors real-world behaviors and expressions.

\textbf{Why Digital Avatars are a big deal:}
Digital avatars, powered by Generative Adversarial Networks (GANs)~\cite{goodfellow2014generativeadversarialnetworks}, have made it possible for virtual characters to behave just like humans. In the metaverse, digital avatars represent you, interact with others, and even attend meetings on your behalf. The ease of use and the ability to fine-tune details—such as facial expressions and micro-interactions—mean anyone can create a digital version of themselves or a fictional character with little technical expertise.

\textbf{Virtual avatars and influence:} Trevor McFedries, founder of Brud, the company behind the AI influencer Lil Miquela, once said, "The future is a world where people will be engaging with personalities that aren’t even human." That future is now, and digital avatars are leading the charge.

\section{AI in Natural Language Processing (NLP)}

\subsection{ChatGPT}

\textbf{ChatGPT}~\cite{OpenAI2023} hardly needs an introduction. This conversational AI from OpenAI can chat with you about anything—from writing a poem to answering tough technical questions. ChatGPT has become the go-to tool for creative writing, coding, and even providing instant customer service. The web interface is user-friendly, allowing for real-time interaction where the AI responds within seconds, making it highly efficient for both individual users and businesses.

\begin{figure}[h]
    \centering
    \includegraphics[width=1.0\linewidth]{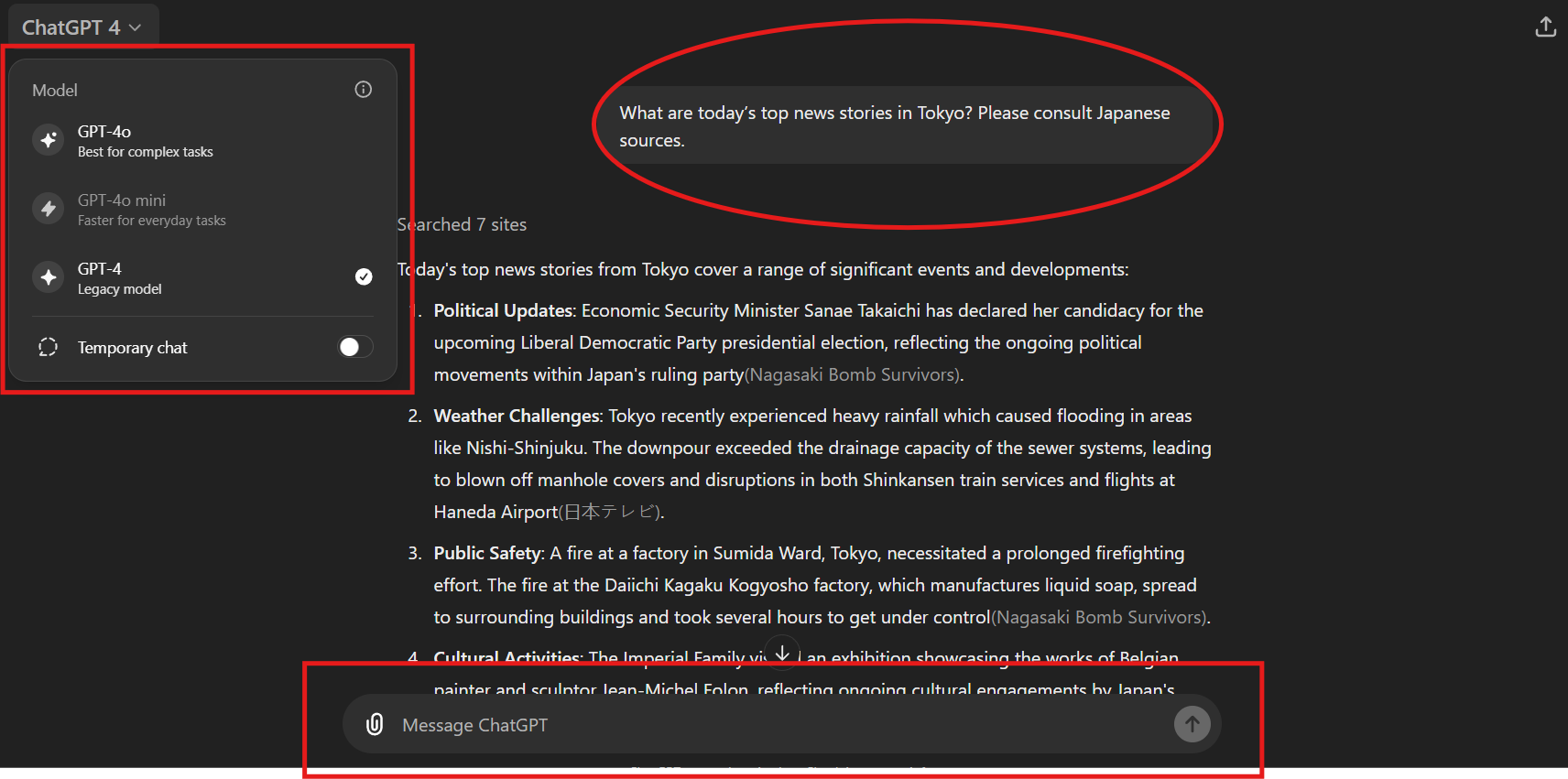}
    \caption{ChatGPT in action answering a query. The interface provides a seamless experience, with responses generated in real time, offering both casual conversation and technical support with ease.}
    \label{fig:chatgpt-demo}
\end{figure}

\textbf{The magic behind ChatGPT:}
ChatGPT runs on the GPT (Generative Pretrained Transformer)~\cite{yenduri2023generativepretrainedtransformercomprehensive} architecture, trained on billions of texts to generate human-like responses. It’s incredibly versatile and can adapt its responses to different tones and styles, making it a powerful assistant in almost any field. Whether you need it for professional use or casual questions, the platform’s easy-to-navigate GUI means anyone can engage with the AI effortlessly.

\textbf{What Sam Altman said:} OpenAI’s CEO Sam Altman once remarked, "AI is the most important technology that humanity has ever developed." ChatGPT certainly lives up to that claim, enabling users worldwide to harness the power of advanced language models with just a few keystrokes.

\subsection{Gemini: The AI Personal Assistant}

\textbf{Gemini}~\cite{geminiteam2024geminifamilyhighlycapable} takes AI personal assistants to the next level. Alexa and Siri help you with basic tasks, but Gemini learns from your habits and anticipates your needs. Whether you're drafting emails, scheduling meetings, or responding to texts, Gemini's intuitive interface makes organizing your life simple.

\begin{figure}[h]
    \centering
    \includegraphics[width=0.2\linewidth]{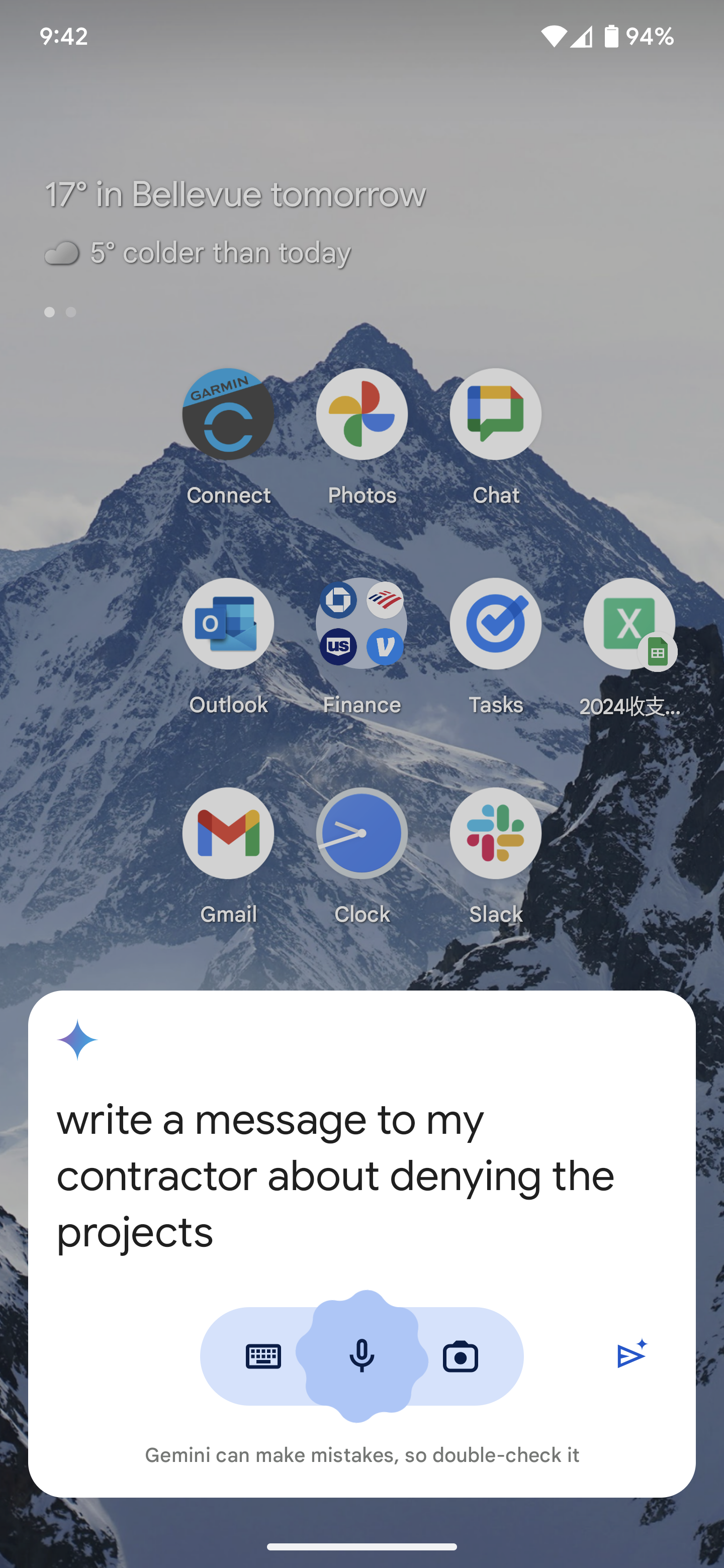}
    \includegraphics[width=0.2\linewidth]{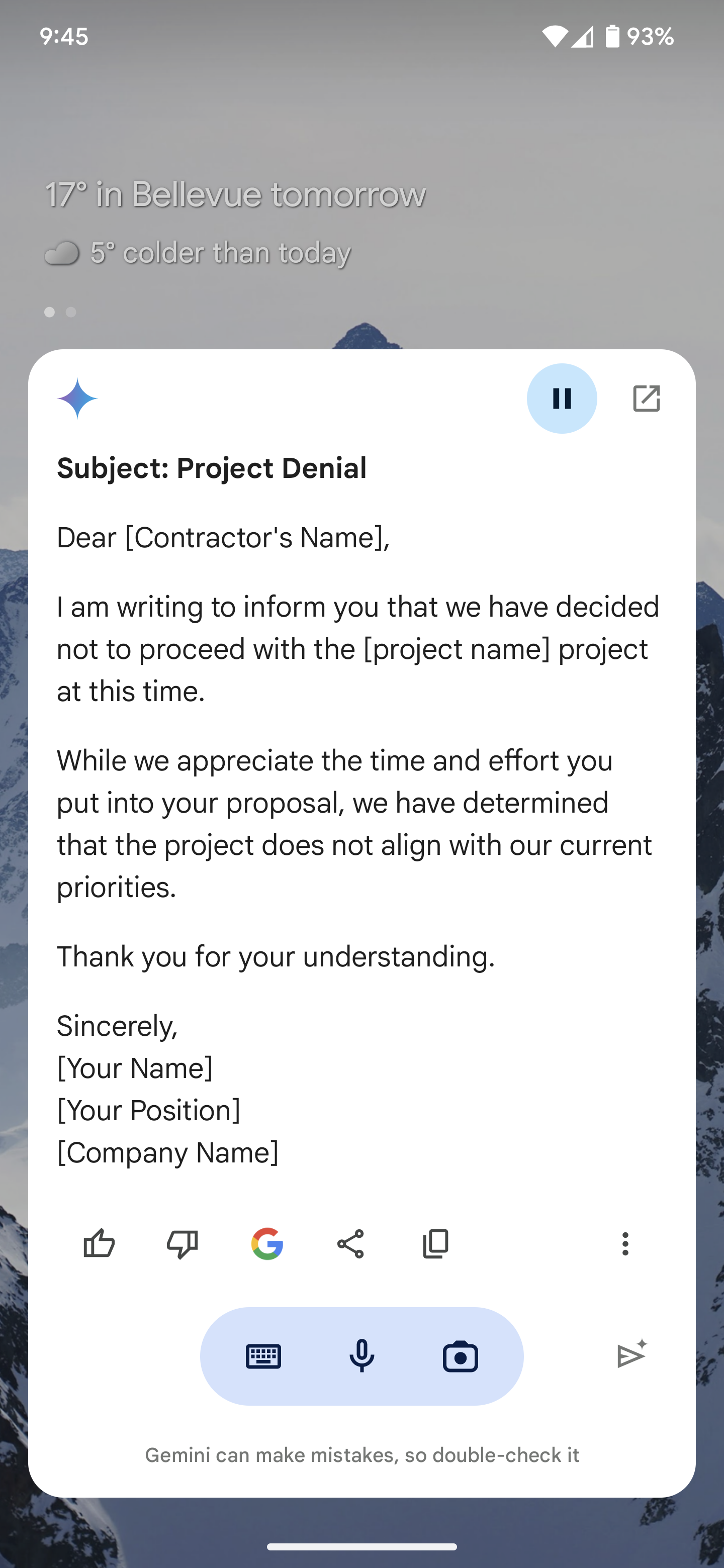}
    \includegraphics[width=0.2\linewidth]{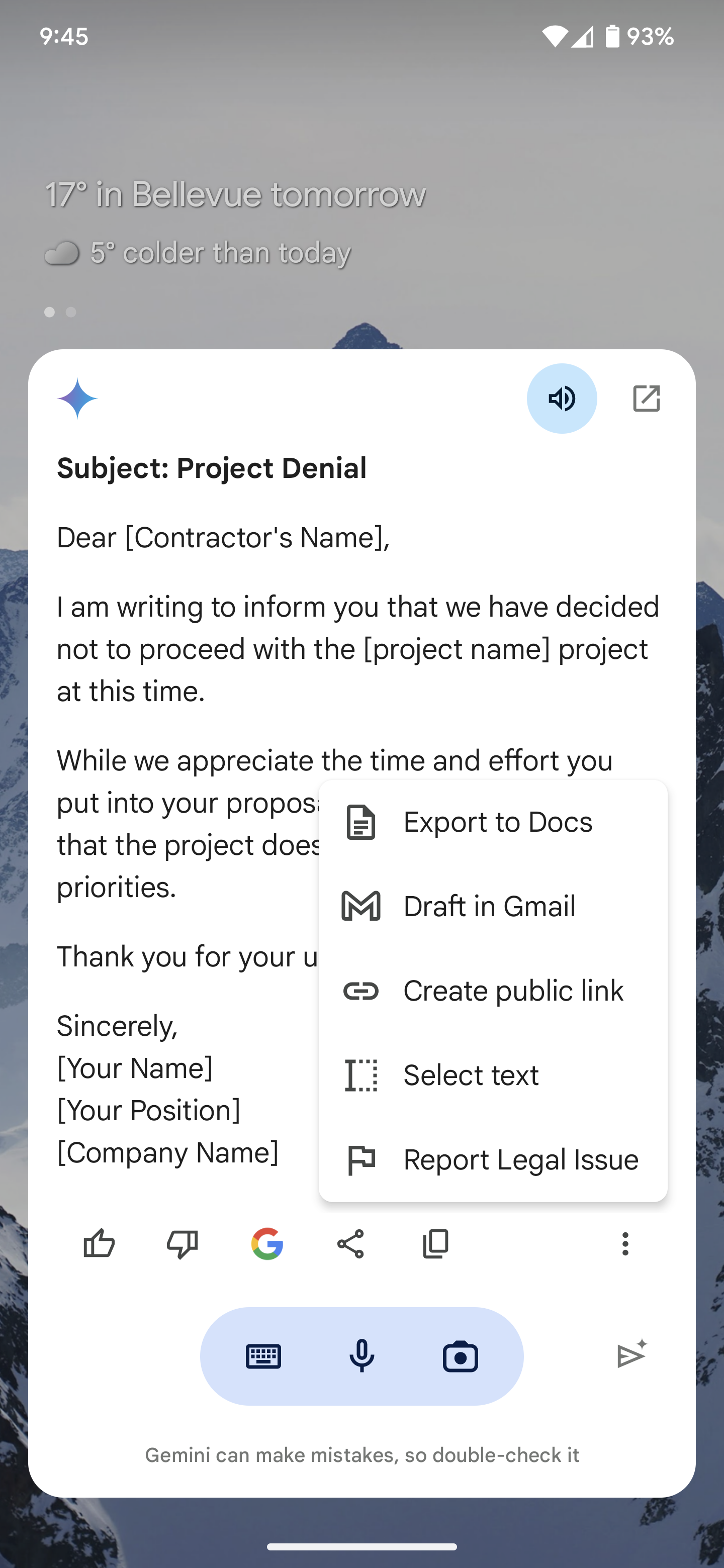}
    \includegraphics[width=0.2\linewidth]{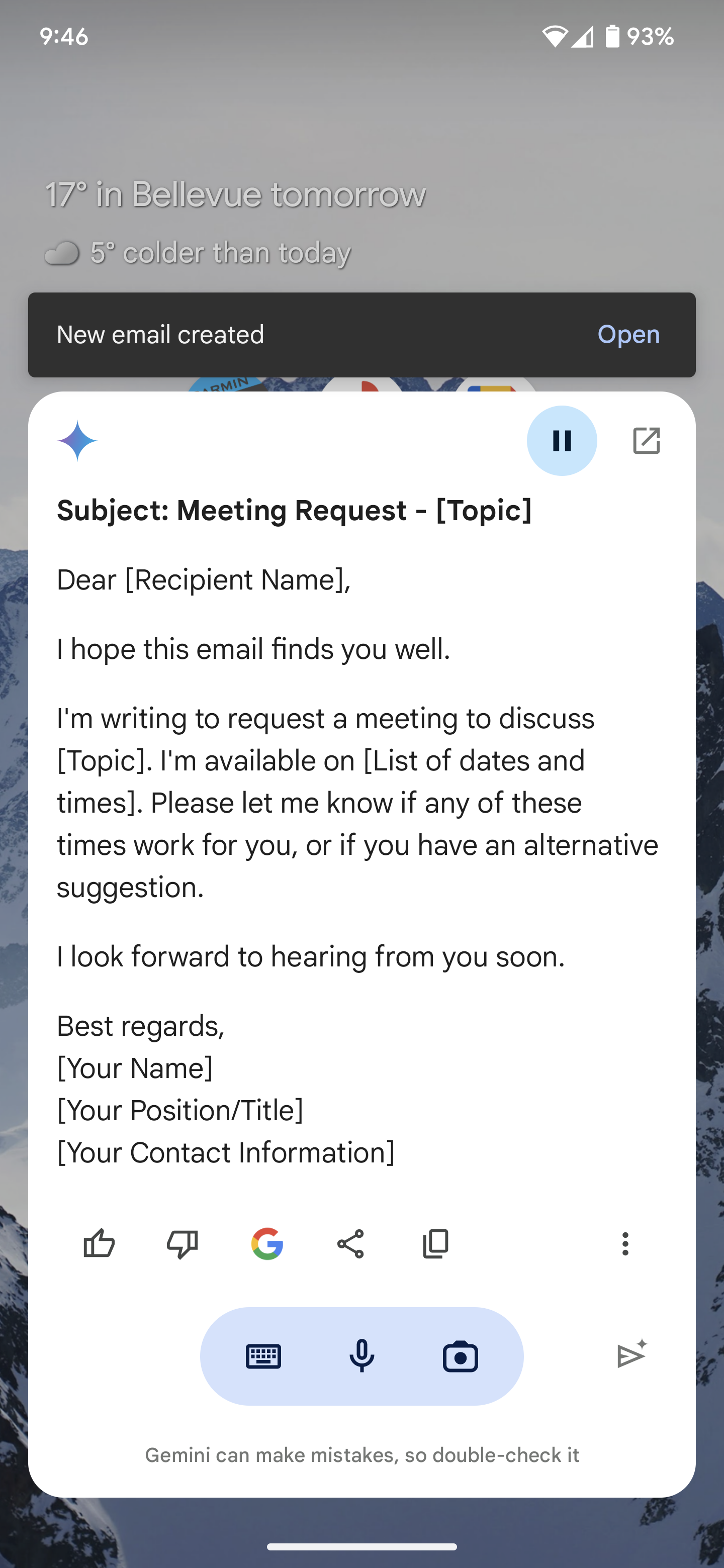}
    \caption{Starting a conversation with Gemini. The interface is intuitive and engaging, allowing users to begin tasks with simple commands.Gemini responds efficiently with personalized insights and tasks. Here it shows a list of actions that it can handle without any need for manual setup.Gemini can perform various operations, from scheduling to reminders, shown in this interface as options highlighted for user input.Gemini generating an email draft based on the user’s preferences—without typing a single word. This feature shows how the assistant handles repetitive tasks effortlessly.}
    \label{fig:gemini-step1}
\end{figure}

\textbf{What sets Gemini apart:}
Unlike typical assistants, Gemini uses reinforcement learning to get better the more you interact with it. Whether you’re planning meetings, answering emails, or setting reminders, Gemini can handle all of it without skipping a beat. The interface is designed to minimize the steps required to complete tasks, making it much faster than traditional personal assistants.

\textbf{Elon Musk’s take:} "We’re headed toward a world where AI will assist us in almost everything." Musk's vision seems to come to life with assistants like Gemini, which promise a future where technology is seamlessly woven into our daily tasks.

\section{AI for Creative Industries}

\subsection{Flux: AI for Photography}

Editing photos has never been easier thanks to \textbf{Flux}~\cite{chang2024fluxfastsoftwarebasedcommunication}, an AI-powered photography tool that can enhance your images with a single click. Whether you want to adjust lighting, remove unwanted objects, or apply artistic filters, Flux takes care of it all. The tool’s web interface is designed to be as simple as possible, offering sliders and one-click presets that instantly improve your photos.

\textbf{How Flux is reshaping photography:}
With convolutional neural networks (CNNs)~\cite{oshea2015introductionconvolutionalneuralnetworks} as its backbone, Flux identifies elements like faces and objects, optimizing them for perfect lighting, balance, and color. This means even amateur photographers can achieve professional results in a fraction of the time. The tool’s AI also suggests creative adjustments, turning ordinary shots into visually stunning images.
\subsection{AI Coding Tools: The Future of Software Development}
The world of programming is being revolutionized by AI tools designed to boost productivity and assist developers at every step. Let’s dive into three major players in the AI coding space: \textbf{Cursor}~\cite{cursor2023}, \textbf{GitHub Copilot}~\cite{githubcopilot2023}, and \textbf{OpenDevin}~\cite{wang2024opendevinopenplatformai}.

\textbf{Cursor:}
Cursor is your AI coding buddy—whether you’re debugging or writing code, Cursor is there to help. It suggests code snippets, fixes bugs, and even generates entire functions based on your input, making coding more efficient than ever. The web GUI is streamlined, allowing developers to see real-time code suggestions and documentation, making it easy to implement solutions with minimal manual coding.

\begin{figure}[h]
    \centering
    \includegraphics[width=0.48\linewidth]{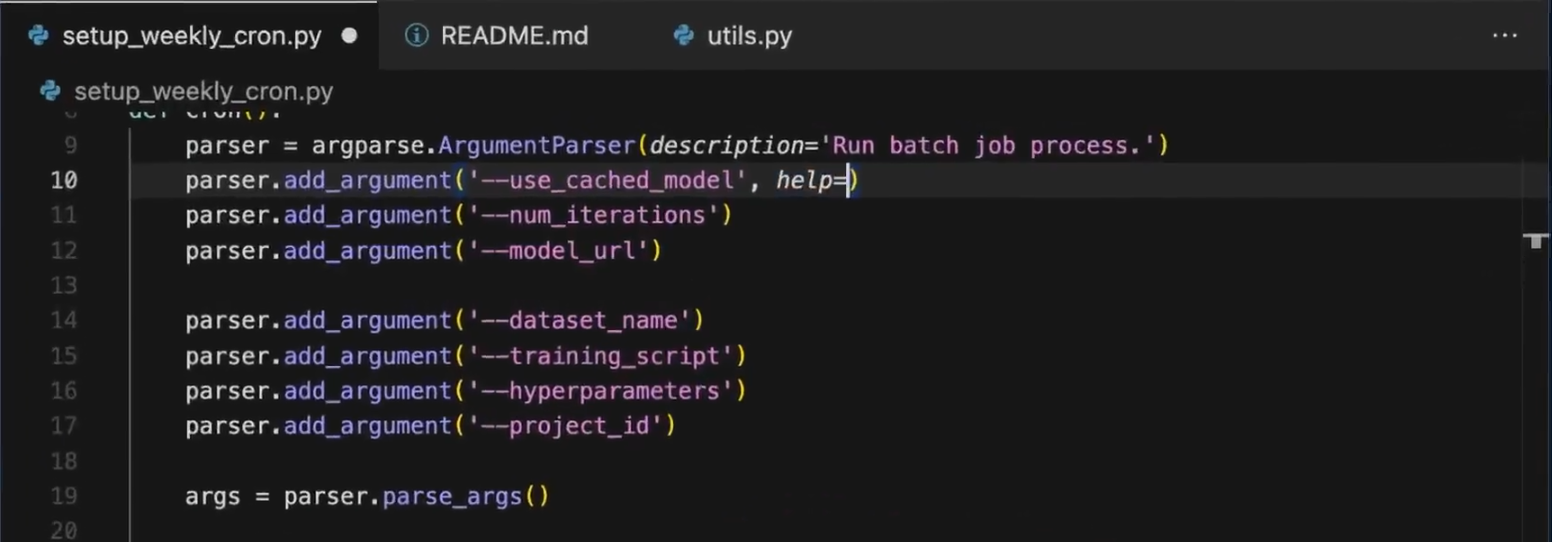}
    \includegraphics[width=0.48\linewidth]{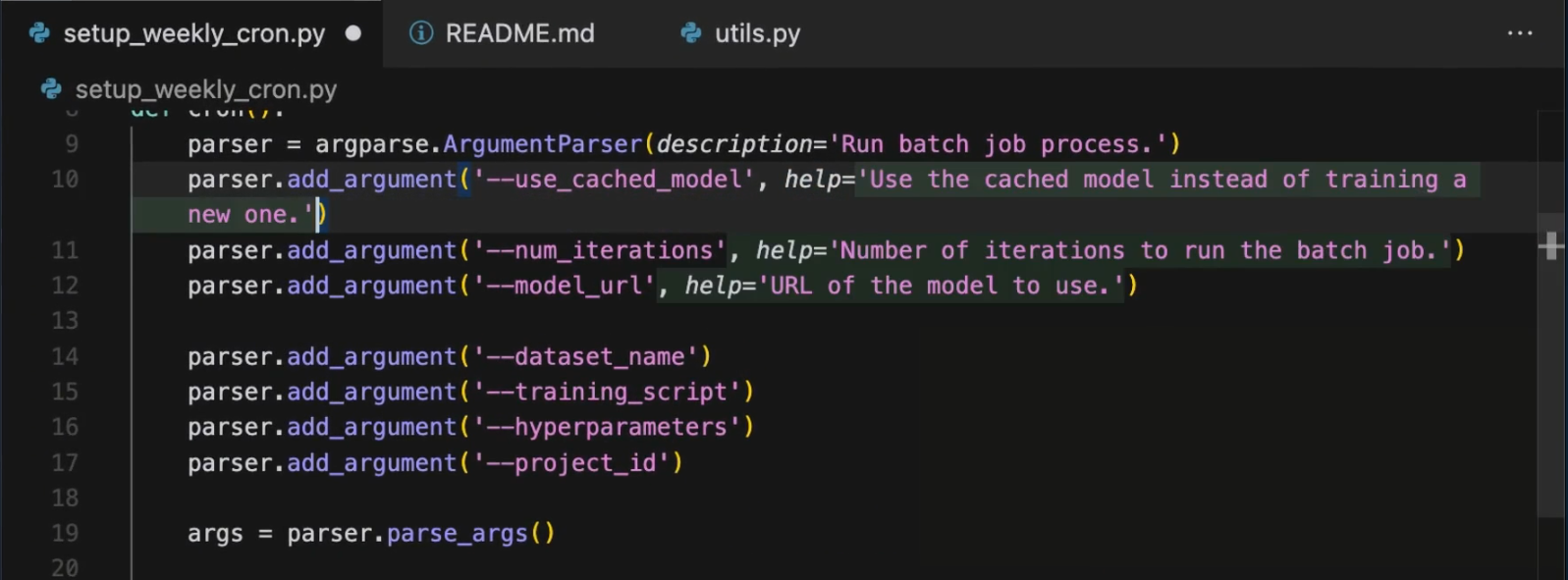}
    \caption{Left: Cursor analyzing code and suggesting fixes; Right: Cursor’s detailed code generation based on minimal input. This AI tool can significantly reduce debugging and coding time.}
    \label{fig:cursor-demo}
\end{figure}

\textbf{GitHub Copilot:}
Powered by OpenAI’s Codex model, \textbf{GitHub Copilot} is like having a co-pilot while you code. It predicts what you’re trying to write, suggests code completions, and even writes boilerplate code, freeing you from repetitive tasks. The simplicity of GitHub Copilot’s interface means that even novice developers can easily integrate it into their workflow. By hovering over the suggestion box, you can apply entire blocks of code in one click, increasing efficiency dramatically.

\begin{figure}[H]
    \centering
    \includegraphics[width=1.0\linewidth]{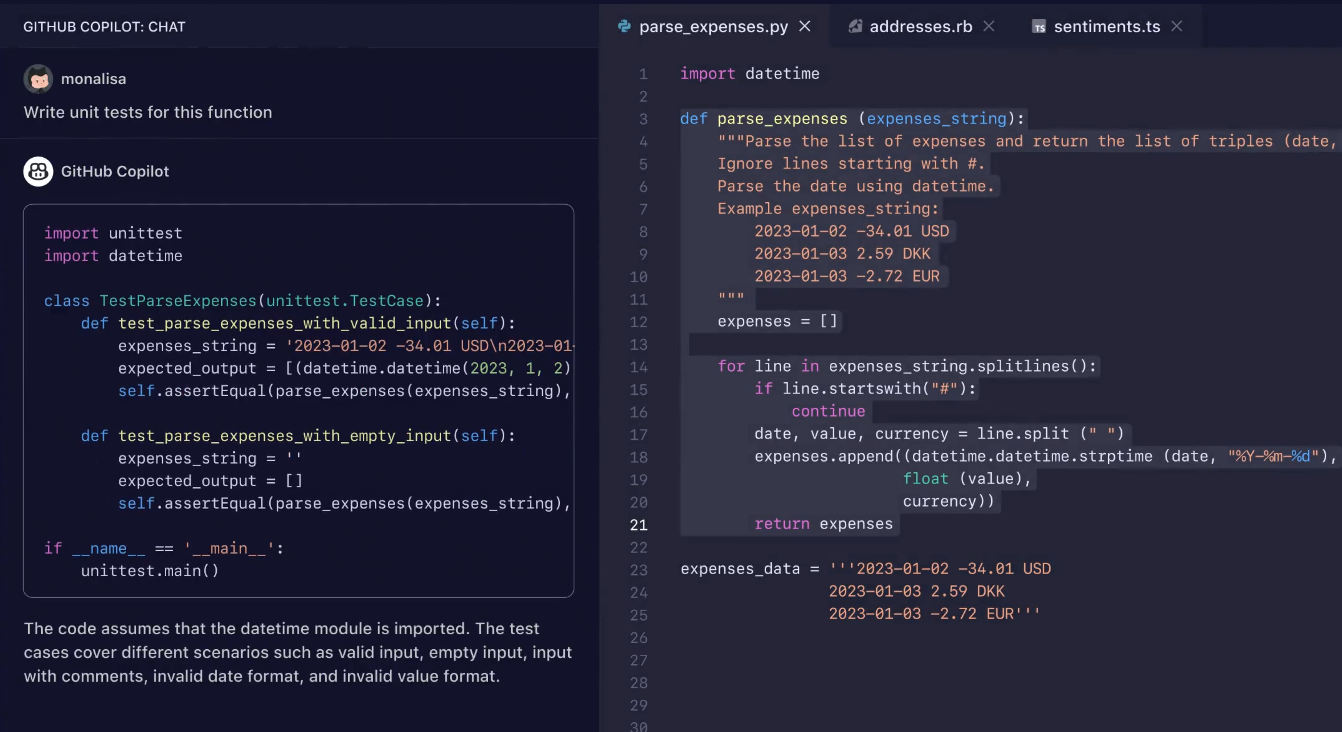}
    \caption{GitHub Copilot in action, suggesting code completions and writing functions automatically based on user prompts. This significantly reduces the time spent on writing repetitive code.}
    \label{fig:copilot-demo}
\end{figure}

\textbf{OpenDevin:}
Developed to provide smart insights, \textbf{OpenDevin} helps developers identify coding patterns, predict errors, and generate code documentation. Think of it as an intelligent assistant for ensuring that your code is not just functional but also optimized. OpenDevin’s intuitive dashboard offers a holistic view of your codebase, ensuring that every part of your project is analyzed for efficiency.

\textbf{Developers’ Perspectives:}
\begin{itemize}[noitemsep]
    \item Greg Brockman, co-founder of OpenAI, once said, "The future of programming is no programming." Tools like GitHub Copilot and Cursor are turning that vision into reality by doing the heavy lifting for developers.
    \item Andrej Karpathy, AI researcher and former director of AI at Tesla, has been vocal about the role AI will play in writing and maintaining code. "AI is fundamentally a new way of creating software."
\end{itemize}

\section{AI in Geospatial Intelligence}

\subsection{Geospy.ai: Geospatial AI}

Geospatial intelligence has become a critical part of environmental management, disaster response, and urban planning. \textbf{Geospy.ai}~\cite{geospy2023} uses AI to analyze satellite imagery, providing real-time insights into land use, forest cover, and even urban growth.

\textbf{What Geospy.ai brings to the table:}
Geospy.ai is built on convolutional neural networks (CNNs) designed to process vast amounts of geospatial data. It can detect changes in infrastructure, track deforestation, and provide critical insights that would otherwise take years to gather manually.

\textbf{Mark Zuckerberg’s view:} Mark Zuckerberg once said, "In the future, every company will use AI." Geospy.ai embodies that idea by integrating AI into industries that might not seem obvious, like environmental conservation and city planning.

\section{Summary}

From ChatGPT’s conversations to Geospy.ai’s real-time satellite analysis, AI is advancing faster than ever, transforming industries and simplifying complex tasks. Developers, designers, and everyday users are benefiting from AI's ability to automate, assist, and inspire. Whether you're using AI to write code with tools like Cursor and GitHub Copilot or enhancing images with Flux, the future is bright, and AI is leading the charge.

\chapter{An Introduction to 
Generative art: A Beginner's Exploration}

\section{What are Diffusion models?}

It is a type of probabilistic model that has been getting attention for its ability to generate high-quality, realistic images in the world of artificial intelligence (AI) and machine learning. In diffusion~\cite{ho2020denoisingdiffusionprobabilisticmodels}, a simple random noise distribution is gradually transformed into a meaningful structure or image over time, as a result of learning and refining. The diffusion process can be understood as the reverse of how physical particles spread or "diffuse" from an ordered state to a disordered one. As a core principle of AI, diffusion models are based on the reverse process of taking random noise and transforming it into detailed, structured data, such as an image.

In simple terms, diffusion models attempt to generate something meaningful (like an image) from randomness by learning how to reverse the natural process of noise. Through iterative refinement, they produce high-quality results that rival other AI methods, such as Generative Adversarial Networks (GANs)~\cite{goodfellow2014generativeadversarialnetworks}, but with greater stability and fewer generation problems.

\section{Stable Diffusion: A Game-Changer in AI Art Generation}

The Stable Diffusion~\cite{rombach2022highresolutionimagesynthesislatent} model is a state-of-the-art deep learning model designed for AI-based image generation. After its release in 2022, it was quickly recognized as a breakthrough in AI art. Previous models such as GANs and autoregressive models had shown promise in generating realistic images but often came with limitations such as instability, mode collapse (repeating the same type of output repeatedly), and limited generation capabilities.

Stable Diffusion, as its name suggests, brings greater stability and flexibility. It takes the diffusion process and applies it to generate high-resolution, coherent images from random noise, learning from massive datasets of images and their corresponding descriptions. What sets Stable Diffusion apart from other image-generation techniques is its ability to create highly detailed images, even when starting from simple prompts. It can generate artwork, landscapes, portraits, and even abstract imagery based on text descriptions, making it one of the most powerful tools in the AI art world.

The model works by iteratively refining noisy images, using a process called denoising diffusion probabilistic models (DDPM). Through this, it learns to reverse the noise and generate a clear, realistic image from random patterns, effectively "stabilizing" the generation process.

\begin{figure}[H]
    \centering
    \includegraphics[width=1.0\linewidth]{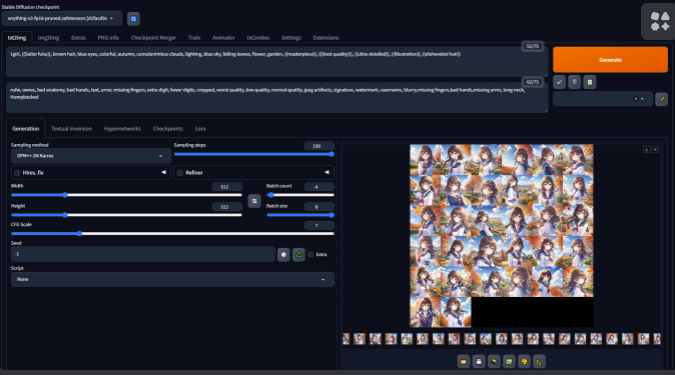}
    \caption{Snapshot of the image generation process using Stable Diffusion with WebUI. The settings and parameters used for generating high-quality images can be easily adjusted via the graphical interface.}
    \label{fig:image-generative-demo-0}
\end{figure}

\begin{figure}[H]
    \centering
    \includegraphics[width=0.2\linewidth]{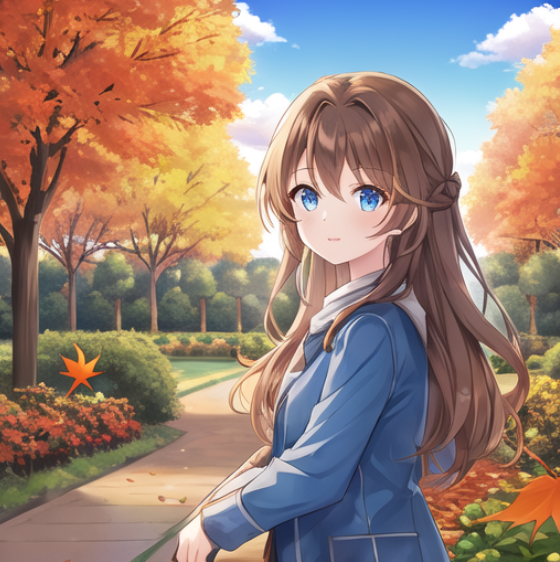}
    \includegraphics[width=0.2\linewidth]{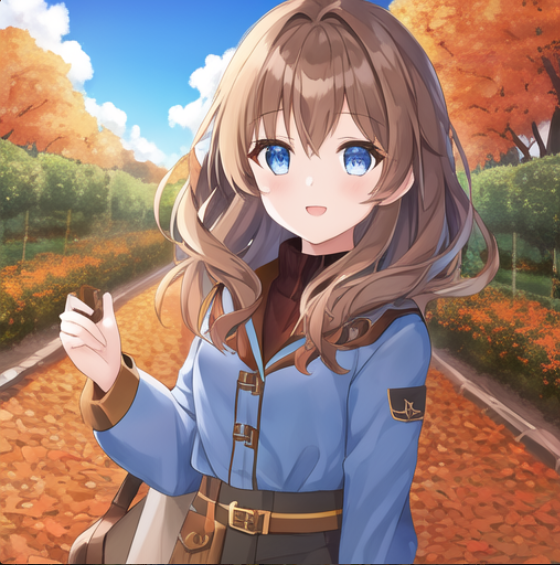}
    \includegraphics[width=0.2\linewidth]{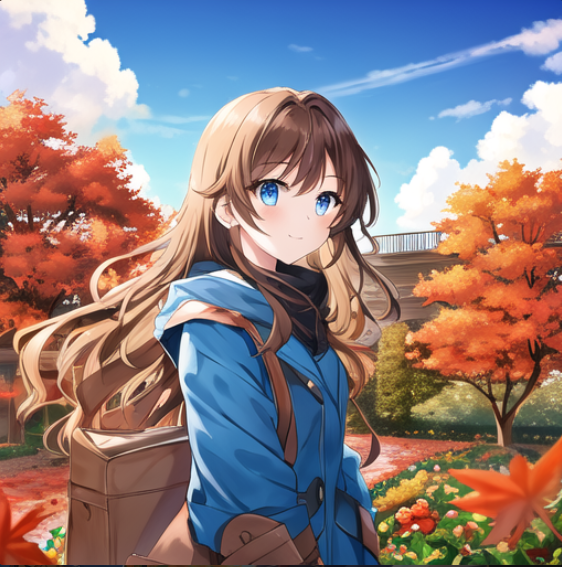}
    \includegraphics[width=0.2\linewidth]{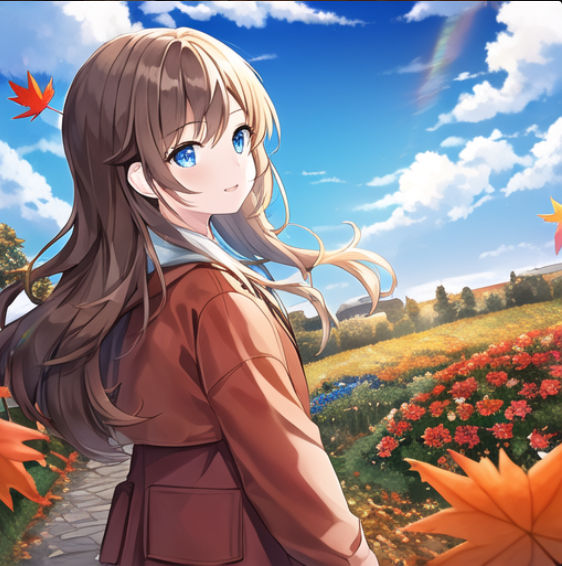}
    \caption{Animated sequence showing how Stable Diffusion refines the noise iteratively to create coherent and high-quality images. These frames capture various stages of the image-generation process.}
    \label{fig:animate-png-0-3}
\end{figure}

\begin{figure}[H]
    \centering
    \includegraphics[width=1.0\linewidth]{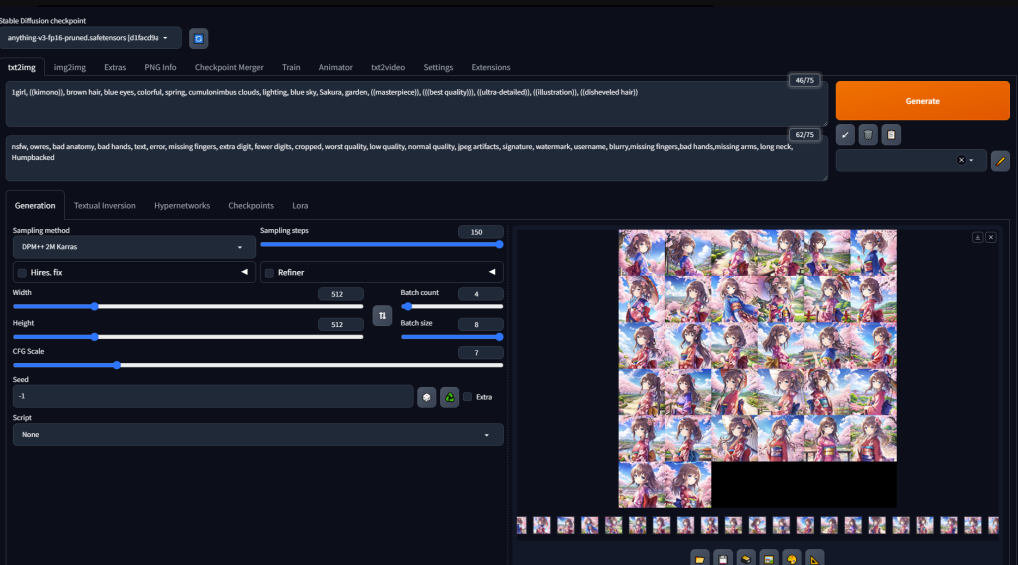}
    \caption{A demo of Stable Diffusion using the WebUI interface, showcasing its flexibility in creating multiple images at once with different prompts and settings.}
    \label{fig:image-generative-demo-1}
\end{figure}

\begin{figure}[H]
    \centering
    \includegraphics[width=0.2\linewidth]{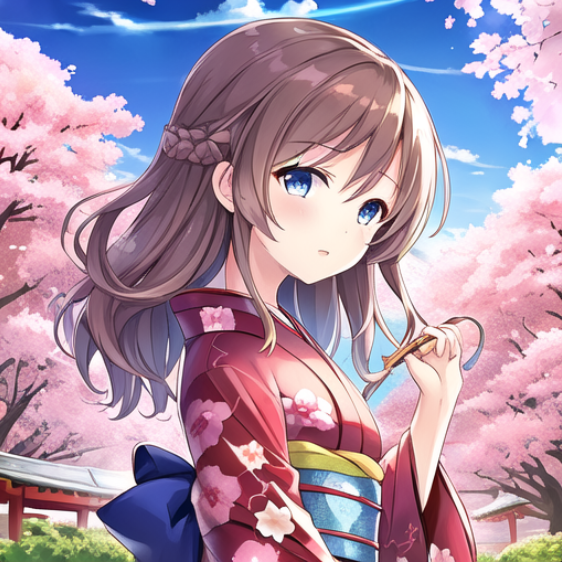}
    \includegraphics[width=0.2\linewidth]{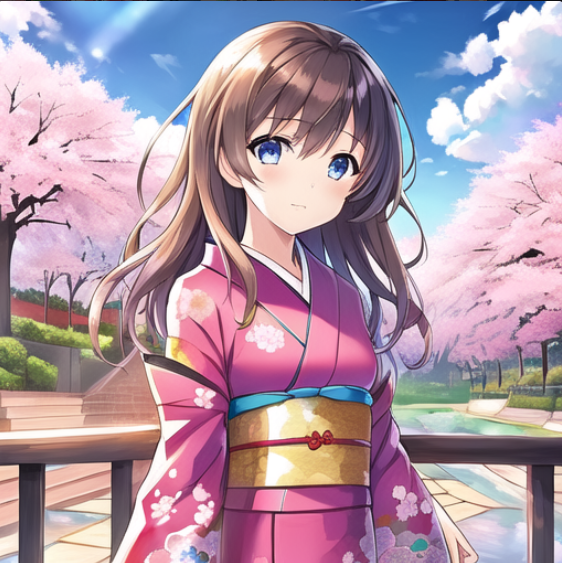}
    \includegraphics[width=0.2\linewidth]{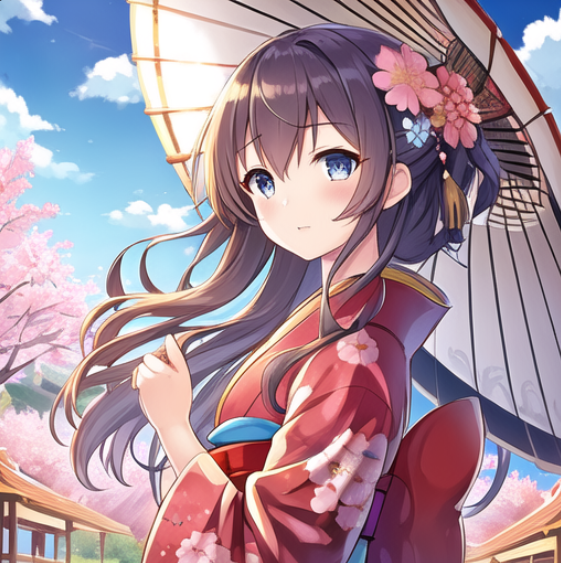}
    \includegraphics[width=0.2\linewidth]{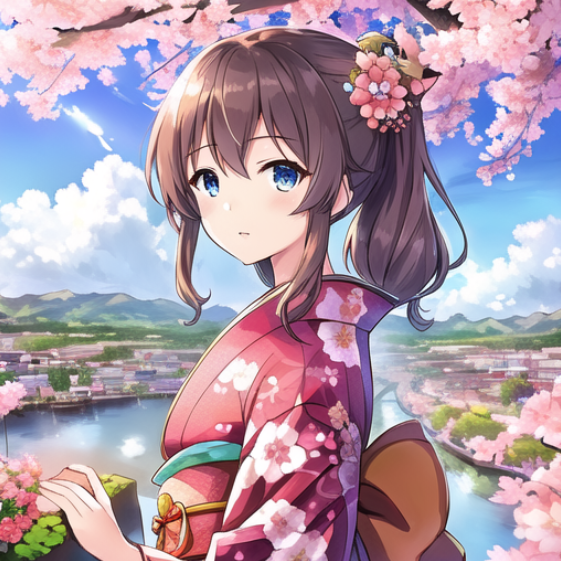}
    \caption{Another set of animated frames from the Stable Diffusion process, showing how different details emerge during the denoising process to create intricate, final images.}
    \label{fig:animate-png-4-7}
\end{figure}

\section{AI Art: The Fusion of Creativity and Technology}

AI art has revolutionized the way we think about creativity. Artists, designers, and hobbyists are now leveraging AI tools like Stable Diffusion to generate complex, imaginative artwork that previously required years of skill to produce. By feeding the model a textual description, users can see their ideas transformed into vivid imagery within minutes. This blend of human creativity and machine learning is shaping a new era of digital art, where AI acts as a collaborator, expanding the boundaries of artistic possibilities.

Artificial intelligence can create surreal, abstract compositions as well as hyper-realistic portraits and landscapes. The most exciting thing about this technology is that it allows anyone to realize their visions, regardless of their artistic background. As AI models become more refined, they also begin to learn individual artistic styles, giving creators greater control.

\section{Two Main Interfaces: WebUI vs. ComfyUI}

User interfaces (GUIs) are commonly used to make complex technology more accessible when using Stable Diffusion or other diffusion models to generate art. Two of the most popular interfaces are \textbf{WebUI} and \textbf{ComfyUI}. While both offer access to diffusion models for image generation, the needs and preferences of their users vary slightly.

\subsection{WebUI: The User-Friendly Option}

WebUI~\cite{automatic1111webui}, short for "Web User Interface," is one of the most popular platforms for interacting with Stable Diffusion. It’s known for being user-friendly and accessible to a broad audience, from beginners to more advanced users. WebUI is designed with simplicity in mind, providing an intuitive interface that allows users to easily input text prompts and generate images without needing extensive technical knowledge.

WebUI’s key features include:
\begin{itemize}[noitemsep,topsep=0pt]
    \item \textbf{Ease of Use}: A simple, clean design makes it easy to start generating images right away.
    \item \textbf{Customization}: Users can tweak various settings, such as image resolution, sampling steps, and denoising strength, to fine-tune their results.
    \item \textbf{Compatibility}: WebUI supports a variety of models, including the latest versions of Stable Diffusion, and integrates with plugins and extensions to offer more advanced functionality.
\end{itemize}

For many users, especially those just beginning to explore AI-generated art, WebUI is the ideal starting point. It provides the essential tools to generate stunning images quickly and with minimal technical barriers.

\subsection{ComfyUI: Advanced Workflow Customization}

While WebUI is excellent for ease of use, ComfyUI~\cite{comfyui} is geared toward users who want more control over the generation process. ComfyUI allows for advanced workflow customization, where users can break down the image generation process into a series of interconnected steps or "nodes," similar to how you would work in a visual programming environment.

Some key features of ComfyUI include:
\begin{itemize}[noitemsep, topsep=0pt]
    \item \textbf{Node-based Workflow}: Users can create custom pipelines for image generation by connecting different processing steps. This allows for greater experimentation and fine-tuning.
    \item \textbf{Flexibility}: ComfyUI is designed for users who want to experiment with different diffusion models and algorithms. It offers more granular control over the image creation process, enabling users to manipulate every aspect of the generation pipeline.
    \item \textbf{Powerful Customization}: This platform is best suited for users who are comfortable with more technical aspects of image generation, allowing them to test various parameters and methods to achieve unique and precise results.
\end{itemize}

While ComfyUI has a steeper learning curve compared to WebUI, it’s a powerful tool for those who want to dive deeper into the mechanics of diffusion models and experiment with different workflows.


\section{Comparison of WebUI and ComfyUI}

The following table provides a comparison of the pros and cons of both WebUI and ComfyUI:

\begin{table}[H]
\centering
\begin{tabular}{|>{\centering\arraybackslash}p{7cm}|>{\centering\arraybackslash}p{7cm}|}
\hline
\textbf{WebUI} & \textbf{ComfyUI} \\ \hline
\multicolumn{2}{|c|}{\textbf{Pros}} \\ \hline
\begin{itemize}[leftmargin=*, noitemsep]
    \item User-friendly interface, ideal for beginners
    \item Simple setup and minimal technical knowledge required
    \item Quick image generation with default settings
    \item Supports various models and plugins
    \item Customizable image parameters like resolution, sampling steps, and denoising
\end{itemize} 
& 
\begin{itemize}[leftmargin=*, noitemsep]
    \item Node-based workflow allows for precise control
    \item Highly customizable pipelines for advanced users on detailed control over each generation step
    \item Ideal for creating complex and specific workflows
    \item Workflows are transferable and can be finalized as a Python module to run in an automatic pipeline.
\end{itemize} \\ \hline
\multicolumn{2}{|c|}{\textbf{Cons}} \\ \hline
\begin{itemize}[leftmargin=*, noitemsep]
    \item Limited customization compared to more advanced tools
    \item May not offer enough control for experienced users
    \item Less suited for highly technical or experimental workflows
    \item The number of users is shrinking
\end{itemize}
& 
\begin{itemize}[leftmargin=*, noitemsep]
    \item Steeper learning curve, requires more technical understanding
    \item Can be overwhelming for beginners
    \item Longer setup and image generation process due to the complexity of the pipeline
    \item May require more time to fine-tune and test parameters
\end{itemize} \\ \hline
\multicolumn{2}{|c|}{\textbf{{\color{red}* It is strongly recommended that beginners focus their learning on ComfyUI}}} \\ \hline
\end{tabular}
\caption{Comparison of WebUI and ComfyUI: Pros and Cons}
\end{table}

Since WebUI is relatively simple and user-friendly, it's a great starting point for those new to AI-generated art. By exploring WebUI first, users can quickly grasp the basic concepts of diffusion models and image generation without being overwhelmed by complex features. This foundational experience will help in understanding the more advanced functionalities that ComfyUI offers later on, making the transition smoother and more rewarding when moving to a more powerful and customizable tool like ComfyUI.

\section{Getting Started with WebUI}

WebUI is a graphic interface that lets users play with diffusion-based generative models easily and prototyping quickly. This section will walk through the process of installing WebUI and generating your first image.

\subsection{Installation}

To get started with WebUI, you will first need to download and install it. Follow these steps to install WebUI on your system:

\begin{enumerate}
    \item \textbf{Download WebUI}: Head over to the official \href{https://github.com/AUTOMATIC1111/stable-diffusion-webui}{\textbf{WebUI GitHub repository}} and clone the repository using the following command:
    \begin{lstlisting}[style=cmd]
    git clone https://github.com/AUTOMATIC1111/stable-diffusion-webui
    \end{lstlisting}
    After the clone, you can follow the instructions in\href{https://github.com/AUTOMATIC1111/stable-diffusion-webui?tab=readme-ov-file#installation-and-running}{\textbf{Installation and Running}}. Keep in mind to match your operating system and graphics hardware.
    
    \item \textbf{Launch WebUI}: Once installed, run the following command to launch WebUI on Linux:
    \begin{lstlisting}[style=cmd]
    bash webui.sh
    \end{lstlisting}
    For Windows users, you can launch it using:
    \begin{lstlisting}[style=cmd]
    webui-user.bat
    Then the WebUI should be opening on your web browser.
    \end{lstlisting}
\end{enumerate}

\subsection{Generate Your First Image}

WebUI comes with a stable diffusion 1.5 checkpoint, to replace any version of the diffusion model, please follow this folder structure to put your diffusion generative model:

\begin{lstlisting}[style=cmd]
stable-diffusion-webui/
|-- models/                           # Contains subfolders for different model types
|   |-- Stable-diffusion/             # Stable Diffusion model checkpoint files
|   |-- Lora/                         # LoRA model files
|   |-- VAE/                          # Variational AutoEncoder models
|   |...
\end{lstlisting}

\begin{figure}
    \centering
    \includegraphics[width=0.9\linewidth]{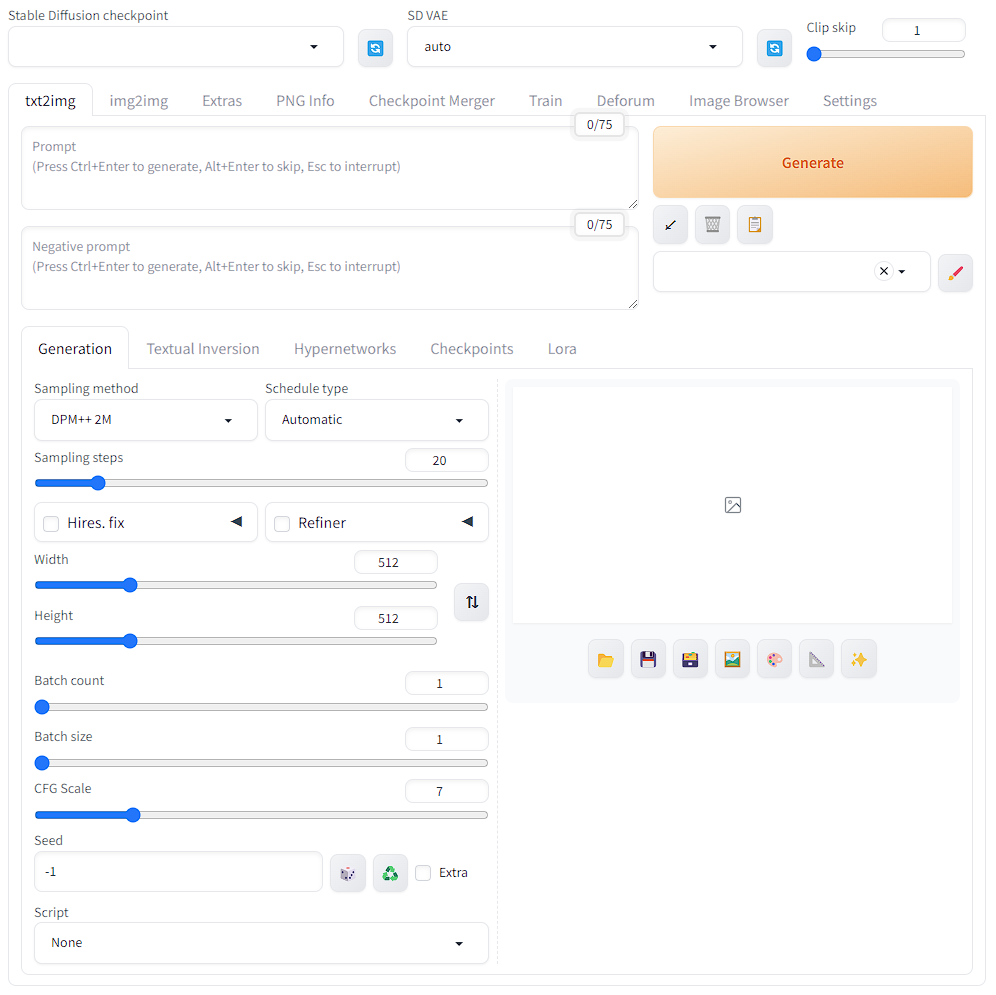}
    \caption{The WebUI Interface}
    \label{fig:enter-label1}
\end{figure}

\begin{figure}
    \centering
    \includegraphics[width=0.5\linewidth]{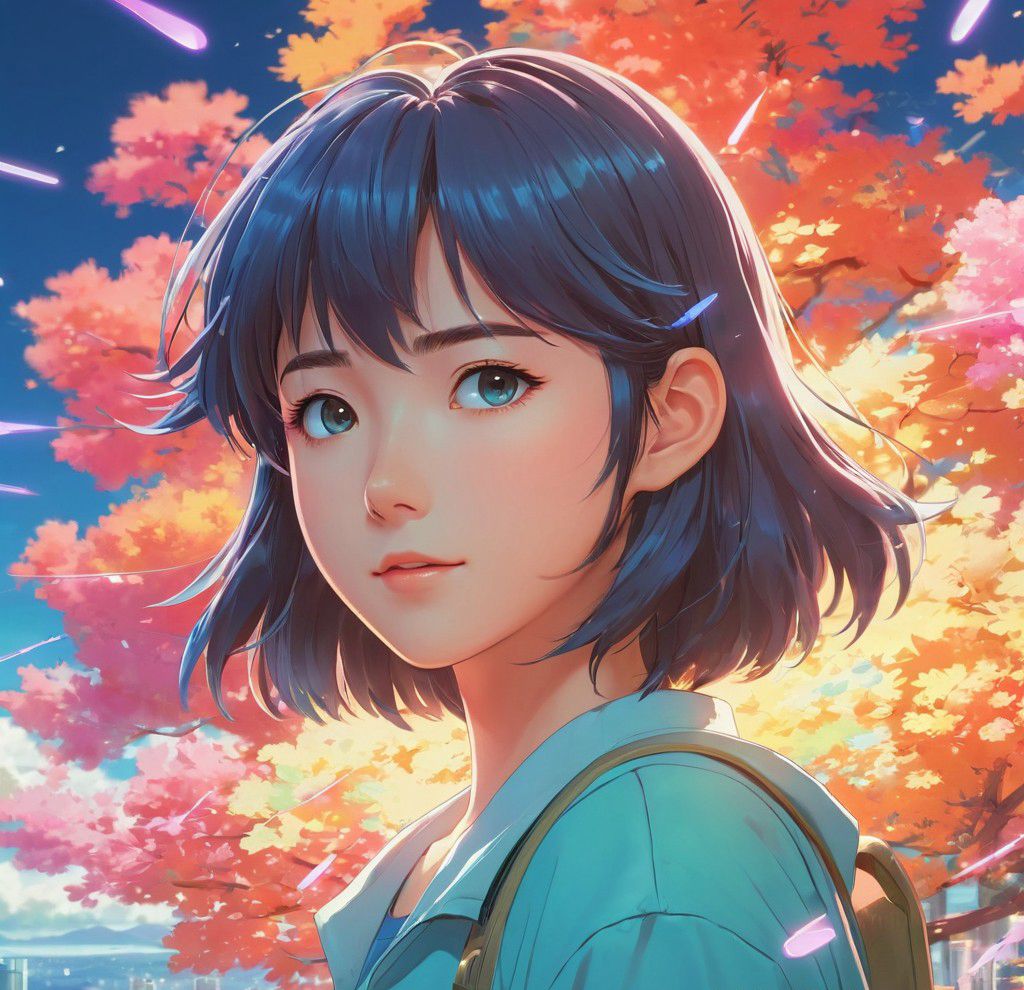}
    \caption{Prompt: A highly detailed matte painting of huge sector v treehouse by Studio Ghibli, Makoto Shinkai, by Artgerm, by Greg Rutkowski, volumetric lighting, octane render, 4K resolution, trending on artstation, masterpiece}
    \label{fig:enter-label2}
\end{figure}

Let's start with the basic usage: generate an image from text (txt2img tab).

\begin{itemize}
    \item \textbf{Stable Diffusion Checkpoint}: In this dropdown, select the stable diffusion model you want to use from the \texttt{\textcolor{orange}{Stable-diffusion}} folder.
    
    \item \textbf{Prompts}: Describe want you want to see in the \texttt{\textcolor{orange}{Prompt}} text box. Within \texttt{\textcolor{orange}{Negative Prompt}} text box, you can put things you don't want in the image. 
    
    \item \textbf{Width and Height}: This is the size of the output image. Different diffusion models might have different output resolutions, set at least one side to 512 for v1.x stable-diffusion models, and 1024 for stable-diffusion xl models. 

    \item \textbf{Sampling Methods}: Choose the algorithm for the denoising process. You can prototype from \texttt{\textcolor{orange}{DPM++ 2M Karras}}~\cite{lu2023dpmsolverfastsolverguided} since it provides a good balance between speed and quality.
    
    \item \textbf{Sampling Steps}: The number of steps for the denoising algorithm above. When you choose different denoising algorithms the best number of steps might change. Tweak the sampling steps with the methods.

    \item \textbf{Batch Count}: Number of times you run the image generation process.

    \item \textbf{Batch Size}: Number of images to generate per pipeline run.
\end{itemize}

Once you choose your preferred parameters, click \texttt{\textcolor{orange}{Generate}} then wait for your images to be generated. Remember, you can always leverage LLM to improve your prompts!

\begin{figure}
    \centering
    \includegraphics[width=0.5\linewidth]{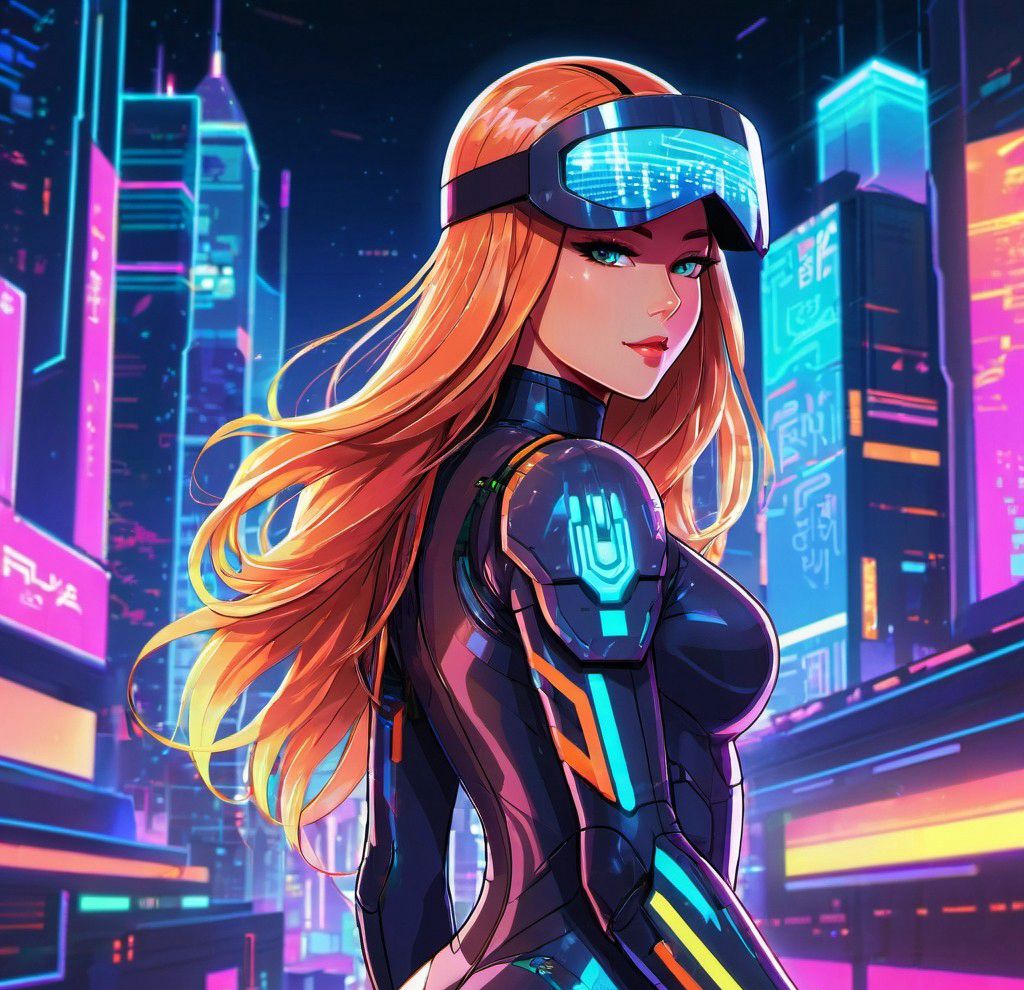}
    \caption{Prompt: A female character design, short hair wearing headphones, background in the city, Pixel style, transparent pvc jacket}
    \label{fig:enter-label3}
\end{figure}

\begin{figure}
    \centering
    \includegraphics[width=0.5\linewidth]{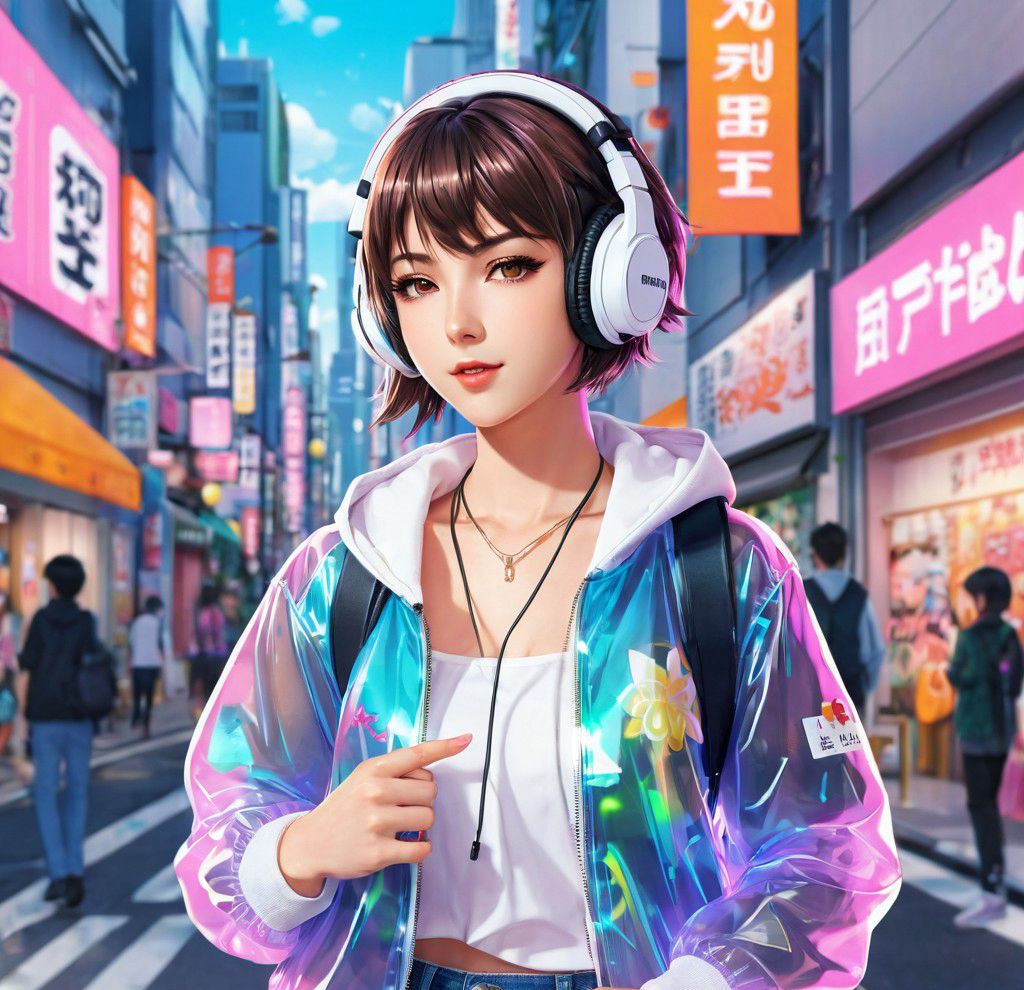}
    \caption{Prompt: A female character design, short hair wearing headphones, background in the city, Harajuku Style, transparent pvc jacket}
    \label{fig:enter-label4}
\end{figure}

\section{Getting Started with ComfyUI}

ComfyUI is a powerful node-based interface designed for generating images using AI models like Stable Diffusion. This section will walk through the process of installing ComfyUI, setting up a basic workflow, and generating your first image.

\subsection{Installation}

To get started with ComfyUI, you will first need to download and install it. Follow these steps to install ComfyUI on your system:

\begin{enumerate}
    \item \textbf{Download ComfyUI}: Head over to the official \href{https://github.com/comfyanonymous/ComfyUI}{\textbf{ComfyUI GitHub repository}} and clone the repository using the following command:
    \begin{lstlisting}[style=cmd]
    git clone https://github.com/comfyanonymous/ComfyUI
    \end{lstlisting}
    Alternatively, you can follow the \href{https://github.com/comfyanonymous/ComfyUI?tab=readme-ov-file#windows}{\textbf{Read Me page}} and download the standalone version for Windows and extract it.
    
    \item \textbf{Install Dependencies}: If you are using Python, navigate to the ComfyUI directory and install the required dependencies using:
    \begin{lstlisting}[style=cmd]
    pip install -r requirements.txt
    \end{lstlisting}
    \item \textbf{Launch ComfyUI}: Once installed, run the following command to launch ComfyUI:
    \begin{lstlisting}[style=cmd]
    python main.py
    \end{lstlisting}
    For Mac users, you can launch it using:
    \begin{lstlisting}[style=cmd]
    python main.py --force-fp16
    \end{lstlisting}
\end{enumerate}

\subsubsection{Try ComfyUI in Google Colab}
\href{https://colab.research.google.com/}{\textbf{Google Colab}}~\cite{googlecolab} (short for Collaboratory) is a free cloud-based service provided by Google which allows users to write and execute Python code in a Jupyter notebook environment. Colab particularly popular for data science, machine learning, and artificial intelligence projects because it provides access to powerful computational resources like GPUs and TPUs without needing a local setup.

To change your Google Colab notebook to use a GPU for free, follow these steps: In the Colab interface, click on \texttt{\textcolor{orange}{Runtime}} in the menu bar. Then, select \texttt{\textcolor{orange}{Change runtime type}} from the dropdown menu. In the dialog box that appears, under \texttt{\textcolor{orange}{Hardware accelerator}}, choose \texttt{\textcolor{orange}{GPU}} from the dropdown. After selecting it, click \texttt{\textcolor{orange}{Save}}. This will enable the GPU for your session, allowing for faster computation, particularly for tasks involving machine learning or deep learning models.

Use this \href{https://colab.research.google.com/github/comfyanonymous/ComfyUI/blob/master/notebooks/comfyui_colab.ipynb}{\textbf{Colab notebook}} to clone the \textbf{ComfyUI} repo and install required packages. The same notebook can be found inside the \href{https://github.com/comfyanonymous/ComfyUI/tree/master/notebooks}{\textbf{ComfyUI GitHub repository}}. This is a well-annotated notebook, and what each cell is used for should be well-annotated and self-explanatory, keep in mind that your free GPU session may subject to early termination.

\subsection{Install ComfyUI Manager}
The \href{https://github.com/ltdrdata/ComfyUI-Manager}{\textbf{ComfyUI-Manager}} is an extension designed to enhance the usability of ComfyUI. To install and use the ComfyUI Manager, follow these steps, which will guide you through the installation process and the basic usage of the manager. This will help you manage custom nodes in ComfyUI efficiently.

\begin{figure}
    \centering
    \includegraphics[width=0.9\linewidth]{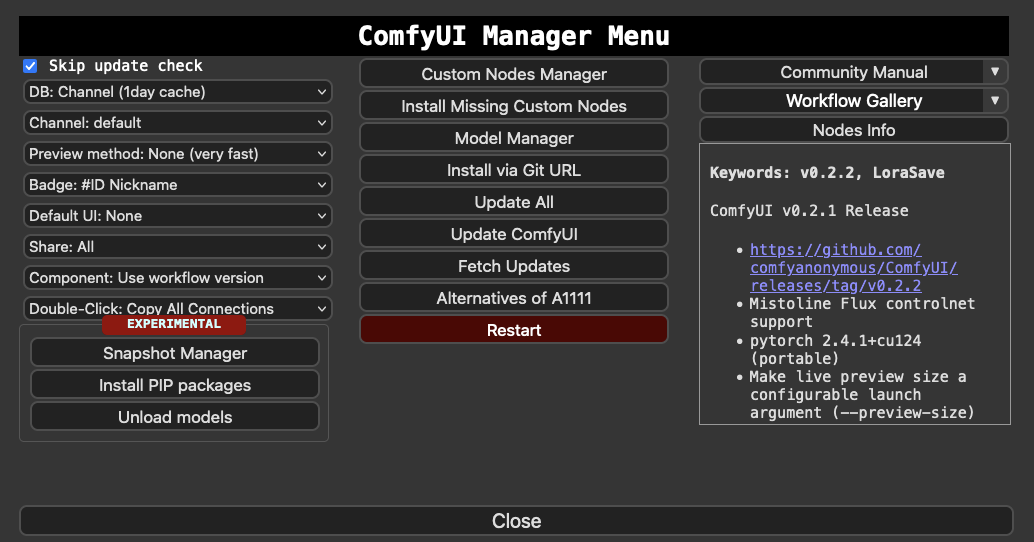}
    \caption{The ComfyUI Manager Menu}
    \label{fig:enter-label5}
\end{figure}

\begin{itemize}[leftmargin=*, noitemsep]
    \item First, make sure you have ComfyUI installed. If it's not installed yet, download it from the ComfyUI GitHub repository and follow the installation instructions.
    \item Open a terminal or command prompt, and navigate to the custom\_nodes directory in your ComfyUI folder: 
    \begin{lstlisting}[style=cmd] 
    cd path/to/ComfyUI/custom_nodes/ 
    \end{lstlisting}
    \item Clone the ComfyUI Manager repository into the custom\_nodes directory using: 
    \begin{lstlisting}[style=cmd] 
    git clone https://github.com/ltdrdata/ComfyUI-Manager.git 
    \end{lstlisting}
    \item Restart ComfyUI to recognize the ComfyUI Manager. This can be done by closing and reopening the application.
    \item Once ComfyUI restarts, you can find the ComfyUI Manager on the interface. Click \texttt{\textcolor{orange}{Manager}} in the main menu to open the management interface for custom nodes.
    \item In the ComfyUI Manager, you can use the \texttt{\textcolor{orange}{Install Custom Nodes}} option to automatically detect and install any missing nodes. This feature is especially helpful when uploading new workflows, as it simplifies the process by listing missing nodes in red.
    \item Additionally, the Manager allows you to update, disable, or remove nodes. Click on \texttt{\textcolor{orange}{Update All}} to refresh and ensure all nodes are up to date.
    \item If you come across conflicting nodes (shown in yellow), they can be disabled directly from the ComfyUI Manager interface to avoid issues.
\end{itemize}

\subsection{Setting Up Your Default Workflow}

The structure of the ComfyUI folder should look like the following:

\begin{lstlisting}[style=cmd]
ComfyUI/
|-- custom_nodes/                     # Directory for custom node files
|-- input/                            # Folder where uploaded assets are stored
|-- models/                           # Contains subfolders for different model types
|   |-- checkpoints/                  # Stable Diffusion model checkpoint files
|   |-- clip/                         # CLIP model files
|   |-- clip_vision/                  # CLIP Vision model files
|   |-- controlnet/                   # ControlNet models
|   |-- loras/                        # LoRA model files
|   |-- vae/                          # Variational AutoEncoder models
|   |-- unet/                         # UNet architecture files
|   `-- upscale_models/               # Models for upscaling images
|-- notebooks/                        # Jupyter Notebooks for Colab and remote executions
|-- output/                           # Generated images saved here
|-- config/                           # Configuration files for models and environment
|-- extra_model_paths.yaml.example    # Example config file for extra model paths
|-- comfyui.log                       # Runtime log file
|-- README_VERY_IMPORTANT.txt         # Instructions and usage details
`-- run_nvidia_gpu.bat                # Script to run ComfyUI on Nvidia GPUs
\end{lstlisting}

Once ComfyUI is installed, you can load a default workflow for generating images. Here's how:

\begin{enumerate}[noitemsep, topsep=0pt]
    \item \textbf{Download a Model}: Download a model checkpoint, such as Stable Diffusion v1.5, and place it in the directory \texttt{\textcolor{orange}{ComfyUI/models/checkpoints}}. \href{https://civitai.com}{\textbf{Civitai}}~\cite{civitai} is a community-driven platform for sharing and discovering custom models and assets used in AI-generated art, primarily focused on Stable Diffusion and other diffusion models.

    \item \textbf{Load the Default Workflow}: Open ComfyUI and click the \texttt{\textcolor{orange}{Load Default}} button to load the basic workflow. In the \texttt{\textcolor{orange}{Load Checkpoint}} node, select the model you just downloaded.
\end{enumerate}

\subsection{Generate Your First Image}

\begin{figure}
    \centering
    \includegraphics[width=1.0\linewidth]{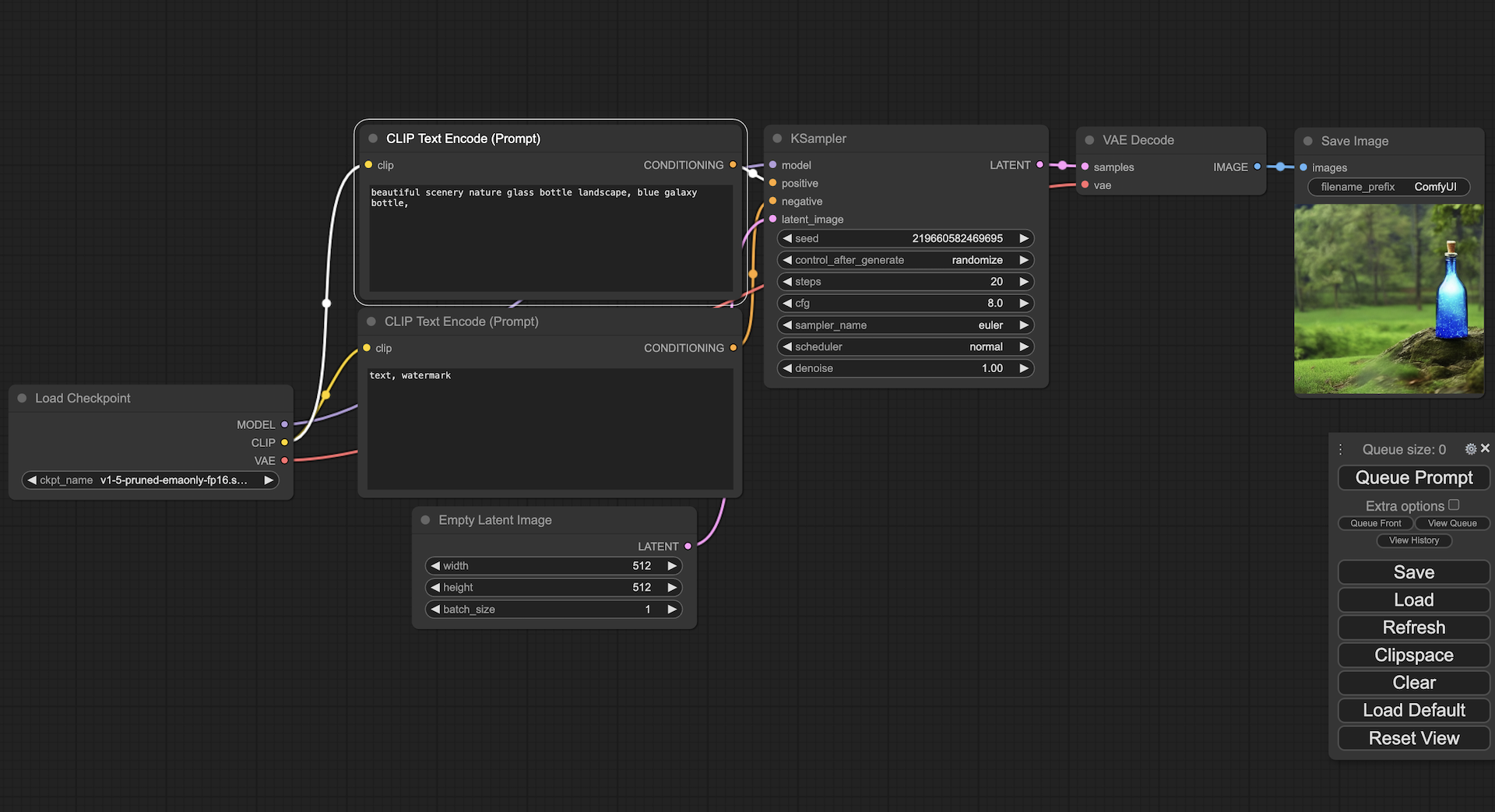}
    \caption{The ComfyUI interface with a simple image generation workflow}
    \label{fig:comfyui-interface}
\end{figure}

ComfyUI uses a node-based interface (see \textbf{Figure \ref{fig:comfyui-interface}}) where all operations are performed by connecting different nodes. To begin generating images:

\begin{itemize}
    \item \textbf{Add Nodes}: Right-click on the canvas and select \texttt{\textcolor{orange}{Add Nodes}}, or double left-click the canvas and search for a node. For image generation, important nodes include \texttt{\textcolor{orange}{Load Checkpoint}}, \texttt{\textcolor{orange}{CLIP Text Encode}}, and \texttt{\textcolor{orange}{KSampler}}.
    
    \item \textbf{Prompts and Parameters}: Set your prompts in the \texttt{\textcolor{orange}{CLIP Text Encode}} node (positive and negative prompt nodes are connected to \texttt{\textcolor{orange}{KSampler}} respectively) to guide image generation. Adjust parameters such as the number of diffusion \texttt{\textcolor{orange}{steps}} and the \texttt{\textcolor{orange}{CFG scale}} in the \texttt{\textcolor{orange}{KSampler}} node.
    
    \item \textbf{Queue and Generate}: Once the nodes are connected, click \texttt{\textcolor{orange}{Queue Prompt}} to generate the image. Experiment with different prompts to refine the results.
\end{itemize}

\subsection{Try Other Models}
To use other models in ComfyUI, follow these step-by-step instructions to ensure proper installation and workflow setup (this example will show you how to use the Anything XL model):

\begin{itemize}[leftmargin=*, noitemsep]

\item \textbf{Download the Anything XL Model}: First, download the Anything XL model from a platform like Hugging Face or CivitAI. Make sure the file is in .safetensors or .ckpt format.

\item \textbf{Place the Model in the Correct Directory}: Navigate to your ComfyUI installation directory, and place the downloaded Anything XL model in the ComfyUI/models/checkpoints folder.

\item \textbf{Launch ComfyUI}: Open ComfyUI by running the appropriate batch file depending on your system (e.g., \texttt{\textcolor{orange}{run\_nvidia\_gpu.bat}} for NVIDIA GPUs).

\item \textbf{Clear the Workspace}: Once ComfyUI opens, click \texttt{\textcolor{orange}{Clear}} to clear any pre-loaded workflows on the workspace. This will give you a clean slate to start from scratch.

\item \textbf{Load the Anything XL Model}: Right-click on the workspace, and navigate to \texttt{\textcolor{orange}{Add Node > Loaders > Load Checkpoint}}. From the options in the dropdown, select the Anything XL model you placed in the checkpoints folder earlier.

\item \textbf{Add the KSampler Node}: Right-click on the workspace again and go to \texttt{\textcolor{orange}{Add Node > Sampling > KSampler}}. This node is essential for generating images from random noise.

\item \textbf{Connect the Model and KSampler}: Connect the output from the \texttt{\textcolor{orange}{Load Checkpoint}} node (MODEL) to the input of the \texttt{\textcolor{orange}{KSampler}} node (model). This links the model to the sampling process.

\item \textbf{Set Up Prompts (Positive/Negative}): To tell the model what to generate, you need to add prompt nodes. Right-click the workspace and select \texttt{\textcolor{orange}{Add Node > Conditioning > CLIP Text Encode (Prompt)}}. This will be your positive prompt node. Repeat this step for the negative prompt. Link the positive prompt's output to the \texttt{\textcolor{orange}{conditioning}} input of the KSampler node. Similarly, link the negative prompt's output to the \texttt{\textcolor{orange}{conditioning (negative)}} input on the KSampler.

\item \textbf{Add Empty Latent Image}: Right-click on the workspace and select \texttt{\textcolor{orange}{Add Node > latent > Empty Latent Image}}. This node will act as the starting point for the image generation. Connect its output to the latent image input on the KSampler.

\item \textbf{VAE Decode}: The generated latent image needs to be decoded into a viewable format. Right-click and select \texttt{\textcolor{orange}{Add Node > latent > VAE Decode}}. Connect the KSampler's latent output to the \texttt{\textcolor{orange}{samples}} input on the VAE Decode node.

\item \textbf{Save the Image}: Add the final node by selecting \texttt{\textcolor{orange}{Add Node > image > Save Image}}. This node will save the generated image to your output folder. Connect the output of the VAE Decode node to the Save Image node.

\item \textbf{Adjust Parameters}: You can now fine-tune your settings. Adjust the image size in the Empty Latent Image node to match Anything XL’s training resolution (1024x1024). You can also modify the sampling steps and CFG scale in the KSampler node to refine the output.

\item \textbf{Generate the Image}: Once all nodes are connected, and parameters are set, click \texttt{\textcolor{orange}{Queue Prompt}} to generate the image. The process might take a few minutes, depending on your system and settings.

\item \textbf{Review and Save}: After the image is generated, it will be saved automatically in the output folder, but you can preview and make any additional adjustments to your prompts or settings if needed. 
\end{itemize}

\begin{figure}
    \centering
    \includegraphics[width=0.66\linewidth]{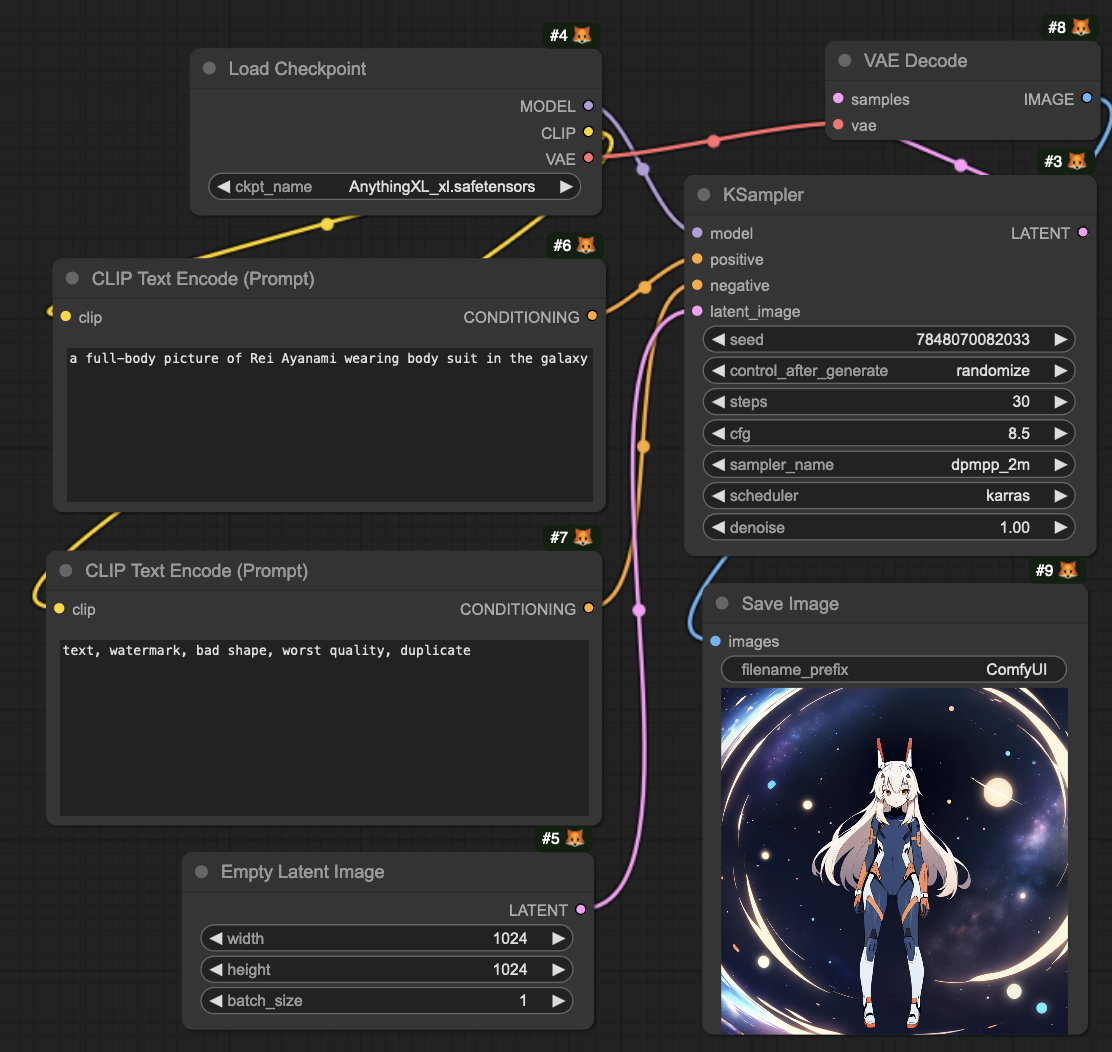}
    \caption{The ComfyUI interface with a simple image generation workflow}
    \label{fig:anythingxl-comfy}
\end{figure}

When using the Anything XL model with ComfyUI, configuring the KSampler node correctly quite important. Here's an overview of the recommended KSampler settings, along with an explanation of their functions:

\begin{itemize}[leftmargin=*, noitemsep] 
\item \textbf{Steps}: The number of sampling steps. For Anything XL, you typically want to set this to around 20-40 steps. Higher steps usually result in better image quality but will take longer to process. For quick previews, you might start with a lower value like 20, but for high-quality final outputs, going up to 40 or even 50 steps can produce more refined images.

\item \textbf{CFG} (Classifier-Free Guidance) Scale: This parameter controls how much the model adheres to the text prompts. For Anything XL, a good starting value is 7-12. Lower values like 7 will give more creative freedom to the model, while higher values like 12 will enforce stricter adherence to the prompt. Start with 7.5 for balanced results, and adjust based on your needs.

\item \textbf{Seed}: This is used for deterministic image generation. By setting a specific seed, you can regenerate the exact same image. If you want random images each time, leave this as -1 (which stands for random seed).

\item \textbf{Sampler Algorithm}: KSampler offers different algorithms to control how the noise is removed over time. For Anything XL, Euler A or DPM++ 2M Karras are popular choices for producing sharper and more detailed results. Euler A is faster but might not be as precise as DPM++ 2M Karras. If you prioritize speed, go with Euler A, otherwise try DPM++ 2M Karras for high-quality outputs.


\item \textbf{Denoising Strength}: This parameter is especially useful if you're using techniques like Img2Img or refining an existing image. A value around 0.7-0.9 works well for generating detailed results. For text-to-image generation, you usually leave this at 1.

\end{itemize}

You can checkout the example in \textbf{Figure \ref{fig:anythingxl-comfy}}


\subsection{Advanced Features}


%
ComfyUI also supports more advanced workflows, such as image-to-image (Img2Img) generation, where you can input an existing image to modify. Additionally, you can explore various node shortcuts like \texttt{CTRL + drag} to select multiple nodes, or \texttt{SHIFT + drag} to move them.

This setup will allow you to get started with ComfyUI and generate your first batch of AI-powered images.

\section{Summary}

As we explored in this chapter, diffusion models and their roles in AI-driven image generation were the foundations of generative art. Diffusion models have emerged as a powerful method for transforming random noise into coherent, high-quality images, providing a stable alternative to earlier techniques like Generative Adversarial Networks (GANs). The development of models like Stable Diffusion has revolutionized AI art generation, enabling artists and hobbyists alike to create detailed and vivid imagery from simple text prompts.

We also examined two main interfaces—WebUI and ComfyUI—through which users can interact with diffusion models. WebUI, with its user-friendly design, offers an accessible platform for beginners to generate AI art with minimal technical knowledge. On the other hand, ComfyUI is tailored for more advanced users, providing granular control over the generation process through a node-based workflow, allowing for greater experimentation and customization.

The chapter showed how AI art merges creativity and technology in a dynamic way. As AI continues to push the boundaries of what's possible in the world of art, even people without traditional artistic backgrounds can explore new realms of digital creativity with tools like Stable Diffusion.









\appendix
\chapter{Image Demos}

This appendix includes various image demos that illustrate different aspects of the content discussed in the book.

\begin{figure}[h]
    \centering
    \includegraphics[width=1\linewidth]{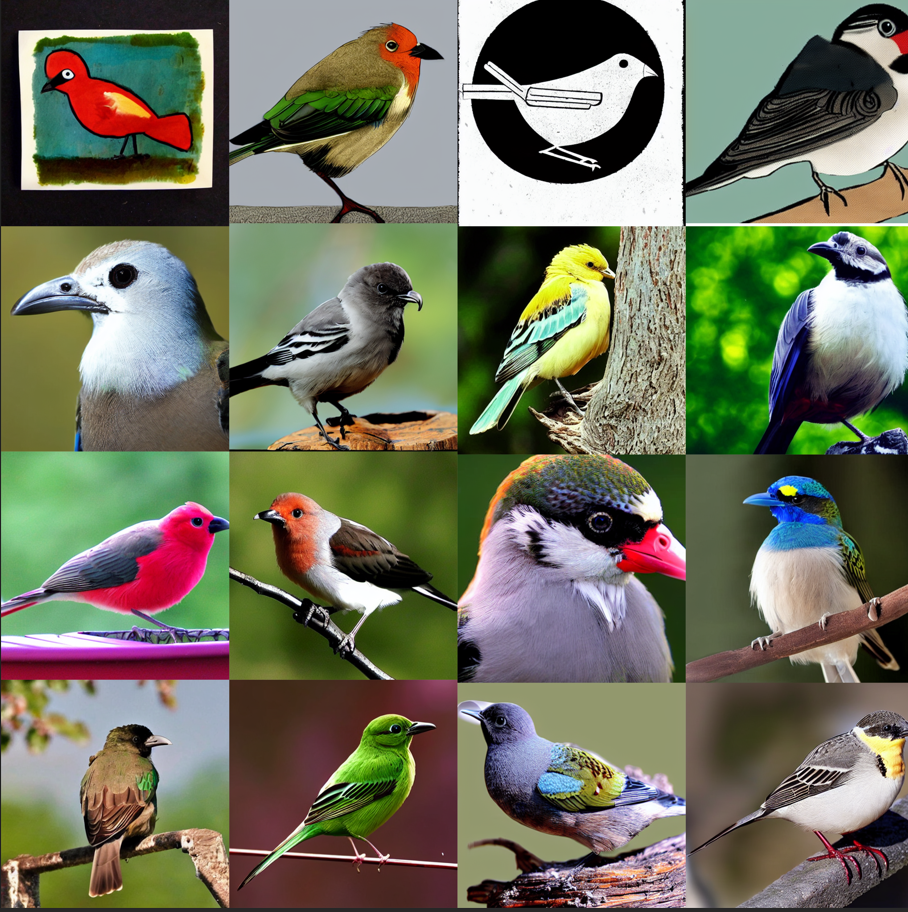}
    \caption{Bird Demo. These images showcase generative models creating realistic representations of animals.}
   
\end{figure}
\begin{figure}[h]
    \centering
    \includegraphics[width=1\linewidth]{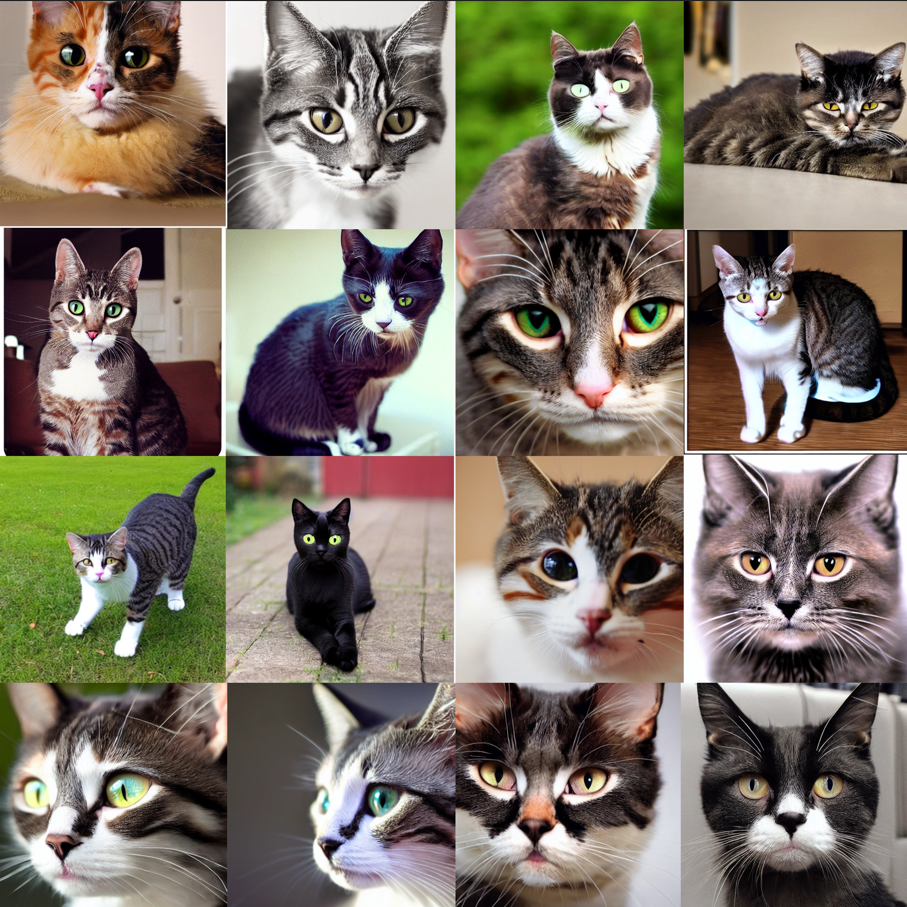}
    \caption{Cat Demo. These images showcase generative models creating realistic representations of animals.}
    
\end{figure}

\begin{figure}[h]
    \centering
    \includegraphics[width=1\linewidth]{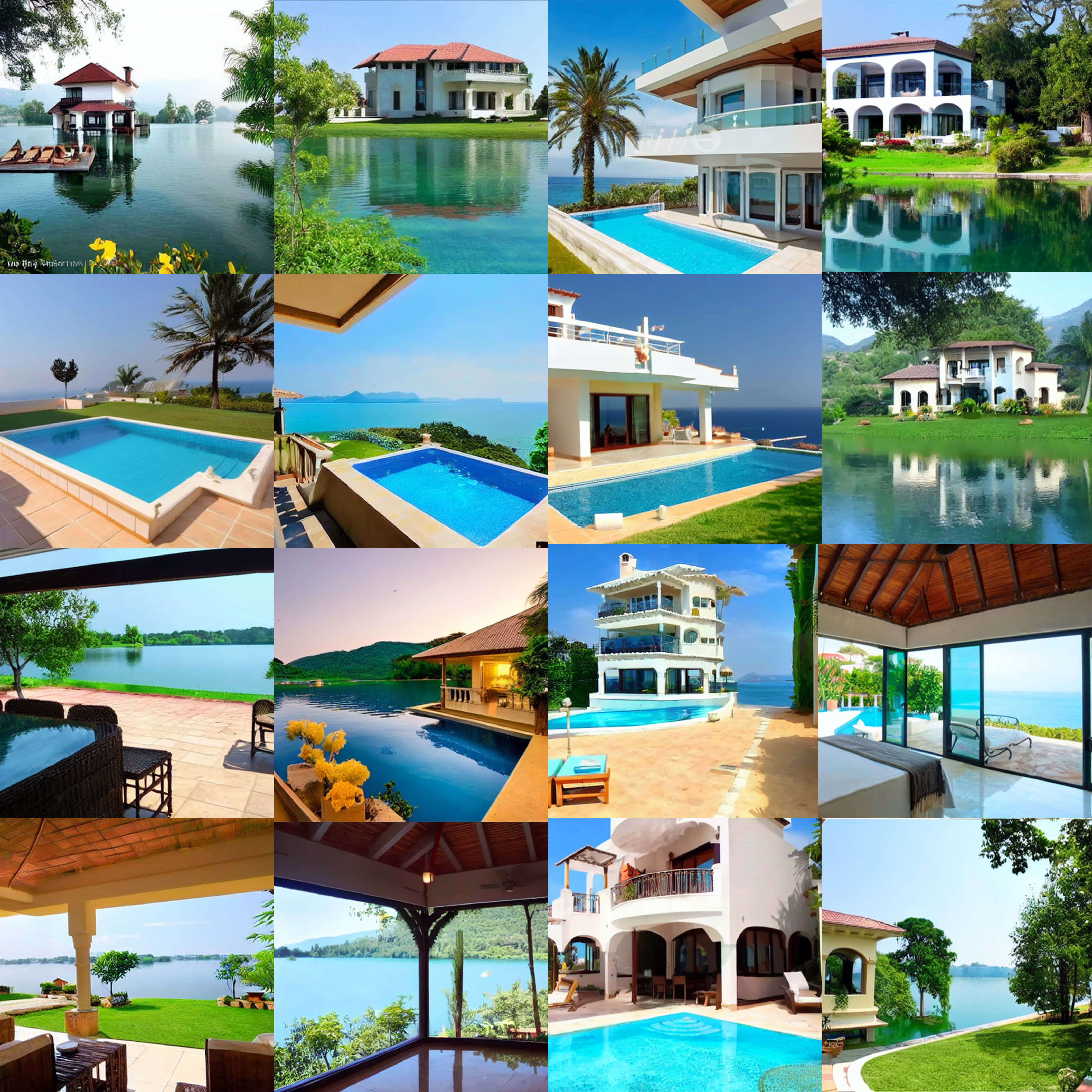}
    \caption{House Demo. Examples of generative models producing architectural and still life images via prompt: Beautiful lake view villa}
    
\end{figure}
\begin{figure}[h]
    \centering
    \includegraphics[width=1\linewidth]{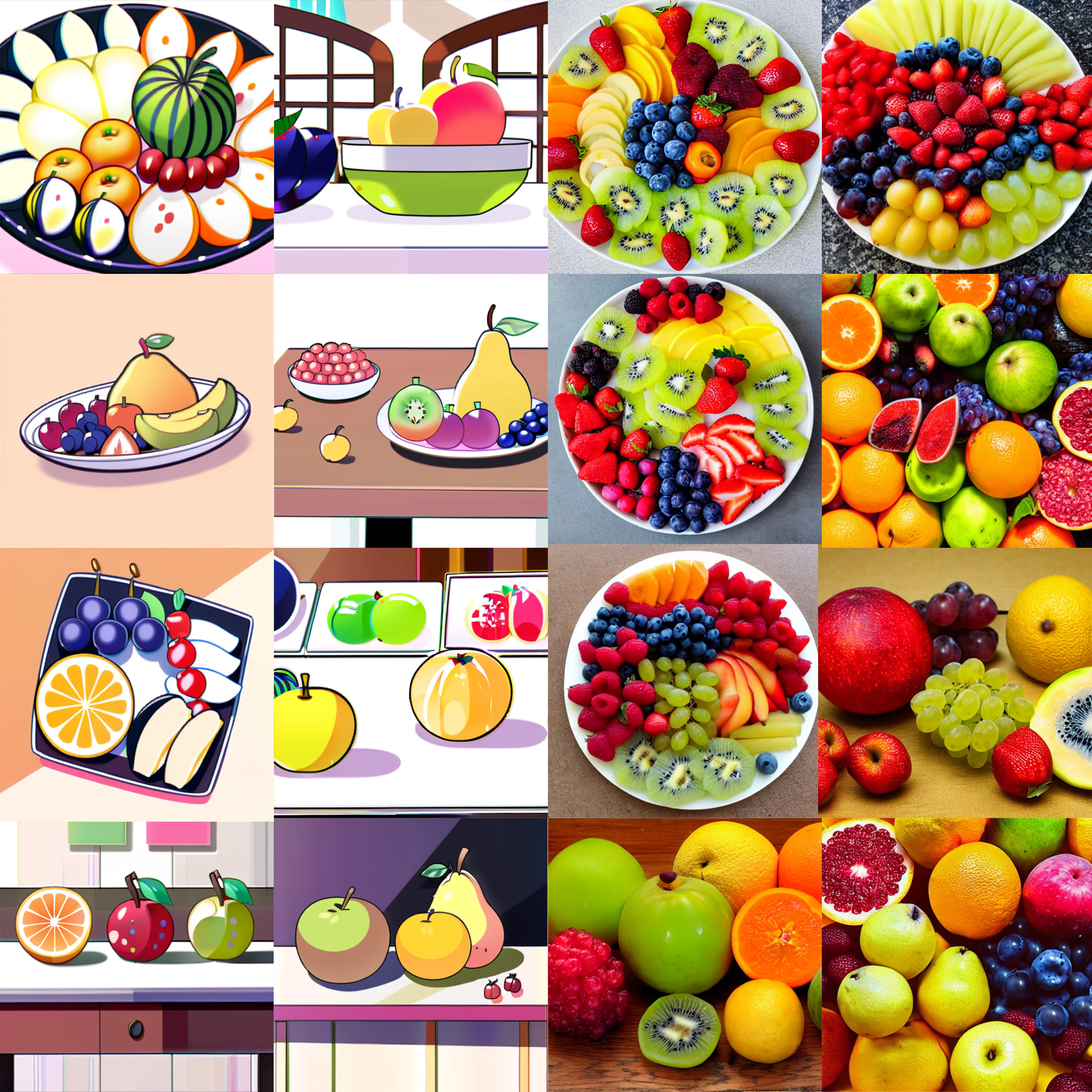}
    \caption{ Fruit Demo. Examples of generative models producing architectural and still life images via prompt: Various fruits on the table and Fruit platter}
    
\end{figure}

\begin{figure}[h]
    \centering
    \includegraphics[width=1\linewidth]{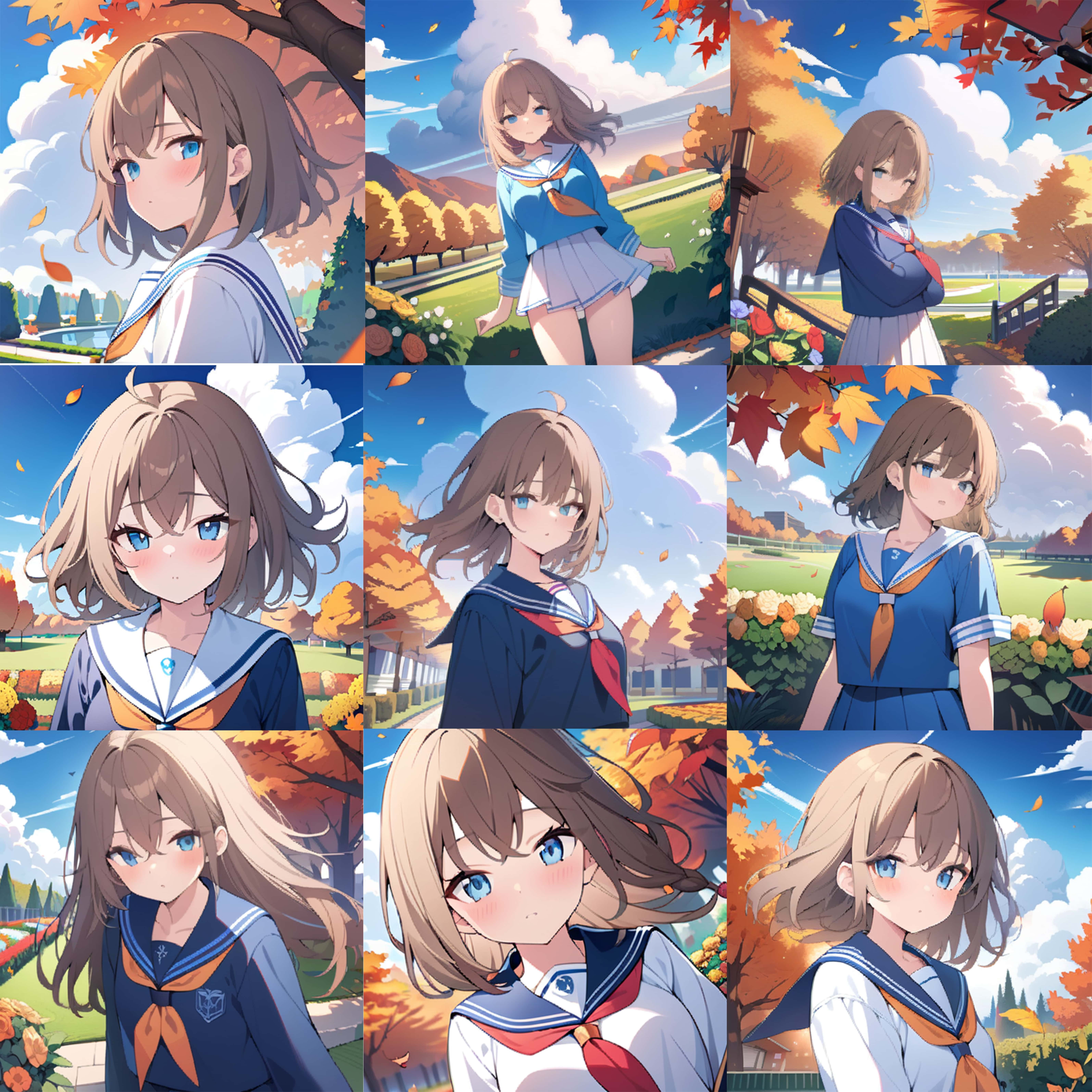}

    \caption{ Animation Demo 1. These frames illustrate how the model generates animations.Promt:1girl, ((Sailor fuku)), brown hair, blue eyes, colorful, autumn, cumulonimbus clouds, lighting, blue sky, falling leaves, flower, garden, ((masterpiece)), (((best quality))), ((ultra-detailed)), ((illustration)), ((disheveled hair))}
   
\end{figure}

\begin{figure}[h]
    \centering
  
    \includegraphics[width=1\linewidth]{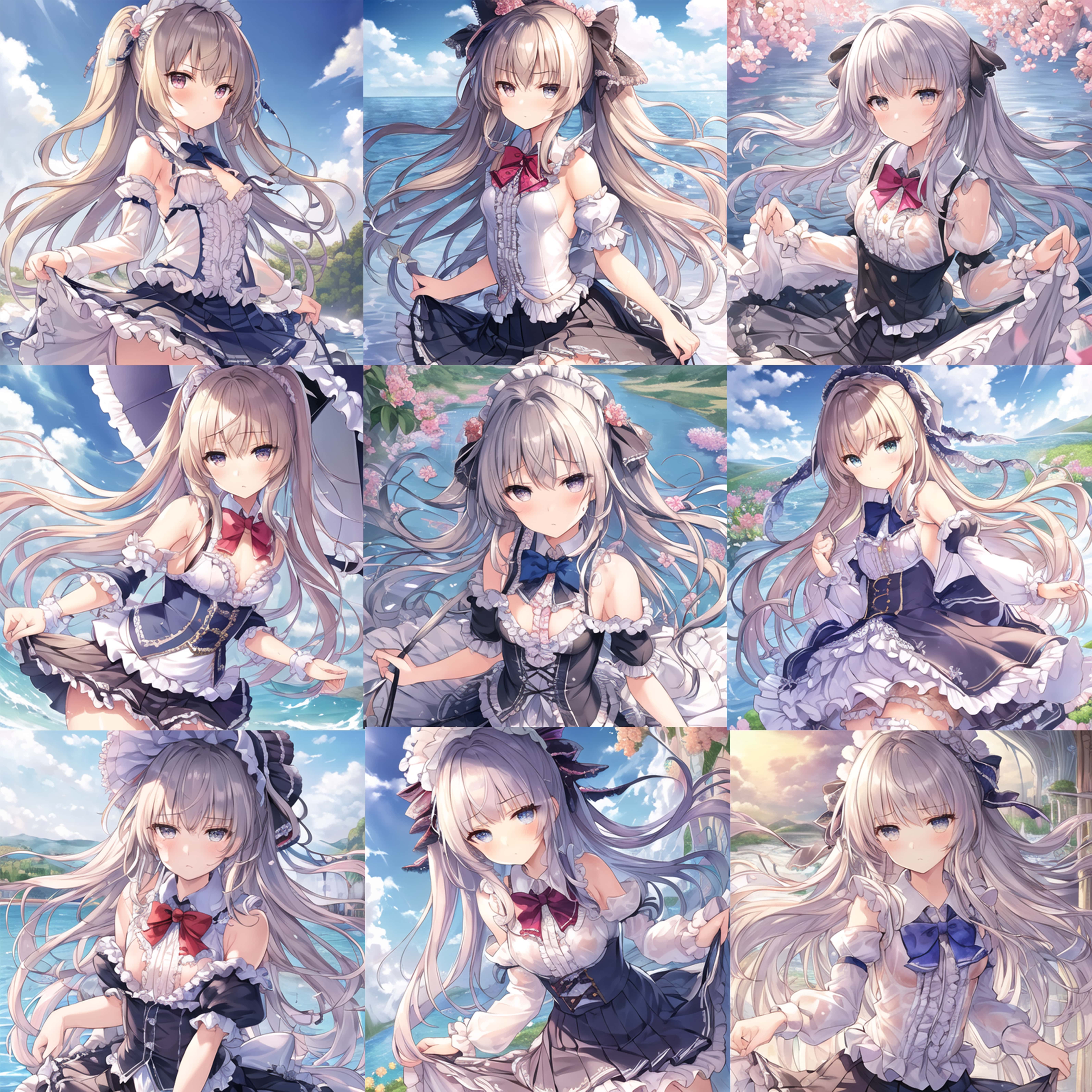}
    \caption{Animation Demo 2. These frames illustrate how the model generates animations. prompt: 1girl, brown hair, green eyes, colorful, autumn, cumulonimbus clouds, lighting, blue sky, falling leaves, garden}

\end{figure}

\begin{figure}[h]
    \centering
    \includegraphics[width=1\linewidth]{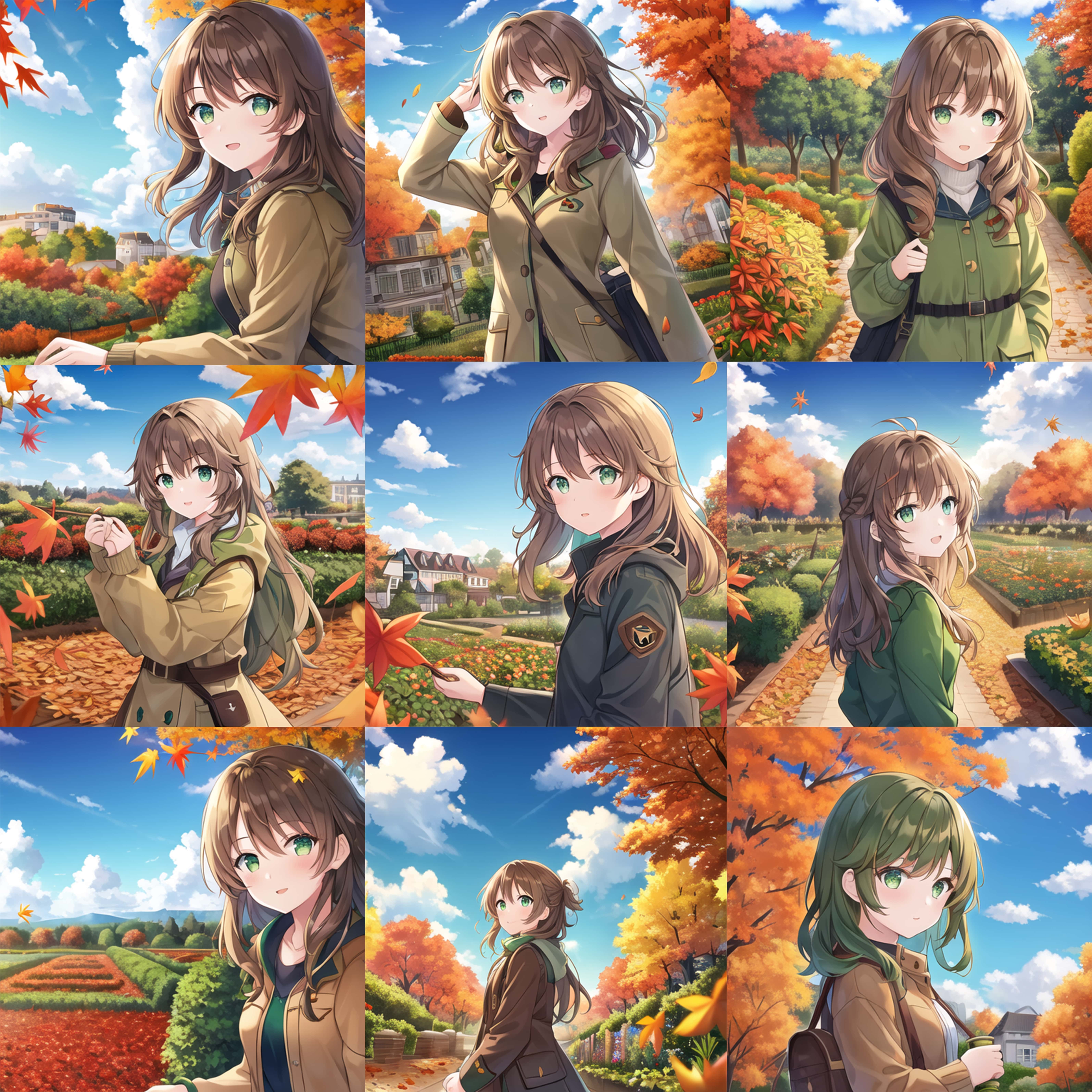}

    \caption{Animation Demo 3.  Further frames from the animation generation process. Prompt: 1girl, brown hair, blue eyes, colorful, autumn, cumulonimbus clouds, lighting, blue sky, falling leaves, garden}

\end{figure}
\begin{figure}[h]
    \centering

    \includegraphics[width=1\linewidth]{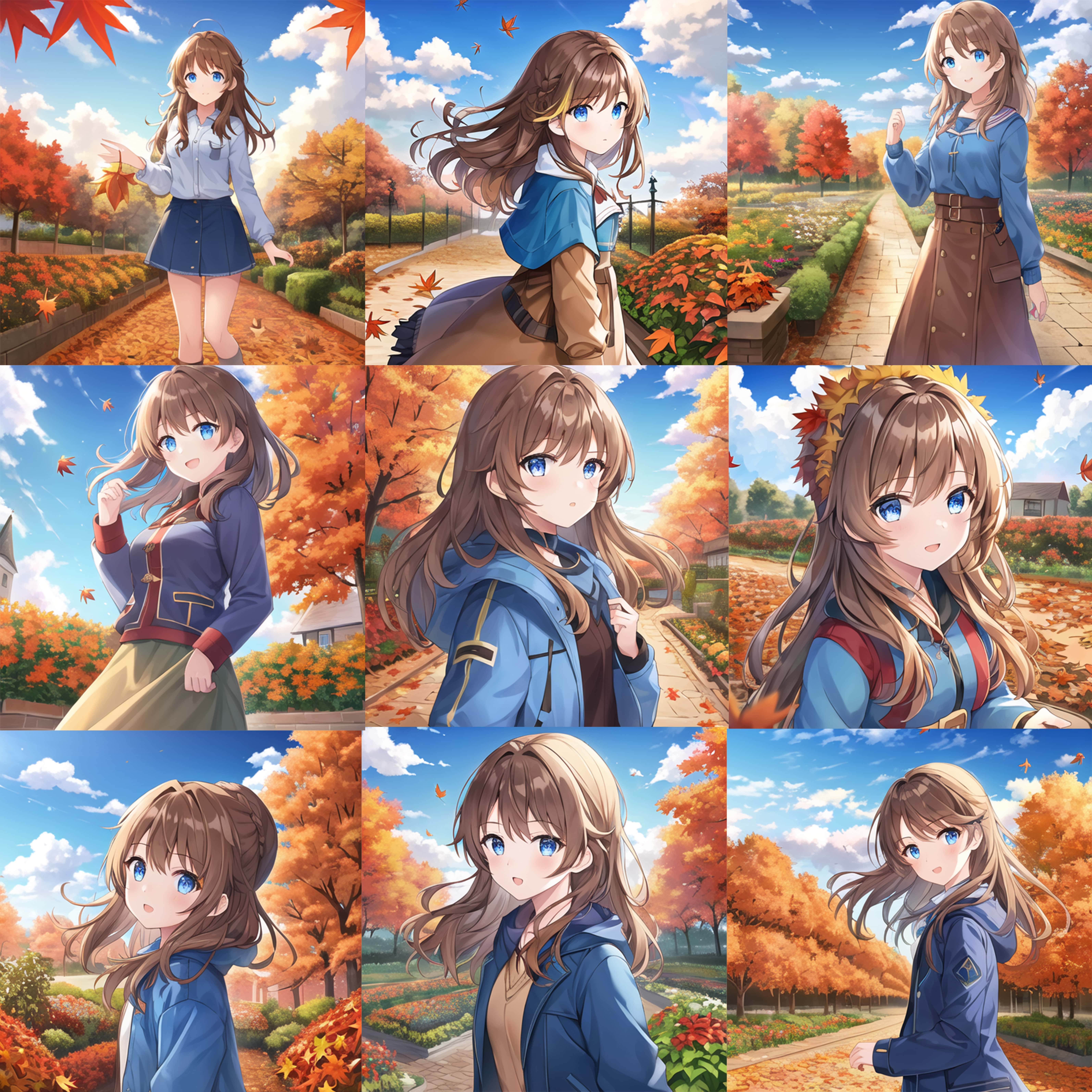}
    \caption{ Animation Demo 4. Further frames from the animation generation process.Prompt:1girl, brown hair, blue eyes, colorful, autumn, cumulonimbus clouds, lighting, blue sky, falling leaves, garden
}

\end{figure}

\begin{figure}[h]
    \centering
    \includegraphics[width=1\linewidth]{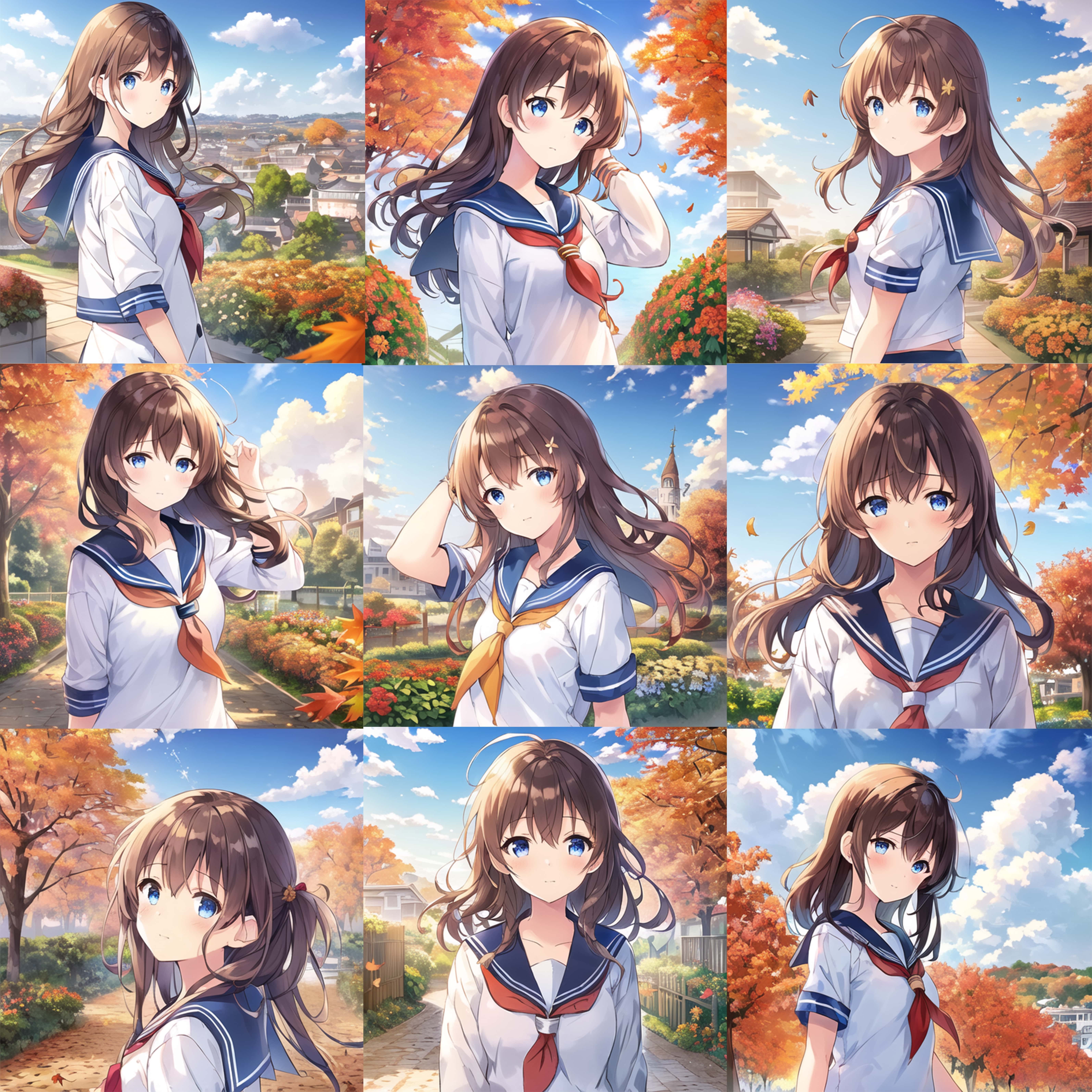}

     \caption{ Animation Demo 4. Further frames from the animation generation process.Prompt:1girl, ((Sailor fuku)), brown hair, blue eyes, colorful, autumn, cumulonimbus clouds, lighting, blue sky, falling leaves, flower, garden, ((masterpiece)), (((best quality))), ((ultra-detailed)), ((illustration)), ((disheveled hair))
}

\end{figure}

\begin{figure}[h]
    \centering

    \includegraphics[width=1\linewidth]{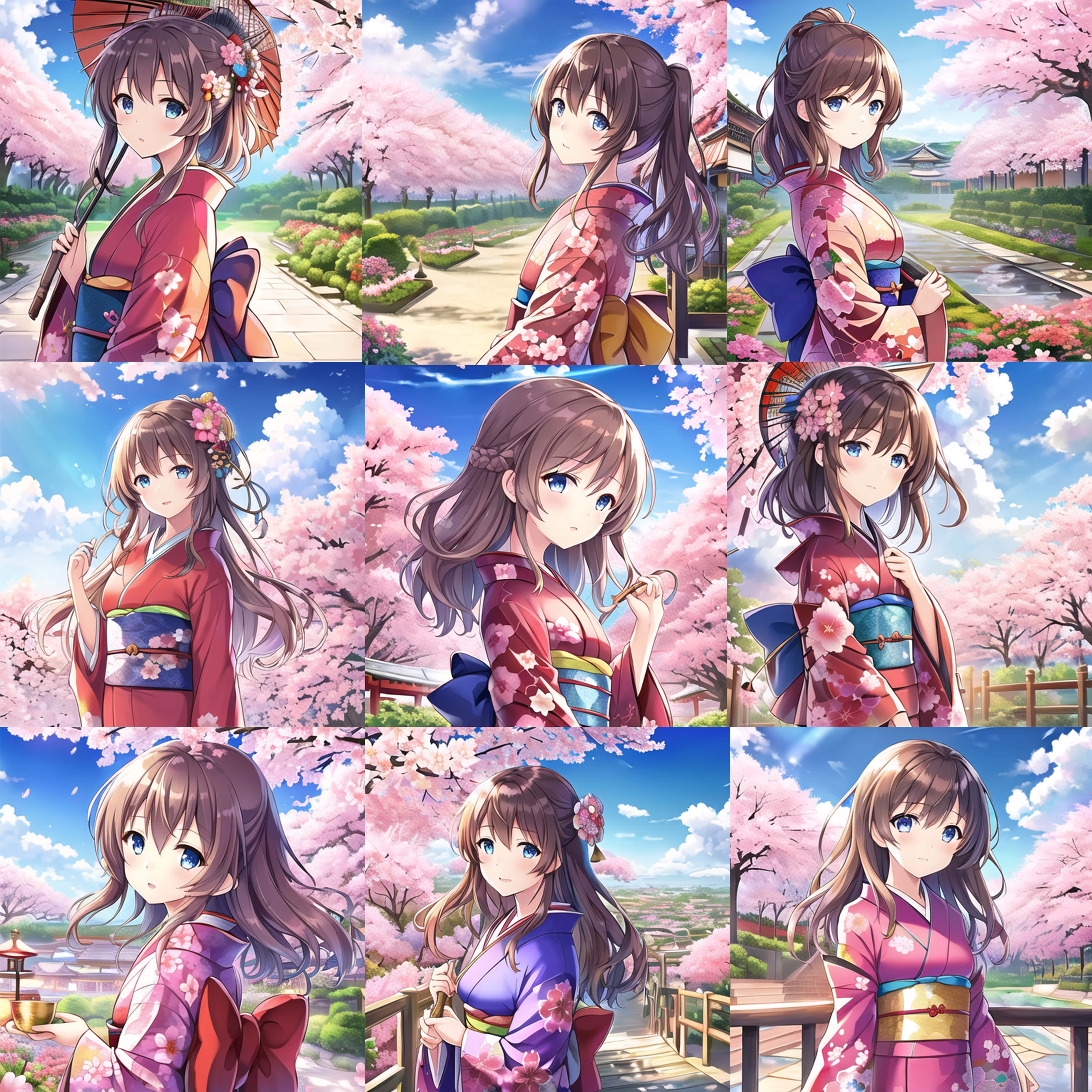}
    \caption{Animation Demo 6. The final frames in the animation sequence.Prompt: 1girl, ((kimono)), brown hair, blue eyes, colorful, spring, cumulonimbus clouds, lighting, blue sky, Sakura, garden, ((masterpiece)), (((best quality))), ((ultra-detailed)), ((illustration)), ((disheveled hair))}

\end{figure}

\bibliographystyle{ieeetr}
\bibliography{sample}

@article{The_transformative_AI,
title = {The transformative potential of artificial intelligence},
journal = {Futures},
volume = {135},
pages = {102884},
year = {2022},
issn = {0016-3287},
doi = {https://doi.org/10.1016/j.futures.2021.102884},
url = {https://www.sciencedirect.com/science/article/pii/S0016328721001932},
author = {Ross Gruetzemacher and Jess Whittlestone}
}

@article{Valuing_Data_as_an_Asset,
    author = {Veldkamp, Laura},
    title = "{Valuing Data as an Asset*}",
    journal = {Review of Finance},
    volume = {27},
    number = {5},
    pages = {1545-1562},
    year = {2023},
    month = {01},
    issn = {1572-3097},
    doi = {10.1093/rof/rfac073},
    url = {https://doi.org/10.1093/rof/rfac073},
    eprint = {https://academic.oup.com/rof/article-pdf/27/5/1545/51555961/rfac073.pdf},
}

@article{Competitive_advantage,
title = {Competitive advantage: A longitudinal analysis of the roles of data-driven innovation capabilities, marketing agility, and market turbulence},
journal = {Journal of Retailing and Consumer Services},
volume = {76},
pages = {103547},
year = {2024},
issn = {0969-6989},
doi = {https://doi.org/10.1016/j.jretconser.2023.103547},
url = {https://www.sciencedirect.com/science/article/pii/S0969698923002989},
author = {OmarA. Alghamdi and Gomaa Agag}
}

@article{ML_DL_Reviwe,
title = {Big data analytics deep learning techniques and applications: A survey},
journal = {Information Systems},
volume = {120},
pages = {102318},
year = {2024},
issn = {0306-4379},
doi = {https://doi.org/10.1016/j.is.2023.102318},
url = {https://www.sciencedirect.com/science/article/pii/S0306437923001540},
author = {Hend A. Selmy and Hoda K. Mohamed and Walaa Medhat}
}

@article{Sarker2021,
  author       = {Iqbal H. Sarker},
  title        = {Deep Learning: A Comprehensive Overview on Techniques, Taxonomy, Applications and Research Directions},
  journal      = {SN Computer Science},
  year         = {2021},
  volume       = {2},
  number       = {6},
  pages        = {420},
  doi          = {10.1007/s42979-021-00815-1},
  url          = {https://pubmed.ncbi.nlm.nih.gov/34426802/},
  publisher    = {Springer Nature Singapore Pte Ltd},
  address      = {Singapore},
  note         = {© The Author(s), under exclusive licence to Springer Nature Singapore Pte Ltd 2021},
  pmid         = {34426802},
  pmcid        = {PMC8372231},
  issn         = {2661-8907},
  eissn        = {2662-995X},
}

@article{ML_DL_app_in_health,
title = {From machine learning to deep learning: Advances of the recent data-driven paradigm shift in medicine and healthcare},
journal = {Current Research in Biotechnology},
volume = {7},
pages = {100164},
year = {2024},
issn = {2590-2628},
doi = {https://doi.org/10.1016/j.crbiot.2023.100164},
url = {https://www.sciencedirect.com/science/article/pii/S2590262823000461},
author = {Chiranjib Chakraborty and Manojit Bhattacharya and Soumen Pal and Sang-Soo Lee}
}

@Article{computers12050091,
AUTHOR = {Taye, Mohammad Mustafa},
TITLE = {Understanding of Machine Learning with Deep Learning: Architectures, Workflow, Applications and Future Directions},
JOURNAL = {Computers},
VOLUME = {12},
YEAR = {2023},
NUMBER = {5},
ARTICLE-NUMBER = {91},
URL = {https://www.mdpi.com/2073-431X/12/5/91},
ISSN = {2073-431X},
DOI = {10.3390/computers12050091}
}

@misc{vaswani2023attentionneed,
      title={Attention Is All You Need}, 
      author={Ashish Vaswani and Noam Shazeer and Niki Parmar and Jakob Uszkoreit and Llion Jones and Aidan N. Gomez and Lukasz Kaiser and Illia Polosukhin},
      year={2023},
      eprint={1706.03762},
      archivePrefix={arXiv},
      primaryClass={cs.CL},
      url={https://arxiv.org/abs/1706.03762}, 
}

@inproceedings{wolf-etal-2020-transformers,
    title = "Transformers: State-of-the-Art Natural Language Processing",
    author = "Wolf, Thomas  and
      Debut, Lysandre  and
      Sanh, Victor  and
      Chaumond, Julien  and
      Delangue, Clement  and
      Moi, Anthony  and
      Cistac, Pierric  and
      Rault, Tim  and
      Louf, Remi  and
      Funtowicz, Morgan  and
      Davison, Joe  and
      Shleifer, Sam  and
      von Platen, Patrick  and
      Ma, Clara  and
      Jernite, Yacine  and
      Plu, Julien  and
      Xu, Canwen  and
      Le Scao, Teven  and
      Gugger, Sylvain  and
      Drame, Mariama  and
      Lhoest, Quentin  and
      Rush, Alexander",
    editor = "Liu, Qun  and
      Schlangen, David",
    booktitle = "Proceedings of the 2020 Conference on Empirical Methods in Natural Language Processing: System Demonstrations",
    month = oct,
    year = "2020",
    address = "Online",
    publisher = "Association for Computational Linguistics",
    url = "https://aclanthology.org/2020.emnlp-demos.6",
    doi = "10.18653/v1/2020.emnlp-demos.6",
    pages = "38--45",
}

@article{Shahriar_2023,
   title={Let’s Have a Chat! A Conversation with ChatGPT: Technology, Applications, and Limitations},
   volume={2},
   ISSN={2811-0854},
   url={http://dx.doi.org/10.47852/bonviewAIA3202939},
   DOI={10.47852/bonviewaia3202939},
   number={1},
   journal={Artificial Intelligence and Applications},
   publisher={BON VIEW PUBLISHING PTE},
   author={Shahriar, Sakib and Hayawi, Kadhim},
   year={2023},
   month=jun, pages={11–20} }

@misc{openai2024gpt4technicalreport,
      title={GPT-4 Technical Report}, 
      author={OpenAI and Josh Achiam and Steven Adler and Sandhini Agarwal and Lama Ahmad and Ilge Akkaya and Florencia Leoni Aleman and Diogo Almeida and Janko Altenschmidt and Sam Altman and Shyamal Anadkat and Red Avila and Igor Babuschkin and Suchir Balaji and Valerie Balcom and Paul Baltescu and Haiming Bao and Mohammad Bavarian and Jeff Belgum and Irwan Bello and Jake Berdine and Gabriel Bernadett-Shapiro and Christopher Berner and Lenny Bogdonoff and Oleg Boiko and Madelaine Boyd and Anna-Luisa Brakman and Greg Brockman and Tim Brooks and Miles Brundage and Kevin Button and Trevor Cai and Rosie Campbell and Andrew Cann and Brittany Carey and Chelsea Carlson and Rory Carmichael and Brooke Chan and Che Chang and Fotis Chantzis and Derek Chen and Sully Chen and Ruby Chen and Jason Chen and Mark Chen and Ben Chess and Chester Cho and Casey Chu and Hyung Won Chung and Dave Cummings and Jeremiah Currier and Yunxing Dai and Cory Decareaux and Thomas Degry and Noah Deutsch and Damien Deville and Arka Dhar and David Dohan and Steve Dowling and Sheila Dunning and Adrien Ecoffet and Atty Eleti and Tyna Eloundou and David Farhi and Liam Fedus and Niko Felix and Simón Posada Fishman and Juston Forte and Isabella Fulford and Leo Gao and Elie Georges and Christian Gibson and Vik Goel and Tarun Gogineni and Gabriel Goh and Rapha Gontijo-Lopes and Jonathan Gordon and Morgan Grafstein and Scott Gray and Ryan Greene and Joshua Gross and Shixiang Shane Gu and Yufei Guo and Chris Hallacy and Jesse Han and Jeff Harris and Yuchen He and Mike Heaton and Johannes Heidecke and Chris Hesse and Alan Hickey and Wade Hickey and Peter Hoeschele and Brandon Houghton and Kenny Hsu and Shengli Hu and Xin Hu and Joost Huizinga and Shantanu Jain and Shawn Jain and Joanne Jang and Angela Jiang and Roger Jiang and Haozhun Jin and Denny Jin and Shino Jomoto and Billie Jonn and Heewoo Jun and Tomer Kaftan and Łukasz Kaiser and Ali Kamali and Ingmar Kanitscheider and Nitish Shirish Keskar and Tabarak Khan and Logan Kilpatrick and Jong Wook Kim and Christina Kim and Yongjik Kim and Jan Hendrik Kirchner and Jamie Kiros and Matt Knight and Daniel Kokotajlo and Łukasz Kondraciuk and Andrew Kondrich and Aris Konstantinidis and Kyle Kosic and Gretchen Krueger and Vishal Kuo and Michael Lampe and Ikai Lan and Teddy Lee and Jan Leike and Jade Leung and Daniel Levy and Chak Ming Li and Rachel Lim and Molly Lin and Stephanie Lin and Mateusz Litwin and Theresa Lopez and Ryan Lowe and Patricia Lue and Anna Makanju and Kim Malfacini and Sam Manning and Todor Markov and Yaniv Markovski and Bianca Martin and Katie Mayer and Andrew Mayne and Bob McGrew and Scott Mayer McKinney and Christine McLeavey and Paul McMillan and Jake McNeil and David Medina and Aalok Mehta and Jacob Menick and Luke Metz and Andrey Mishchenko and Pamela Mishkin and Vinnie Monaco and Evan Morikawa and Daniel Mossing and Tong Mu and Mira Murati and Oleg Murk and David Mély and Ashvin Nair and Reiichiro Nakano and Rajeev Nayak and Arvind Neelakantan and Richard Ngo and Hyeonwoo Noh and Long Ouyang and Cullen O'Keefe and Jakub Pachocki and Alex Paino and Joe Palermo and Ashley Pantuliano and Giambattista Parascandolo and Joel Parish and Emy Parparita and Alex Passos and Mikhail Pavlov and Andrew Peng and Adam Perelman and Filipe de Avila Belbute Peres and Michael Petrov and Henrique Ponde de Oliveira Pinto and Michael and Pokorny and Michelle Pokrass and Vitchyr H. Pong and Tolly Powell and Alethea Power and Boris Power and Elizabeth Proehl and Raul Puri and Alec Radford and Jack Rae and Aditya Ramesh and Cameron Raymond and Francis Real and Kendra Rimbach and Carl Ross and Bob Rotsted and Henri Roussez and Nick Ryder and Mario Saltarelli and Ted Sanders and Shibani Santurkar and Girish Sastry and Heather Schmidt and David Schnurr and John Schulman and Daniel Selsam and Kyla Sheppard and Toki Sherbakov and Jessica Shieh and Sarah Shoker and Pranav Shyam and Szymon Sidor and Eric Sigler and Maddie Simens and Jordan Sitkin and Katarina Slama and Ian Sohl and Benjamin Sokolowsky and Yang Song and Natalie Staudacher and Felipe Petroski Such and Natalie Summers and Ilya Sutskever and Jie Tang and Nikolas Tezak and Madeleine B. Thompson and Phil Tillet and Amin Tootoonchian and Elizabeth Tseng and Preston Tuggle and Nick Turley and Jerry Tworek and Juan Felipe Cerón Uribe and Andrea Vallone and Arun Vijayvergiya and Chelsea Voss and Carroll Wainwright and Justin Jay Wang and Alvin Wang and Ben Wang and Jonathan Ward and Jason Wei and CJ Weinmann and Akila Welihinda and Peter Welinder and Jiayi Weng and Lilian Weng and Matt Wiethoff and Dave Willner and Clemens Winter and Samuel Wolrich and Hannah Wong and Lauren Workman and Sherwin Wu and Jeff Wu and Michael Wu and Kai Xiao and Tao Xu and Sarah Yoo and Kevin Yu and Qiming Yuan and Wojciech Zaremba and Rowan Zellers and Chong Zhang and Marvin Zhang and Shengjia Zhao and Tianhao Zheng and Juntang Zhuang and William Zhuk and Barret Zoph},
      year={2024},
      eprint={2303.08774},
      archivePrefix={arXiv},
      primaryClass={cs.CL},
      url={https://arxiv.org/abs/2303.08774}, 
}

@inproceedings{Chen_2016, series={KDD ’16},
   title={XGBoost: A Scalable Tree Boosting System},
   volume={11},
   url={http://dx.doi.org/10.1145/2939672.2939785},
   DOI={10.1145/2939672.2939785},
   booktitle={Proceedings of the 22nd ACM SIGKDD International Conference on Knowledge Discovery and Data Mining},
   publisher={ACM},
   author={Chen, Tianqi and Guestrin, Carlos},
   year={2016},
   month=aug, pages={785–794},
   collection={KDD ’16} }

@INPROCEEDINGS{Bain_etal,
  author={Bain, Rose and Lynch, Conor and McDonnell, Darren and Witheephanich, Kritchai},
  booktitle={2022 33rd Irish Signals and Systems Conference (ISSC)}, 
  title={An XGBoost approach for industrial component degradation classification}, 
  year={2022},
  volume={},
  number={},
  pages={1-7},
  doi={10.1109/ISSC55427.2022.9826181}}

@article{NIAZKAR2024105971,
title = {Applications of XGBoost in water resources engineering: A systematic literature review (Dec 2018–May 2023)},
journal = {Environmental Modelling \& Software},
volume = {174},
pages = {105971},
year = {2024},
issn = {1364-8152},
doi = {https://doi.org/10.1016/j.envsoft.2024.105971},
url = {https://www.sciencedirect.com/science/article/pii/S136481522400032X},
author = {Majid Niazkar and Andrea Menapace and Bruno Brentan and Reza Piraei and David Jimenez and Pranav Dhawan and Maurizio Righetti}
}

@article{Silver2017,
  author    = {David Silver and Julian Schrittwieser and Karen Simonyan and Ioannis Antonoglou and Aja Huang and Arthur Guez and Thomas Hubert and Lucas Baker and Matthew Lai and Adrian Bolton and Yutian Chen and Timothy Lillicrap and Fan Hui and Laurent Sifre and George van den Driessche and Thore Graepel and Demis Hassabis},
  title     = {Mastering the game of Go without human knowledge},
  journal   = {Nature},
  year      = {2017},
  volume    = {550},
  number    = {7676},
  pages     = {354--359},
  doi       = {10.1038/nature24270},
  url       = {https://doi.org/10.1038/nature24270},
  issn      = {1476-4687}
}

@misc{he2015deepresiduallearningimage,
      title={Deep Residual Learning for Image Recognition}, 
      author={Kaiming He and Xiangyu Zhang and Shaoqing Ren and Jian Sun},
      year={2015},
      eprint={1512.03385},
      archivePrefix={arXiv},
      primaryClass={cs.CV},
      url={https://arxiv.org/abs/1512.03385}, 
}

@inproceedings{devlin-etal-2019-bert,
    title = "{BERT}: Pre-training of Deep Bidirectional Transformers for Language Understanding",
    author = "Devlin, Jacob  and
      Chang, Ming-Wei  and
      Lee, Kenton  and
      Toutanova, Kristina",
    editor = "Burstein, Jill  and
      Doran, Christy  and
      Solorio, Thamar",
    booktitle = "Proceedings of the 2019 Conference of the North {A}merican Chapter of the Association for Computational Linguistics: Human Language Technologies, Volume 1 (Long and Short Papers)",
    month = jun,
    year = "2019",
    address = "Minneapolis, Minnesota",
    publisher = "Association for Computational Linguistics",
    url = "https://aclanthology.org/N19-1423",
    doi = "10.18653/v1/N19-1423",
    pages = "4171--4186",
}

@misc{redmon2016lookonceunifiedrealtime,
      title={You Only Look Once: Unified, Real-Time Object Detection}, 
      author={Joseph Redmon and Santosh Divvala and Ross Girshick and Ali Farhadi},
      year={2016},
      eprint={1506.02640},
      archivePrefix={arXiv},
      primaryClass={cs.CV},
      url={https://arxiv.org/abs/1506.02640}, 
}

@article{Terven_2023,
   title={A Comprehensive Review of YOLO Architectures in Computer Vision: From YOLOv1 to YOLOv8 and YOLO-NAS},
   volume={5},
   ISSN={2504-4990},
   url={http://dx.doi.org/10.3390/make5040083},
   DOI={10.3390/make5040083},
   number={4},
   journal={Machine Learning and Knowledge Extraction},
   publisher={MDPI AG},
   author={Terven, Juan and Córdova-Esparza, Diana-Margarita and Romero-González, Julio-Alejandro},
   year={2023},
   month=nov, pages={1680–1716} }

@article{Jumper2021,
  author = {John Jumper and Richard Evans and Alexander Pritzel and Tim Green and Michael Figurnov and Olaf Ronneberger and Kathryn Tunyasuvunakool and Russ Bates and Augustin Žídek and Anna Potapenko and Alex Bridgland and Clemens Meyer and Simon A. A. Kohl and Andrew J. Ballard and Andrew Cowie and Bernardino Romera-Paredes and Stanislav Nikolov and Rishub Jain and Jonas Adler and Trevor Back and Stig Petersen and David Reiman and Ellen Clancy and Michal Zielinski and Martin Steinegger and Michalina Pacholska and Tamas Berghammer and Sebastian Bodenstein and David Silver and Oriol Vinyals and Andrew W. Senior and Koray Kavukcuoglu and Pushmeet Kohli and Demis Hassabis},
  title = {Highly accurate protein structure prediction with AlphaFold},
  journal = {Nature},
  volume = {596},
  number = {7873},
  pages = {583--589},
  year = {2021},
  month = {aug},
  doi = {10.1038/s41586-021-03819-2},
  url = {https://doi.org/10.1038/s41586-021-03819-2},
  issn = {1476-4687}
}

@misc{goodfellow2014generativeadversarialnetworks,
      title={Generative Adversarial Networks}, 
      author={Ian J. Goodfellow and Jean Pouget-Abadie and Mehdi Mirza and Bing Xu and David Warde-Farley and Sherjil Ozair and Aaron Courville and Yoshua Bengio},
      year={2014},
      eprint={1406.2661},
      archivePrefix={arXiv},
      primaryClass={stat.ML},
      url={https://arxiv.org/abs/1406.2661}, 
}

@article{Creswell_2018,
   title={Generative Adversarial Networks: An Overview},
   volume={35},
   ISSN={1558-0792},
   url={http://dx.doi.org/10.1109/MSP.2017.2765202},
   DOI={10.1109/msp.2017.2765202},
   number={1},
   journal={IEEE Signal Processing Magazine},
   publisher={Institute of Electrical and Electronics Engineers (IEEE)},
   author={Creswell, Antonia and White, Tom and Dumoulin, Vincent and Arulkumaran, Kai and Sengupta, Biswa and Bharath, Anil A.},
   year={2018},
   month=jan, pages={53–65} }

@misc{mnih2013playingatarideepreinforcement,
      title={Playing Atari with Deep Reinforcement Learning}, 
      author={Volodymyr Mnih and Koray Kavukcuoglu and David Silver and Alex Graves and Ioannis Antonoglou and Daan Wierstra and Martin Riedmiller},
      year={2013},
      eprint={1312.5602},
      archivePrefix={arXiv},
      primaryClass={cs.LG},
      url={https://arxiv.org/abs/1312.5602}, 
}

@misc{hausknecht2017deeprecurrentqlearningpartially,
      title={Deep Recurrent Q-Learning for Partially Observable MDPs}, 
      author={Matthew Hausknecht and Peter Stone},
      year={2017},
      eprint={1507.06527},
      archivePrefix={arXiv},
      primaryClass={cs.LG},
      url={https://arxiv.org/abs/1507.06527}, 
}

@misc{wang2016duelingnetworkarchitecturesdeep,
      title={Dueling Network Architectures for Deep Reinforcement Learning}, 
      author={Ziyu Wang and Tom Schaul and Matteo Hessel and Hado van Hasselt and Marc Lanctot and Nando de Freitas},
      year={2016},
      eprint={1511.06581},
      archivePrefix={arXiv},
      primaryClass={cs.LG},
      url={https://arxiv.org/abs/1511.06581}, 
}

@misc{vanhasselt2015deepreinforcementlearningdouble,
      title={Deep Reinforcement Learning with Double Q-learning}, 
      author={Hado van Hasselt and Arthur Guez and David Silver},
      year={2015},
      eprint={1509.06461},
      archivePrefix={arXiv},
      primaryClass={cs.LG},
      url={https://arxiv.org/abs/1509.06461}, 
}

@article{Ferreira_2017,
   title={Building an Effective Data Warehousing for Financial Sector},
   volume={3},
   ISSN={2375-1649},
   url={http://dx.doi.org/10.12691/acis-3-1-4},
   DOI={10.12691/acis-3-1-4},
   number={1},
   journal={Automatic Control and Information Sciences},
   publisher={Science and Education Publishing Co., Ltd.},
   author={Ferreira, José and Almeida, Fernando and Monteiro, José},
   year={2017},
   month=aug, pages={16–25} }

@inproceedings{Silva2016,
author = {Silva, Yasin N. and Almeida, Isadora and Queiroz, Michell},
title = {SQL: From Traditional Databases to Big Data},
year = {2016},
isbn = {9781450336857},
publisher = {Association for Computing Machinery},
address = {New York, NY, USA},
url = {https://doi.org/10.1145/2839509.2844560},
doi = {10.1145/2839509.2844560},
booktitle = {Proceedings of the 47th ACM Technical Symposium on Computing Science Education},
pages = {413–418},
numpages = {6},
location = {Memphis, Tennessee, USA},
series = {SIGCSE '16}
}

@INPROCEEDINGS{Gupta2017,
  author={Gupta, Adity and Tyagi, Swati and Panwar, Nupur and Sachdeva, Shelly and Saxena, Upaang},
  booktitle={2017 International Conference on Computing and Communication Technologies for Smart Nation (IC3TSN)}, 
  title={NoSQL databases: Critical analysis and comparison}, 
  year={2017},
  volume={},
  number={},
  pages={293-299},
  doi={10.1109/IC3TSN.2017.8284494}}

@misc{tian2022worldgraphdatabasesindustry,
      title={The World of Graph Databases from An Industry Perspective}, 
      author={Yuanyuan Tian},
      year={2022},
      eprint={2211.13170},
      archivePrefix={arXiv},
      primaryClass={cs.DB},
      url={https://arxiv.org/abs/2211.13170}, 
}

@article{Gao2023,
author = {Gao, Chen and Zheng, Yu and Li, Nian and Li, Yinfeng and Qin, Yingrong and Piao, Jinghua and Quan, Yuhan and Chang, Jianxin and Jin, Depeng and He, Xiangnan and Li, Yong},
title = {A Survey of Graph Neural Networks for Recommender Systems: Challenges, Methods, and Directions},
year = {2023},
issue_date = {March 2023},
publisher = {Association for Computing Machinery},
address = {New York, NY, USA},
volume = {1},
number = {1},
url = {https://doi.org/10.1145/3568022},
doi = {10.1145/3568022},
journal = {ACM Trans. Recomm. Syst.},
month = {mar},
articleno = {3},
numpages = {51}
}

@INPROCEEDINGS{hadoop,
  author={Shvachko, Konstantin and Kuang, Hairong and Radia, Sanjay and Chansler, Robert},
  booktitle={2010 IEEE 26th Symposium on Mass Storage Systems and Technologies (MSST)}, 
  title={The Hadoop Distributed File System}, 
  year={2010},
  volume={},
  number={},
  pages={1-10},
  doi={10.1109/MSST.2010.5496972}}

@inproceedings{spark,
author = {Zaharia, Matei and Chowdhury, Mosharaf and Franklin, Michael J. and Shenker, Scott and Stoica, Ion},
title = {Spark: cluster computing with working sets},
year = {2010},
publisher = {USENIX Association},
address = {USA},
booktitle = {Proceedings of the 2nd USENIX Conference on Hot Topics in Cloud Computing},
pages = {10},
numpages = {1},
location = {Boston, MA},
series = {HotCloud'10}
}

@Article{ML_4.0,
AUTHOR = {Mazzei, Daniele and Ramjattan, Reshawn},
TITLE = {Machine Learning for Industry 4.0: A Systematic Review Using Deep Learning-Based Topic Modelling},
JOURNAL = {Sensors},
VOLUME = {22},
YEAR = {2022},
NUMBER = {22},
ARTICLE-NUMBER = {8641},
URL = {https://www.mdpi.com/1424-8220/22/22/8641},
PubMedID = {36433236},
ISSN = {1424-8220},
DOI = {10.3390/s22228641}
}

@article{Choudhary2022,
  author = {Kamal Choudhary and Brian DeCost and Chi Chen and Anubhav Jain and Francesca Tavazza and Ryan Cohn and Cheol Woo Park and Alok Choudhary and Ankit Agrawal and Simon J. L. Billinge and Elizabeth Holm and Shyue Ping Ong and Chris Wolverton},
  title = {Recent advances and applications of deep learning methods in materials science},
  journal = {npj Computational Materials},
  year = {2022},
  volume = {8},
  number = {1},
  pages = {59},
  doi = {10.1038/s41524-022-00734-6},
  url = {https://doi.org/10.1038/s41524-022-00734-6},
  issn = {2057-3960},
  date = {2022-04-05}
}

@misc{pip,
  author = {The Python Packaging Authority},
  title = {pip},
  year = {2023},
  url = {https://pypi.org/project/pip/},
  note = {Version 23.1.2},
}

@misc{Waskom2021,
  author = {Michael Waskom},
  title = {Seaborn: Statistical Data Visualization},
  year = {2021},
  publisher = {Zenodo},
  doi = {10.5281/zenodo.592845},
  url = {https://doi.org/10.5281/zenodo.592845},
  note = {Version 0.11.2}
}

@article{Hunter2007,
  author = {John D. Hunter},
  title = {Matplotlib: A 2D Graphics Environment},
  journal = {Computing in Science \& Engineering},
  year = {2007},
  volume = {9},
  number = {3},
  pages = {90--95},
  doi = {10.1109/MCSE.2007.55},
  url = {https://matplotlib.org/},
}

@article{Pedregosa2011,
  author = {Pedregosa, Fabian and Varoquaux, Ga\"{e}l and Gramfort, Alexandre and Michel, Vincent and Thirion, Bertrand and Grisel, Olivier and Blondel, Mathieu and Prettenhofer, Peter and Weiss, Ron and Dubourg, Vincent and Vanderplas, Jake and Passos, Alexandre and Cournapeau, David and Brucher, Matthieu and Perrot, Matthieu and Duchesnay, {\'E}douard},
  title = {Scikit-learn: Machine Learning in Python},
  journal = {Journal of Machine Learning Research},
  year = {2011},
  volume = {12},
  pages = {2825--2830},
  url = {http://jmlr.org/papers/v12/pedregosa11a.html}
}

@misc{OpenAI2023,
  author = {OpenAI},
  title = {ChatGPT: September 2023 version},
  year = {2023},
  url = {https://openai.com/chatgpt},
  note = {Large language model}
}

@misc{tensorflow2015-whitepaper,
title={{TensorFlow}: Large-Scale Machine Learning on Heterogeneous Systems},
url={http://tensorflow.org/},
note={Software available from tensorflow.org},
author={
    Mart\'{\i}n~Abadi and
    Ashish~Agarwal and
    Paul~Barham and
    Eugene~Brevdo and
    Zhifeng~Chen and
    Craig~Citro and
    Greg~S.~Corrado and
    Andy~Davis and
    Jeffrey~Dean and
    Matthieu~Devin and
    Sanjay~Ghemawat and
    Ian~Goodfellow and
    Andrew~Harp and
    Geoffrey~Irving and
    Michael~Isard and
    Yangqing Jia and
    Rafal~Jozefowicz and
    Lukasz~Kaiser and
    Manjunath~Kudlur and
    Josh~Levenberg and
    Dan~Man\'{e} and
    Rajat~Monga and
    Sherry~Moore and
    Derek~Murray and
    Chris~Olah and
    Mike~Schuster and
    Jonathon~Shlens and
    Benoit~Steiner and
    Ilya~Sutskever and
    Kunal~Talwar and
    Paul~Tucker and
    Vincent~Vanhoucke and
    Vijay~Vasudevan and
    Fernanda~Vi\'{e}gas and
    Oriol~Vinyals and
    Pete~Warden and
    Martin~Wattenberg and
    Martin~Wicke and
    Yuan~Yu and
    Xiaoqiang~Zheng},
  year={2015},
}

@article{paszke2019pytorch,
  title={PyTorch: An Imperative Style, High-Performance Deep Learning Library},
  author={Paszke, Adam and Gross, Sam and Massa, Francisco and Lerer, Adam and Bradbury, James and Chanan, Gregory and Killeen, Trevor and Lin, Zeming and Gimelshein, Natalia and Antiga, Luca and Desmaison, Alban and Kopf, Andreas and Yang, Edward and DeVito, Zachary and Raison, Martin and Tejani, Alykhan and Chilamkurthy, Sasank and Steiner, Benoit and Fang, Lu and Bai, Junjie and Chintala, Soumith},
  journal={Advances in Neural Information Processing Systems},
  volume={32},
  pages={8024--8035},
  year={2019},
  url={https://papers.nips.cc/paper/9015-pytorch-an-imperative-style-high-performance-deep-learning-library.pdf}
}

@ARTICLE{MNIST,
  author={Deng, Li},
  journal={IEEE Signal Processing Magazine}, 
  title={The MNIST Database of Handwritten Digit Images for Machine Learning Research [Best of the Web]}, 
  year={2012},
  volume={29},
  number={6},
  pages={141-142},
  doi={10.1109/MSP.2012.2211477}}

@misc{ultralytics2023yolov8,
  author = {Ultralytics},
  title = {YOLOv8: State-of-the-Art Object Detection Models},
  year = {2023},
  howpublished = {\url{https://github.com/ultralytics/ultralytics}},
  note = {Version 8.0}
}

@misc{kirillov2023segment,
      title={Segment Anything}, 
      author={Alexander Kirillov and Eric Mintun and Nikhila Ravi and Hanzi Mao and Chloe Rolland and Laura Gustafson and Tete Xiao and Spencer Whitehead and Alexander C. Berg and Wan-Yen Lo and Piotr Dollár and Ross Girshick},
      year={2023},
      eprint={2304.02643},
      archivePrefix={arXiv},
      primaryClass={cs.CV},
      url={https://arxiv.org/abs/2304.02643}, 
}

@article{Digital_Avatars,
author = {Pérez-Vereda, Alejandro and Hervas, Ramon and Canal, Carlos},
year = {2022},
month = {11},
pages = {101718},
title = {Digital Avatars: A programming framework for personalized human interactions through virtual profiles},
volume = {87},
journal = {Pervasive and Mobile Computing},
doi = {10.1016/j.pmcj.2022.101718}
}

@misc{geminiteam2024geminifamilyhighlycapable,
      title={Gemini: A Family of Highly Capable Multimodal Models}, 
      author={Gemini Team and Rohan Anil and Sebastian Borgeaud and Jean-Baptiste Alayrac and Jiahui Yu and Radu Soricut and Johan Schalkwyk and Andrew M. Dai and Anja Hauth and Katie Millican and David Silver and Melvin Johnson and Ioannis Antonoglou and Julian Schrittwieser and Amelia Glaese and Jilin Chen and Emily Pitler and Timothy Lillicrap and Angeliki Lazaridou and Orhan Firat and James Molloy and Michael Isard and Paul R. Barham and Tom Hennigan and Benjamin Lee and Fabio Viola and Malcolm Reynolds and Yuanzhong Xu and Ryan Doherty and Eli Collins and Clemens Meyer and Eliza Rutherford and Erica Moreira and Kareem Ayoub and Megha Goel and Jack Krawczyk and Cosmo Du and Ed Chi and Heng-Tze Cheng and Eric Ni and Purvi Shah and Patrick Kane and Betty Chan and Manaal Faruqui and Aliaksei Severyn and Hanzhao Lin and YaGuang Li and Yong Cheng and Abe Ittycheriah and Mahdis Mahdieh and Mia Chen and Pei Sun and Dustin Tran and Sumit Bagri and Balaji Lakshminarayanan and Jeremiah Liu and Andras Orban and Fabian Güra and Hao Zhou and Xinying Song and Aurelien Boffy and Harish Ganapathy and Steven Zheng and HyunJeong Choe and Ágoston Weisz and Tao Zhu and Yifeng Lu and Siddharth Gopal and Jarrod Kahn and Maciej Kula and Jeff Pitman and Rushin Shah and Emanuel Taropa and Majd Al Merey and Martin Baeuml and Zhifeng Chen and Laurent El Shafey and Yujing Zhang and Olcan Sercinoglu and George Tucker and Enrique Piqueras and Maxim Krikun and Iain Barr and Nikolay Savinov and Ivo Danihelka and Becca Roelofs and Anaïs White and Anders Andreassen and Tamara von Glehn and Lakshman Yagati and Mehran Kazemi and Lucas Gonzalez and Misha Khalman and Jakub Sygnowski and Alexandre Frechette and Charlotte Smith and Laura Culp and Lev Proleev and Yi Luan and Xi Chen and James Lottes and Nathan Schucher and Federico Lebron and Alban Rrustemi and Natalie Clay and Phil Crone and Tomas Kocisky and Jeffrey Zhao and Bartek Perz and Dian Yu and Heidi Howard and Adam Bloniarz and Jack W. Rae and Han Lu and Laurent Sifre and Marcello Maggioni and Fred Alcober and Dan Garrette and Megan Barnes and Shantanu Thakoor and Jacob Austin and Gabriel Barth-Maron and William Wong and Rishabh Joshi and Rahma Chaabouni and Deeni Fatiha and Arun Ahuja and Gaurav Singh Tomar and Evan Senter and Martin Chadwick and Ilya Kornakov and Nithya Attaluri and Iñaki Iturrate and Ruibo Liu and Yunxuan Li and Sarah Cogan and Jeremy Chen and Chao Jia and Chenjie Gu and Qiao Zhang and Jordan Grimstad and Ale Jakse Hartman and Xavier Garcia and Thanumalayan Sankaranarayana Pillai and Jacob Devlin and Michael Laskin and Diego de Las Casas and Dasha Valter and Connie Tao and Lorenzo Blanco and Adrià Puigdomènech Badia and David Reitter and Mianna Chen and Jenny Brennan and Clara Rivera and Sergey Brin and Shariq Iqbal and Gabriela Surita and Jane Labanowski and Abhi Rao and Stephanie Winkler and Emilio Parisotto and Yiming Gu and Kate Olszewska and Ravi Addanki and Antoine Miech and Annie Louis and Denis Teplyashin and Geoff Brown and Elliot Catt and Jan Balaguer and Jackie Xiang and Pidong Wang and Zoe Ashwood and Anton Briukhov and Albert Webson and Sanjay Ganapathy and Smit Sanghavi and Ajay Kannan and Ming-Wei Chang and Axel Stjerngren and Josip Djolonga and Yuting Sun and Ankur Bapna and Matthew Aitchison and Pedram Pejman and Henryk Michalewski and Tianhe Yu and Cindy Wang and Juliette Love and Junwhan Ahn and Dawn Bloxwich and Kehang Han and Peter Humphreys and Thibault Sellam and James Bradbury and Varun Godbole and Sina Samangooei and Bogdan Damoc and Alex Kaskasoli and Sébastien M. R. Arnold and Vijay Vasudevan and Shubham Agrawal and Jason Riesa and Dmitry Lepikhin and Richard Tanburn and Srivatsan Srinivasan and Hyeontaek Lim and Sarah Hodkinson and Pranav Shyam and Johan Ferret and Steven Hand and Ankush Garg and Tom Le Paine and Jian Li and Yujia Li and Minh Giang and Alexander Neitz and Zaheer Abbas and Sarah York and Machel Reid and Elizabeth Cole and Aakanksha Chowdhery and Dipanjan Das and Dominika Rogozińska and Vitaliy Nikolaev and Pablo Sprechmann and Zachary Nado and Lukas Zilka and Flavien Prost and Luheng He and Marianne Monteiro and Gaurav Mishra and Chris Welty and Josh Newlan and Dawei Jia and Miltiadis Allamanis and Clara Huiyi Hu and Raoul de Liedekerke and Justin Gilmer and Carl Saroufim and Shruti Rijhwani and Shaobo Hou and Disha Shrivastava and Anirudh Baddepudi and Alex Goldin and Adnan Ozturel and Albin Cassirer and Yunhan Xu and Daniel Sohn and Devendra Sachan and Reinald Kim Amplayo and Craig Swanson and Dessie Petrova and Shashi Narayan and Arthur Guez and Siddhartha Brahma and Jessica Landon and Miteyan Patel and Ruizhe Zhao and Kevin Villela and Luyu Wang and Wenhao Jia and Matthew Rahtz and Mai Giménez and Legg Yeung and James Keeling and Petko Georgiev and Diana Mincu and Boxi Wu and Salem Haykal and Rachel Saputro and Kiran Vodrahalli and James Qin and Zeynep Cankara and Abhanshu Sharma and Nick Fernando and Will Hawkins and Behnam Neyshabur and Solomon Kim and Adrian Hutter and Priyanka Agrawal and Alex Castro-Ros and George van den Driessche and Tao Wang and Fan Yang and Shuo-yiin Chang and Paul Komarek and Ross McIlroy and Mario Lučić and Guodong Zhang and Wael Farhan and Michael Sharman and Paul Natsev and Paul Michel and Yamini Bansal and Siyuan Qiao and Kris Cao and Siamak Shakeri and Christina Butterfield and Justin Chung and Paul Kishan Rubenstein and Shivani Agrawal and Arthur Mensch and Kedar Soparkar and Karel Lenc and Timothy Chung and Aedan Pope and Loren Maggiore and Jackie Kay and Priya Jhakra and Shibo Wang and Joshua Maynez and Mary Phuong and Taylor Tobin and Andrea Tacchetti and Maja Trebacz and Kevin Robinson and Yash Katariya and Sebastian Riedel and Paige Bailey and Kefan Xiao and Nimesh Ghelani and Lora Aroyo and Ambrose Slone and Neil Houlsby and Xuehan Xiong and Zhen Yang and Elena Gribovskaya and Jonas Adler and Mateo Wirth and Lisa Lee and Music Li and Thais Kagohara and Jay Pavagadhi and Sophie Bridgers and Anna Bortsova and Sanjay Ghemawat and Zafarali Ahmed and Tianqi Liu and Richard Powell and Vijay Bolina and Mariko Iinuma and Polina Zablotskaia and James Besley and Da-Woon Chung and Timothy Dozat and Ramona Comanescu and Xiance Si and Jeremy Greer and Guolong Su and Martin Polacek and Raphaël Lopez Kaufman and Simon Tokumine and Hexiang Hu and Elena Buchatskaya and Yingjie Miao and Mohamed Elhawaty and Aditya Siddhant and Nenad Tomasev and Jinwei Xing and Christina Greer and Helen Miller and Shereen Ashraf and Aurko Roy and Zizhao Zhang and Ada Ma and Angelos Filos and Milos Besta and Rory Blevins and Ted Klimenko and Chih-Kuan Yeh and Soravit Changpinyo and Jiaqi Mu and Oscar Chang and Mantas Pajarskas and Carrie Muir and Vered Cohen and Charline Le Lan and Krishna Haridasan and Amit Marathe and Steven Hansen and Sholto Douglas and Rajkumar Samuel and Mingqiu Wang and Sophia Austin and Chang Lan and Jiepu Jiang and Justin Chiu and Jaime Alonso Lorenzo and Lars Lowe Sjösund and Sébastien Cevey and Zach Gleicher and Thi Avrahami and Anudhyan Boral and Hansa Srinivasan and Vittorio Selo and Rhys May and Konstantinos Aisopos and Léonard Hussenot and Livio Baldini Soares and Kate Baumli and Michael B. Chang and Adrià Recasens and Ben Caine and Alexander Pritzel and Filip Pavetic and Fabio Pardo and Anita Gergely and Justin Frye and Vinay Ramasesh and Dan Horgan and Kartikeya Badola and Nora Kassner and Subhrajit Roy and Ethan Dyer and Víctor Campos Campos and Alex Tomala and Yunhao Tang and Dalia El Badawy and Elspeth White and Basil Mustafa and Oran Lang and Abhishek Jindal and Sharad Vikram and Zhitao Gong and Sergi Caelles and Ross Hemsley and Gregory Thornton and Fangxiaoyu Feng and Wojciech Stokowiec and Ce Zheng and Phoebe Thacker and Çağlar Ünlü and Zhishuai Zhang and Mohammad Saleh and James Svensson and Max Bileschi and Piyush Patil and Ankesh Anand and Roman Ring and Katerina Tsihlas and Arpi Vezer and Marco Selvi and Toby Shevlane and Mikel Rodriguez and Tom Kwiatkowski and Samira Daruki and Keran Rong and Allan Dafoe and Nicholas FitzGerald and Keren Gu-Lemberg and Mina Khan and Lisa Anne Hendricks and Marie Pellat and Vladimir Feinberg and James Cobon-Kerr and Tara Sainath and Maribeth Rauh and Sayed Hadi Hashemi and Richard Ives and Yana Hasson and Eric Noland and Yuan Cao and Nathan Byrd and Le Hou and Qingze Wang and Thibault Sottiaux and Michela Paganini and Jean-Baptiste Lespiau and Alexandre Moufarek and Samer Hassan and Kaushik Shivakumar and Joost van Amersfoort and Amol Mandhane and Pratik Joshi and Anirudh Goyal and Matthew Tung and Andrew Brock and Hannah Sheahan and Vedant Misra and Cheng Li and Nemanja Rakićević and Mostafa Dehghani and Fangyu Liu and Sid Mittal and Junhyuk Oh and Seb Noury and Eren Sezener and Fantine Huot and Matthew Lamm and Nicola De Cao and Charlie Chen and Sidharth Mudgal and Romina Stella and Kevin Brooks and Gautam Vasudevan and Chenxi Liu and Mainak Chain and Nivedita Melinkeri and Aaron Cohen and Venus Wang and Kristie Seymore and Sergey Zubkov and Rahul Goel and Summer Yue and Sai Krishnakumaran and Brian Albert and Nate Hurley and Motoki Sano and Anhad Mohananey and Jonah Joughin and Egor Filonov and Tomasz Kępa and Yomna Eldawy and Jiawern Lim and Rahul Rishi and Shirin Badiezadegan and Taylor Bos and Jerry Chang and Sanil Jain and Sri Gayatri Sundara Padmanabhan and Subha Puttagunta and Kalpesh Krishna and Leslie Baker and Norbert Kalb and Vamsi Bedapudi and Adam Kurzrok and Shuntong Lei and Anthony Yu and Oren Litvin and Xiang Zhou and Zhichun Wu and Sam Sobell and Andrea Siciliano and Alan Papir and Robby Neale and Jonas Bragagnolo and Tej Toor and Tina Chen and Valentin Anklin and Feiran Wang and Richie Feng and Milad Gholami and Kevin Ling and Lijuan Liu and Jules Walter and Hamid Moghaddam and Arun Kishore and Jakub Adamek and Tyler Mercado and Jonathan Mallinson and Siddhinita Wandekar and Stephen Cagle and Eran Ofek and Guillermo Garrido and Clemens Lombriser and Maksim Mukha and Botu Sun and Hafeezul Rahman Mohammad and Josip Matak and Yadi Qian and Vikas Peswani and Pawel Janus and Quan Yuan and Leif Schelin and Oana David and Ankur Garg and Yifan He and Oleksii Duzhyi and Anton Älgmyr and Timothée Lottaz and Qi Li and Vikas Yadav and Luyao Xu and Alex Chinien and Rakesh Shivanna and Aleksandr Chuklin and Josie Li and Carrie Spadine and Travis Wolfe and Kareem Mohamed and Subhabrata Das and Zihang Dai and Kyle He and Daniel von Dincklage and Shyam Upadhyay and Akanksha Maurya and Luyan Chi and Sebastian Krause and Khalid Salama and Pam G Rabinovitch and Pavan Kumar Reddy M and Aarush Selvan and Mikhail Dektiarev and Golnaz Ghiasi and Erdem Guven and Himanshu Gupta and Boyi Liu and Deepak Sharma and Idan Heimlich Shtacher and Shachi Paul and Oscar Akerlund and François-Xavier Aubet and Terry Huang and Chen Zhu and Eric Zhu and Elico Teixeira and Matthew Fritze and Francesco Bertolini and Liana-Eleonora Marinescu and Martin Bölle and Dominik Paulus and Khyatti Gupta and Tejasi Latkar and Max Chang and Jason Sanders and Roopa Wilson and Xuewei Wu and Yi-Xuan Tan and Lam Nguyen Thiet and Tulsee Doshi and Sid Lall and Swaroop Mishra and Wanming Chen and Thang Luong and Seth Benjamin and Jasmine Lee and Ewa Andrejczuk and Dominik Rabiej and Vipul Ranjan and Krzysztof Styrc and Pengcheng Yin and Jon Simon and Malcolm Rose Harriott and Mudit Bansal and Alexei Robsky and Geoff Bacon and David Greene and Daniil Mirylenka and Chen Zhou and Obaid Sarvana and Abhimanyu Goyal and Samuel Andermatt and Patrick Siegler and Ben Horn and Assaf Israel and Francesco Pongetti and Chih-Wei "Louis" Chen and Marco Selvatici and Pedro Silva and Kathie Wang and Jackson Tolins and Kelvin Guu and Roey Yogev and Xiaochen Cai and Alessandro Agostini and Maulik Shah and Hung Nguyen and Noah Ó Donnaile and Sébastien Pereira and Linda Friso and Adam Stambler and Adam Kurzrok and Chenkai Kuang and Yan Romanikhin and Mark Geller and ZJ Yan and Kane Jang and Cheng-Chun Lee and Wojciech Fica and Eric Malmi and Qijun Tan and Dan Banica and Daniel Balle and Ryan Pham and Yanping Huang and Diana Avram and Hongzhi Shi and Jasjot Singh and Chris Hidey and Niharika Ahuja and Pranab Saxena and Dan Dooley and Srividya Pranavi Potharaju and Eileen O'Neill and Anand Gokulchandran and Ryan Foley and Kai Zhao and Mike Dusenberry and Yuan Liu and Pulkit Mehta and Ragha Kotikalapudi and Chalence Safranek-Shrader and Andrew Goodman and Joshua Kessinger and Eran Globen and Prateek Kolhar and Chris Gorgolewski and Ali Ibrahim and Yang Song and Ali Eichenbaum and Thomas Brovelli and Sahitya Potluri and Preethi Lahoti and Cip Baetu and Ali Ghorbani and Charles Chen and Andy Crawford and Shalini Pal and Mukund Sridhar and Petru Gurita and Asier Mujika and Igor Petrovski and Pierre-Louis Cedoz and Chenmei Li and Shiyuan Chen and Niccolò Dal Santo and Siddharth Goyal and Jitesh Punjabi and Karthik Kappaganthu and Chester Kwak and Pallavi LV and Sarmishta Velury and Himadri Choudhury and Jamie Hall and Premal Shah and Ricardo Figueira and Matt Thomas and Minjie Lu and Ting Zhou and Chintu Kumar and Thomas Jurdi and Sharat Chikkerur and Yenai Ma and Adams Yu and Soo Kwak and Victor Ähdel and Sujeevan Rajayogam and Travis Choma and Fei Liu and Aditya Barua and Colin Ji and Ji Ho Park and Vincent Hellendoorn and Alex Bailey and Taylan Bilal and Huanjie Zhou and Mehrdad Khatir and Charles Sutton and Wojciech Rzadkowski and Fiona Macintosh and Konstantin Shagin and Paul Medina and Chen Liang and Jinjing Zhou and Pararth Shah and Yingying Bi and Attila Dankovics and Shipra Banga and Sabine Lehmann and Marissa Bredesen and Zifan Lin and John Eric Hoffmann and Jonathan Lai and Raynald Chung and Kai Yang and Nihal Balani and Arthur Bražinskas and Andrei Sozanschi and Matthew Hayes and Héctor Fernández Alcalde and Peter Makarov and Will Chen and Antonio Stella and Liselotte Snijders and Michael Mandl and Ante Kärrman and Paweł Nowak and Xinyi Wu and Alex Dyck and Krishnan Vaidyanathan and Raghavender R and Jessica Mallet and Mitch Rudominer and Eric Johnston and Sushil Mittal and Akhil Udathu and Janara Christensen and Vishal Verma and Zach Irving and Andreas Santucci and Gamaleldin Elsayed and Elnaz Davoodi and Marin Georgiev and Ian Tenney and Nan Hua and Geoffrey Cideron and Edouard Leurent and Mahmoud Alnahlawi and Ionut Georgescu and Nan Wei and Ivy Zheng and Dylan Scandinaro and Heinrich Jiang and Jasper Snoek and Mukund Sundararajan and Xuezhi Wang and Zack Ontiveros and Itay Karo and Jeremy Cole and Vinu Rajashekhar and Lara Tumeh and Eyal Ben-David and Rishub Jain and Jonathan Uesato and Romina Datta and Oskar Bunyan and Shimu Wu and John Zhang and Piotr Stanczyk and Ye Zhang and David Steiner and Subhajit Naskar and Michael Azzam and Matthew Johnson and Adam Paszke and Chung-Cheng Chiu and Jaume Sanchez Elias and Afroz Mohiuddin and Faizan Muhammad and Jin Miao and Andrew Lee and Nino Vieillard and Jane Park and Jiageng Zhang and Jeff Stanway and Drew Garmon and Abhijit Karmarkar and Zhe Dong and Jong Lee and Aviral Kumar and Luowei Zhou and Jonathan Evens and William Isaac and Geoffrey Irving and Edward Loper and Michael Fink and Isha Arkatkar and Nanxin Chen and Izhak Shafran and Ivan Petrychenko and Zhe Chen and Johnson Jia and Anselm Levskaya and Zhenkai Zhu and Peter Grabowski and Yu Mao and Alberto Magni and Kaisheng Yao and Javier Snaider and Norman Casagrande and Evan Palmer and Paul Suganthan and Alfonso Castaño and Irene Giannoumis and Wooyeol Kim and Mikołaj Rybiński and Ashwin Sreevatsa and Jennifer Prendki and David Soergel and Adrian Goedeckemeyer and Willi Gierke and Mohsen Jafari and Meenu Gaba and Jeremy Wiesner and Diana Gage Wright and Yawen Wei and Harsha Vashisht and Yana Kulizhskaya and Jay Hoover and Maigo Le and Lu Li and Chimezie Iwuanyanwu and Lu Liu and Kevin Ramirez and Andrey Khorlin and Albert Cui and Tian LIN and Marcus Wu and Ricardo Aguilar and Keith Pallo and Abhishek Chakladar and Ginger Perng and Elena Allica Abellan and Mingyang Zhang and Ishita Dasgupta and Nate Kushman and Ivo Penchev and Alena Repina and Xihui Wu and Tom van der Weide and Priya Ponnapalli and Caroline Kaplan and Jiri Simsa and Shuangfeng Li and Olivier Dousse and Fan Yang and Jeff Piper and Nathan Ie and Rama Pasumarthi and Nathan Lintz and Anitha Vijayakumar and Daniel Andor and Pedro Valenzuela and Minnie Lui and Cosmin Paduraru and Daiyi Peng and Katherine Lee and Shuyuan Zhang and Somer Greene and Duc Dung Nguyen and Paula Kurylowicz and Cassidy Hardin and Lucas Dixon and Lili Janzer and Kiam Choo and Ziqiang Feng and Biao Zhang and Achintya Singhal and Dayou Du and Dan McKinnon and Natasha Antropova and Tolga Bolukbasi and Orgad Keller and David Reid and Daniel Finchelstein and Maria Abi Raad and Remi Crocker and Peter Hawkins and Robert Dadashi and Colin Gaffney and Ken Franko and Anna Bulanova and Rémi Leblond and Shirley Chung and Harry Askham and Luis C. Cobo and Kelvin Xu and Felix Fischer and Jun Xu and Christina Sorokin and Chris Alberti and Chu-Cheng Lin and Colin Evans and Alek Dimitriev and Hannah Forbes and Dylan Banarse and Zora Tung and Mark Omernick and Colton Bishop and Rachel Sterneck and Rohan Jain and Jiawei Xia and Ehsan Amid and Francesco Piccinno and Xingyu Wang and Praseem Banzal and Daniel J. Mankowitz and Alex Polozov and Victoria Krakovna and Sasha Brown and MohammadHossein Bateni and Dennis Duan and Vlad Firoiu and Meghana Thotakuri and Tom Natan and Matthieu Geist and Ser tan Girgin and Hui Li and Jiayu Ye and Ofir Roval and Reiko Tojo and Michael Kwong and James Lee-Thorp and Christopher Yew and Danila Sinopalnikov and Sabela Ramos and John Mellor and Abhishek Sharma and Kathy Wu and David Miller and Nicolas Sonnerat and Denis Vnukov and Rory Greig and Jennifer Beattie and Emily Caveness and Libin Bai and Julian Eisenschlos and Alex Korchemniy and Tomy Tsai and Mimi Jasarevic and Weize Kong and Phuong Dao and Zeyu Zheng and Frederick Liu and Fan Yang and Rui Zhu and Tian Huey Teh and Jason Sanmiya and Evgeny Gladchenko and Nejc Trdin and Daniel Toyama and Evan Rosen and Sasan Tavakkol and Linting Xue and Chen Elkind and Oliver Woodman and John Carpenter and George Papamakarios and Rupert Kemp and Sushant Kafle and Tanya Grunina and Rishika Sinha and Alice Talbert and Diane Wu and Denese Owusu-Afriyie and Cosmo Du and Chloe Thornton and Jordi Pont-Tuset and Pradyumna Narayana and Jing Li and Saaber Fatehi and John Wieting and Omar Ajmeri and Benigno Uria and Yeongil Ko and Laura Knight and Amélie Héliou and Ning Niu and Shane Gu and Chenxi Pang and Yeqing Li and Nir Levine and Ariel Stolovich and Rebeca Santamaria-Fernandez and Sonam Goenka and Wenny Yustalim and Robin Strudel and Ali Elqursh and Charlie Deck and Hyo Lee and Zonglin Li and Kyle Levin and Raphael Hoffmann and Dan Holtmann-Rice and Olivier Bachem and Sho Arora and Christy Koh and Soheil Hassas Yeganeh and Siim Põder and Mukarram Tariq and Yanhua Sun and Lucian Ionita and Mojtaba Seyedhosseini and Pouya Tafti and Zhiyu Liu and Anmol Gulati and Jasmine Liu and Xinyu Ye and Bart Chrzaszcz and Lily Wang and Nikhil Sethi and Tianrun Li and Ben Brown and Shreya Singh and Wei Fan and Aaron Parisi and Joe Stanton and Vinod Koverkathu and Christopher A. Choquette-Choo and Yunjie Li and TJ Lu and Abe Ittycheriah and Prakash Shroff and Mani Varadarajan and Sanaz Bahargam and Rob Willoughby and David Gaddy and Guillaume Desjardins and Marco Cornero and Brona Robenek and Bhavishya Mittal and Ben Albrecht and Ashish Shenoy and Fedor Moiseev and Henrik Jacobsson and Alireza Ghaffarkhah and Morgane Rivière and Alanna Walton and Clément Crepy and Alicia Parrish and Zongwei Zhou and Clement Farabet and Carey Radebaugh and Praveen Srinivasan and Claudia van der Salm and Andreas Fidjeland and Salvatore Scellato and Eri Latorre-Chimoto and Hanna Klimczak-Plucińska and David Bridson and Dario de Cesare and Tom Hudson and Piermaria Mendolicchio and Lexi Walker and Alex Morris and Matthew Mauger and Alexey Guseynov and Alison Reid and Seth Odoom and Lucia Loher and Victor Cotruta and Madhavi Yenugula and Dominik Grewe and Anastasia Petrushkina and Tom Duerig and Antonio Sanchez and Steve Yadlowsky and Amy Shen and Amir Globerson and Lynette Webb and Sahil Dua and Dong Li and Surya Bhupatiraju and Dan Hurt and Haroon Qureshi and Ananth Agarwal and Tomer Shani and Matan Eyal and Anuj Khare and Shreyas Rammohan Belle and Lei Wang and Chetan Tekur and Mihir Sanjay Kale and Jinliang Wei and Ruoxin Sang and Brennan Saeta and Tyler Liechty and Yi Sun and Yao Zhao and Stephan Lee and Pandu Nayak and Doug Fritz and Manish Reddy Vuyyuru and John Aslanides and Nidhi Vyas and Martin Wicke and Xiao Ma and Evgenii Eltyshev and Nina Martin and Hardie Cate and James Manyika and Keyvan Amiri and Yelin Kim and Xi Xiong and Kai Kang and Florian Luisier and Nilesh Tripuraneni and David Madras and Mandy Guo and Austin Waters and Oliver Wang and Joshua Ainslie and Jason Baldridge and Han Zhang and Garima Pruthi and Jakob Bauer and Feng Yang and Riham Mansour and Jason Gelman and Yang Xu and George Polovets and Ji Liu and Honglong Cai and Warren Chen and XiangHai Sheng and Emily Xue and Sherjil Ozair and Christof Angermueller and Xiaowei Li and Anoop Sinha and Weiren Wang and Julia Wiesinger and Emmanouil Koukoumidis and Yuan Tian and Anand Iyer and Madhu Gurumurthy and Mark Goldenson and Parashar Shah and MK Blake and Hongkun Yu and Anthony Urbanowicz and Jennimaria Palomaki and Chrisantha Fernando and Ken Durden and Harsh Mehta and Nikola Momchev and Elahe Rahimtoroghi and Maria Georgaki and Amit Raul and Sebastian Ruder and Morgan Redshaw and Jinhyuk Lee and Denny Zhou and Komal Jalan and Dinghua Li and Blake Hechtman and Parker Schuh and Milad Nasr and Kieran Milan and Vladimir Mikulik and Juliana Franco and Tim Green and Nam Nguyen and Joe Kelley and Aroma Mahendru and Andrea Hu and Joshua Howland and Ben Vargas and Jeffrey Hui and Kshitij Bansal and Vikram Rao and Rakesh Ghiya and Emma Wang and Ke Ye and Jean Michel Sarr and Melanie Moranski Preston and Madeleine Elish and Steve Li and Aakash Kaku and Jigar Gupta and Ice Pasupat and Da-Cheng Juan and Milan Someswar and Tejvi M. and Xinyun Chen and Aida Amini and Alex Fabrikant and Eric Chu and Xuanyi Dong and Amruta Muthal and Senaka Buthpitiya and Sarthak Jauhari and Nan Hua and Urvashi Khandelwal and Ayal Hitron and Jie Ren and Larissa Rinaldi and Shahar Drath and Avigail Dabush and Nan-Jiang Jiang and Harshal Godhia and Uli Sachs and Anthony Chen and Yicheng Fan and Hagai Taitelbaum and Hila Noga and Zhuyun Dai and James Wang and Chen Liang and Jenny Hamer and Chun-Sung Ferng and Chenel Elkind and Aviel Atias and Paulina Lee and Vít Listík and Mathias Carlen and Jan van de Kerkhof and Marcin Pikus and Krunoslav Zaher and Paul Müller and Sasha Zykova and Richard Stefanec and Vitaly Gatsko and Christoph Hirnschall and Ashwin Sethi and Xingyu Federico Xu and Chetan Ahuja and Beth Tsai and Anca Stefanoiu and Bo Feng and Keshav Dhandhania and Manish Katyal and Akshay Gupta and Atharva Parulekar and Divya Pitta and Jing Zhao and Vivaan Bhatia and Yashodha Bhavnani and Omar Alhadlaq and Xiaolin Li and Peter Danenberg and Dennis Tu and Alex Pine and Vera Filippova and Abhipso Ghosh and Ben Limonchik and Bhargava Urala and Chaitanya Krishna Lanka and Derik Clive and Yi Sun and Edward Li and Hao Wu and Kevin Hongtongsak and Ianna Li and Kalind Thakkar and Kuanysh Omarov and Kushal Majmundar and Michael Alverson and Michael Kucharski and Mohak Patel and Mudit Jain and Maksim Zabelin and Paolo Pelagatti and Rohan Kohli and Saurabh Kumar and Joseph Kim and Swetha Sankar and Vineet Shah and Lakshmi Ramachandruni and Xiangkai Zeng and Ben Bariach and Laura Weidinger and Tu Vu and Alek Andreev and Antoine He and Kevin Hui and Sheleem Kashem and Amar Subramanya and Sissie Hsiao and Demis Hassabis and Koray Kavukcuoglu and Adam Sadovsky and Quoc Le and Trevor Strohman and Yonghui Wu and Slav Petrov and Jeffrey Dean and Oriol Vinyals},
      year={2024},
      eprint={2312.11805},
      archivePrefix={arXiv},
      primaryClass={cs.CL},
      url={https://arxiv.org/abs/2312.11805}, 
}

@misc{chang2024fluxfastsoftwarebasedcommunication,
      title={FLUX: Fast Software-based Communication Overlap On GPUs Through Kernel Fusion}, 
      author={Li-Wen Chang and Wenlei Bao and Qi Hou and Chengquan Jiang and Ningxin Zheng and Yinmin Zhong and Xuanrun Zhang and Zuquan Song and Ziheng Jiang and Haibin Lin and Xin Jin and Xin Liu},
      year={2024},
      eprint={2406.06858},
      archivePrefix={arXiv},
      primaryClass={cs.LG},
      url={https://arxiv.org/abs/2406.06858}, 
}

@misc{oshea2015introductionconvolutionalneuralnetworks,
      title={An Introduction to Convolutional Neural Networks}, 
      author={Keiron O'Shea and Ryan Nash},
      year={2015},
      eprint={1511.08458},
      archivePrefix={arXiv},
      primaryClass={cs.NE},
      url={https://arxiv.org/abs/1511.08458}, 
}

@misc{cursor2023,
  author = {Cursor Team},
  title = {Cursor: An AI-Powered Coding Assistant},
  year = {2023},
  howpublished = {\url{https://github.com/getcursor/cursor}},
  note = {GitHub repository}
}

@misc{githubcopilot2023,
  author = {GitHub and OpenAI},
  title = {GitHub Copilot: Your AI Pair Programmer},
  year = {2023},
  howpublished = {\url{https://github.com/features/copilot}},
  note = {Software tool}
}

@misc{wang2024opendevinopenplatformai,
      title={OpenDevin: An Open Platform for AI Software Developers as Generalist Agents}, 
      author={Xingyao Wang and Boxuan Li and Yufan Song and Frank F. Xu and Xiangru Tang and Mingchen Zhuge and Jiayi Pan and Yueqi Song and Bowen Li and Jaskirat Singh and Hoang H. Tran and Fuqiang Li and Ren Ma and Mingzhang Zheng and Bill Qian and Yanjun Shao and Niklas Muennighoff and Yizhe Zhang and Binyuan Hui and Junyang Lin and Robert Brennan and Hao Peng and Heng Ji and Graham Neubig},
      year={2024},
      eprint={2407.16741},
      archivePrefix={arXiv},
      primaryClass={cs.SE},
      url={https://arxiv.org/abs/2407.16741}, 
}

@misc{geospy2023,
  author = {Geospy.ai Team},
  title = {Geospy.ai: Geospatial Intelligence Platform},
  year = {2023},
  howpublished = {\url{https://geospy.ai}}
}

@misc{schmidt2019recurrentneuralnetworksrnns,
      title={Recurrent Neural Networks (RNNs): A gentle Introduction and Overview}, 
      author={Robin M. Schmidt},
      year={2019},
      eprint={1912.05911},
      archivePrefix={arXiv},
      primaryClass={cs.LG},
      url={https://arxiv.org/abs/1912.05911}, 
}

@article{LSTM,
author = {Hochreiter, Sepp and Schmidhuber, Jürgen},
year = {1997},
month = {12},
pages = {1735-80},
title = {Long Short-term Memory},
volume = {9},
journal = {Neural computation},
doi = {10.1162/neco.1997.9.8.1735}
}

@article{radford2018improving,
  title={Improving language understanding by generative pre-training},
  author={Radford, Alec and Narasimhan, Karthik and Salimans, Tim and Sutskever, Ilya},
  year={2018},
  journal={OpenAI},
  url={https://cdn.openai.com/research-covers/language-unsupervised/language_understanding_paper.pdf}
}

@misc{devlin2019bertpretrainingdeepbidirectional,
      title={BERT: Pre-training of Deep Bidirectional Transformers for Language Understanding}, 
      author={Jacob Devlin and Ming-Wei Chang and Kenton Lee and Kristina Toutanova},
      year={2019},
      eprint={1810.04805},
      archivePrefix={arXiv},
      primaryClass={cs.CL},
      url={https://arxiv.org/abs/1810.04805}, 
}

@misc{python,
  author = {Python Software Foundation},
  title = {Python Language Reference, version 3.x},
  year = {2023},
  howpublished = {\url{https://www.python.org/}},
  note = {Available at https://www.python.org/}
}

@misc{waskom2023pythontips,
  author = {Michael Waskom},
  title = {Python Tips Dataset},
  year = {2023},
  howpublished = {\url{https://github.com/mwaskom/seaborn-data}},
  note = {Available in the Seaborn-data repository},
}

@article{Schneider2010,
  author = {Schneider, Astrid and Hommel, Gerhard and Blettner, Maria},
  title = {Linear Regression Analysis: Part 14 of a Series on Evaluation of Scientific Publications},
  journal = {Deutsches Ärzteblatt International},
  year = {2010},
  volume = {107},
  number = {44},
  pages = {776--782},
  month = {November},
  doi = {10.3238/arztebl.2010.0776},
  url = {https://doi.org/10.3238/arztebl.2010.0776},
  pmid = {21116397},
  pmcid = {PMC2992018}
}

@article{diabetes,
author = {Bradley Efron and Trevor Hastie and Iain Johnstone and Robert Tibshirani},
title = {{Least angle regression}},
volume = {32},
journal = {The Annals of Statistics},
number = {2},
publisher = {Institute of Mathematical Statistics},
pages = {407 -- 499},
year = {2004},
doi = {10.1214/009053604000000067},
URL = {https://doi.org/10.1214/009053604000000067}
}

@misc{yenduri2023generativepretrainedtransformercomprehensive,
      title={Generative Pre-trained Transformer: A Comprehensive Review on Enabling Technologies, Potential Applications, Emerging Challenges, and Future Directions}, 
      author={Gokul Yenduri and Ramalingam M and Chemmalar Selvi G and Supriya Y and Gautam Srivastava and Praveen Kumar Reddy Maddikunta and Deepti Raj G and Rutvij H Jhaveri and Prabadevi B and Weizheng Wang and Athanasios V. Vasilakos and Thippa Reddy Gadekallu},
      year={2023},
      eprint={2305.10435},
      archivePrefix={arXiv},
      primaryClass={cs.CL},
      url={https://arxiv.org/abs/2305.10435}, 
}

@misc{ho2020denoisingdiffusionprobabilisticmodels,
      title={Denoising Diffusion Probabilistic Models}, 
      author={Jonathan Ho and Ajay Jain and Pieter Abbeel},
      year={2020},
      eprint={2006.11239},
      archivePrefix={arXiv},
      primaryClass={cs.LG},
      url={https://arxiv.org/abs/2006.11239}, 
}

@misc{rombach2022highresolutionimagesynthesislatent,
      title={High-Resolution Image Synthesis with Latent Diffusion Models}, 
      author={Robin Rombach and Andreas Blattmann and Dominik Lorenz and Patrick Esser and Björn Ommer},
      year={2022},
      eprint={2112.10752},
      archivePrefix={arXiv},
      primaryClass={cs.CV},
      url={https://arxiv.org/abs/2112.10752}, 
}

@misc{automatic1111webui,
  author = {Automatic1111},
  title = {Stable Diffusion WebUI},
  year = {2023},
  howpublished = {\url{https://github.com/AUTOMATIC1111/stable-diffusion-webui}},
  note = {GitHub repository}
}

@misc{comfyui,
  author = {ComfyUI Team},
  title = {ComfyUI: A Powerful and Modular Stable Diffusion Graph Node System},
  year = {2023},
  howpublished = {\url{https://github.com/comfyanonymous/ComfyUI}},
  note = {GitHub repository}
}

@misc{lu2023dpmsolverfastsolverguided,
      title={DPM-Solver++: Fast Solver for Guided Sampling of Diffusion Probabilistic Models}, 
      author={Cheng Lu and Yuhao Zhou and Fan Bao and Jianfei Chen and Chongxuan Li and Jun Zhu},
      year={2023},
      eprint={2211.01095},
      archivePrefix={arXiv},
      primaryClass={cs.LG},
      url={https://arxiv.org/abs/2211.01095}, 
}

@misc{googlecolab,
  author = {Google Research},
  title = {Colaboratory: Collaborative Data Science},
  year = {2023},
  howpublished = {\url{https://colab.research.google.com/}},
}

@misc{civitai,
  author = {Civitai Team},
  title = {Civitai: AI Model Sharing Platform},
  year = {2023},
  howpublished = {\url{https://civitai.com}}
}

@misc{alphafold_image,
  author = {{AlphaFold Protein Structure Database}},
  title = {Image of Protein Structure},
  year = {2024},
  howpublished = {\url{https://alphafold.ebi.ac.uk/}},
  note = {Accessed: 2024-09-22}
}

@misc{google_gan_structure,
  author = {{Google Developers}},
  title = {GAN Structure [Image]},
  year = {2024},
  howpublished = {\url{https://developers.google.com/machine-learning/gan/gan_structure}},
  note = {Accessed: 2024-09-22}
}

@article{chen2024deeplearningmachinelearning,
      title={Deep Learning and Machine Learning, Advancing Big Data Analytics and Management: Tensorflow Pretrained Models}, 
      author={Keyu Chen and Ziqian Bi and Qian Niu and Junyu Liu and Benji Peng and Sen Zhang and Ming Liu and Ming Li and Xuanhe Pan and Jiawei Xu and Jinlang Wang and Pohsun Feng},
      year={2024},
      eprint={2409.13566},
      archivePrefix={arXiv},
      primaryClass={cs.LG},
      url={https://arxiv.org/abs/2409.13566}, 
}

\end{document}